\documentclass[10pt,twocolumn,letterpaper]{article}
\usepackage{cvpr} 
\usepackage{graphicx}
\usepackage{algorithm}
\usepackage{comment}
\usepackage{amsmath,amssymb} 
\newcommand{\myparagraph}[1]{\vspace{.01ex} \noindent  \textbf{#1}}
\usepackage{dsfont}
\usepackage{booktabs}
\usepackage{color}
\usepackage{bm}
\usepackage[accsupp]{axessibility}

\usepackage[pagebackref,breaklinks,colorlinks]{hyperref}

\usepackage[capitalize]{cleveref}
\crefname{section}{Sec.}{Secs.}
\Crefname{section}{Section}{Sections}
\Crefname{table}{Table}{Tables}
\crefname{table}{Tab.}{Tabs.}

\newcommand{\ours}[0]{xGA}

\def\cvprPaperID{3939} 
\def\confName{CVPR}
\def\confYear{2023}

\begin{document}

\title{Cross-GAN Auditing: Unsupervised Identification of Attribute Level Similarities and Differences between Pretrained Generative Models}

\author{Matthew L. Olson\textsuperscript{\rm 1}, Shusen Liu\textsuperscript{\rm 2}, Rushil Anirudh\textsuperscript{\rm 2}, Jayaraman J. Thiagarajan\textsuperscript{\rm 2}, Peer-Timo Bremer\textsuperscript{\rm 2}, \\ and Weng-Keen Wong\textsuperscript{\rm 1} \\
\textsuperscript{\rm 1} Oregon State University - EECS, \textsuperscript{\rm 2}Lawrence Livermore National Laboratory 
 - CASC\\
{\tt\small \{olsomatt,wongwe\}@oregonstate.edu, \{liu42,anirudh1,jayaramanthi1,bremer5\}}@llnl.gov \\
}

\twocolumn[{%
\renewcommand\twocolumn[1][]{#1}%
\maketitle
\begin{center}
        \centering
        \includegraphics[width=0.90\linewidth]{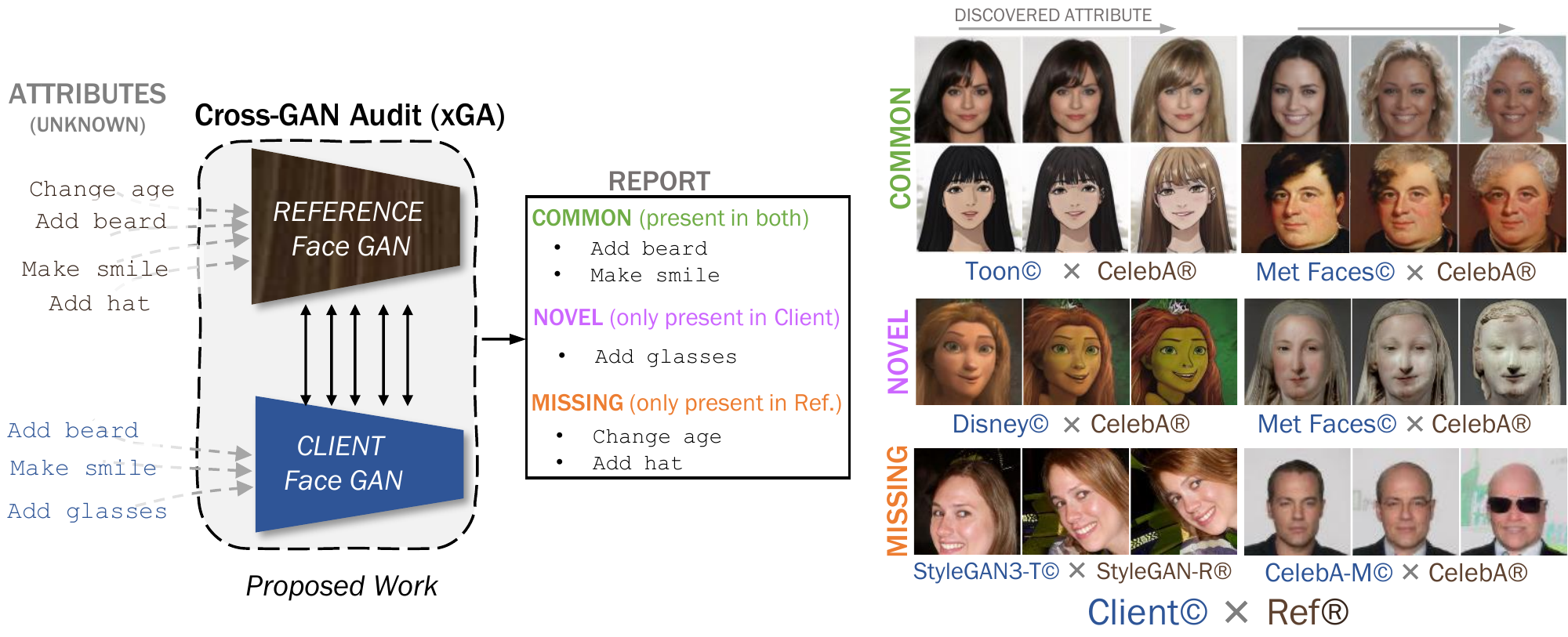}
        \captionof{figure}{
        We introduce (\textbf{xGA}) an approach for fully unsupervised cross-GAN auditing and validation. Given two pre-trained GANs (Reference \& Client), xGA evaluates the client by identifying three types of semantic attributes -- (a) Common: those that exist in both models, (b) Novel: those only present in the client and (c) Missing: those that only exist in the reference.  On the right, we show results across multiple studies, that among others include notable shifts in distribution between the Reference (CelebA) to Client (Toon, Disney, Met Faces). xGA also lends itself easily to comparing models with different properties on the same dataset as shown on the bottom right for StyleGAN3-T vs.\ StyleGAN-R. And CelebA-M is a control dataset we create that does not contain glasses, ties and smiles.
        }
        \label{fig:teaser}
\end{center}%
}]

\begin{abstract}
Generative Adversarial Networks (GANs) are notoriously difficult to train especially for complex distributions and with limited data. This has driven the need for tools to audit trained networks in human intelligible format, for example, to identify biases or ensure fairness. Existing GAN audit tools are restricted to coarse-grained, model-data comparisons based on summary statistics such as FID or recall. In this paper, we propose an alternative approach that compares a newly developed GAN against a prior baseline. To this end, we introduce \emph{Cross-GAN Auditing} ({x}GA) that, given an established ``reference" GAN and a newly proposed ``client" GAN, jointly identifies intelligible attributes that are either \emph{common} across both GANs, \emph{novel} to the client GAN, or \emph{missing} from the client GAN. This provides both users and model developers an intuitive assessment of similarity and differences between GANs. We introduce novel metrics to evaluate attribute-based GAN auditing approaches and use these metrics to demonstrate quantitatively that {x}GA outperforms baseline approaches. We also include qualitative results that illustrate the common, novel and missing attributes identified by {x}GA from GANs trained on a variety of image datasets\footnote{Source code is available at \url{https://github.com/mattolson93/cross_gan_auditing}}.  
\end{abstract}

\section{Introduction}

Generative Adversarial Networks (GANs)~\cite{goodfellow2014generative, karras2019style, karras2020analyzing, karras2021alias} have become ubiquitous in a range of high impact commercial and scientific applications \cite{beaulieu2019privacy, chen2021deepfakes,gupta2020multi, bagal2021molgpt, bian2021generative}. With this prolific use comes a growing need for investigative tools that are able to evaluate, characterize and differentiate one GAN model from another, especially since such differences can arise from a wide range of factors -- biases in training data, model architectures and hyper parameters used in training etc. In practice, this has been mostly restricted to comparing two or more GAN models against the dataset they were trained on using summary metrics such as Fr\'{e}chet Inception Distance (FID)~\cite{Heusel2017FID} and precision/recall~\cite{karras2019style} scores.

However, in many real world scenarios, different models may not even be trained on the same dataset, thereby making such summary metrics incomparable. More formally, if we define the model comparison problem as one being between a known -- and presumably well vetted -- \emph{reference} GAN and a newly developed \emph{client} GAN. For example, the reference GANs can correspond to models purchased from public market places such as AWS~\cite{aws}, Azure~\cite{azure}, or GCP~\cite{gcp}, or to community-wide standards. Furthermore, there is a critical need for more fine-grained, interpretable, investigative tools in the context of fairness and accountability. Broadly, these class of methods can be studied under the umbrella of AI model \emph{auditing}~\cite{bau2019seeing, alaa2022faithful,raji2020closing}. Here, the interpretability is used in the context to indicate that the proposed auditing result will involves of human intelligible attributes, rather than summary statistic that do not have explicit association with meaningful semantics.

While auditing classifiers has received much attention in the past \cite{raji2020closing}, GAN auditing is still a relatively new research problem with existing efforts focusing on model-data comparisons, such as identifying how faithfully a GAN recovers the original data distribution~\cite{alaa2022faithful}. In contrast, we are interested in developing a more general framework that enables a user to visually audit a ``client'' GAN model with respect the ``reference''. This framework is expected to support different kinds of auditing tasks: (i) comparing different GAN models trained on the same dataset (e.g. StyleGAN3-Rotation and StyleGAN3-Translate on FFHQ); (ii) comparing models trained on datasets with different biases (e.g., StyleGAN with race imbalance vs StyleGAN with age imbalance); and finally (iii) comparing models trained using datasets that contain challenging distribution shifts (e.g., CelebA vs Toons). Since these tools are primarily intended for human experts and auditors, interpretability is critical. Hence, it is natural to perform auditing in terms of human intelligible attributes. Though there has been encouraging progress in automatically discovering such attributes from a single GAN in the recent years \cite{yuksel2021latentclr, harkonen2020ganspace,voynov2020unsupervised,peebles2020hessian, wei2021jacobian} they are not applicable to our setting with multiple GANs.

\myparagraph{Proposed work} We introduce cross-GAN auditing (xGA), an unsupervised approach for identifying attribute similarities and differences between client GANs and reference models (which could be pre-trained and potentially unrelated). Since the GANs are trained independently, their latent spaces are disparate and encode different attributes, and thus they are not directly comparable. Consequently, discovering attributes is only one part of the solution; we also need to `align' humanly meaningful and commonly occurring attributes across the individual latent spaces. 

Our audit identifies three distinct sets of attributes: (a)~common: attributes that exist in both client and reference models; (b)~novel: attributes encoded only in the client model; (c)~missing: attributes present only in the reference. In order to identify common attributes, xGA exploits the fact that shared attributes should induce similar changes in the resulting images across both the models. On the other hand, to discover novel/missing attributes, xGA leverages the key insight that attribute manipulations unique to one GAN can be viewed as out of distribution (OOD) to the other GAN. Using empirical studies with a variety of StyleGAN models and benchmark datasets, we demonstrate that xGA is effective in providing a fine-grained characterization of generative models.

\myparagraph{Contributions} (i) We present the first cross-GAN auditing framework that uses an unified, attribute-centric method to automatically discover common, novel, and missing attributes from two or more GANs; (ii) Using an external, robust feature space for optimization, xGA produces high-quality attributes and achieves effective alignment even across challenging distribution shifts; (iii) We introduce novel metrics to evaluate attribute-based GAN auditing approaches; and 
(iv) We evaluate xGA using StyleGANs trained on CelebA, AFHQ, FFHQ, Toons, Disney and MetFaces, and also provide a suite of controlled experiments to evaluate cross-GAN auditing methods.

\section{Related Work}
\label{sec:related}

\myparagraph{Attribute Discovery} 
Several approaches have been successful in extracting attribute directions in StyleGAN's latent space in the past few years. %
InterfaceGAN~\cite{shen2020interfacegan} used an external classifier and human annotations to label sampled images in order to build a simple linear model that captures the attribute direction in a GAN's latent space. GANSpace~\cite{harkonen2020ganspace} applies PCA to these intermediate representations to find the large factors of variation and then reprojects these directions onto a GAN's latent space. Similarly, SeFa~\cite{shen2021closed} directly captures these directions via matrix factorization of the affine mapping weights in styleGAN, which identify directions of large changes without the need to sample the latent space. An alternative strategy is to directly learn the interpretable directions through a jointly-trained predictive model by assuming that the more predictive variations are more likely to be semantically meaningful \cite{voynov2020unsupervised}-- or that using a Hessian penalty \cite{peebles2020hessian}, or Jacobian \cite{wei2021jacobian}, in the image space enables learning of directions.  LatentCLR \cite{yuksel2021latentclr} used a similar optimization framework, but instead of training a separate predictive model, it leveraged the GAN's internal representation and adopted a contrastive loss \cite{chen2020simple} for attribute discovery.

\myparagraph{Model Auditing } With increased awareness of the societal impact of machine learning models, there is an increased interest in characterizing and criticizing model behavior under the broad umbrella of auditing \cite{yan22c-fairness-auditing, raji2020closing}. There has been relatively less work in auditing generative models. For example, \cite{alaa2022faithful} introduce a new performance metric for generative models that measures fidelity, diversity, and generalization. Another related work is from Bau et al., \cite{bau2019seeing} who investigate what a GAN cannot generate, whereas our interest is in distinguishing a client GAN from a reference GAN.

\myparagraph{Interpretation of Domain Shift } 
Some of the most related work comes from methods that aim for characterizing domain shift \cite{olson2021unsupervised, olson2021contrastive}, but these methods are limited to specific settings: either relying on human intervention \cite{olson2021unsupervised} or needing a disentangled generator in the input \cite{olson2021contrastive}. An indirect way to obtain aligned attributes is via \emph{aligned GANs}-- GANs where one is fine-tuned from the other ~\cite{wu2021stylealign}, \cite{pinkney2020resolution}. Here, the attribute direction will be inherent to the children models, eliminating the need to do joint discovery to identify similar attributes. However, obtaining an \emph{aligned GAN} through a separate fine-tuning process for attribute discovery across distributions is neither practical nor feasible.

\section{Methods}
We approach GAN auditing as performing attribute-level comparison to a reference GAN. For simplicity, we consider the setup where there is a single reference and client model to perform auditing, though xGA can be used even with multiple reference or client models (see experiments). Let us define the reference and client generators as $\mathcal{G}_r: \mathcal{Z}_r \mapsto \mathcal{X}_r$ and $\mathcal{G}_c: \mathcal{Z}_c \mapsto \mathcal{X}_c$ respectively. Here, $\mathcal{Z}_{r}$ and $\mathcal{Z}_{c}$ refer to the corresponding latent spaces and the generators are trained to approximate the data distributions $P_r(\mathrm{x})$ and $P_c(\mathrm{x})$. Our formulation encompasses the scenario where $P_r(\mathrm{x}) = P_c(\mathrm{x})$ but the model architectures are different, or the challenging setting of $P_r(\mathrm{x}) \neq P_c(\mathrm{x})$ (e.g., CelebA faces vs Met Faces datasets).

\begin{figure}[t]
\centering
         
         \includegraphics[width=1.00\linewidth]{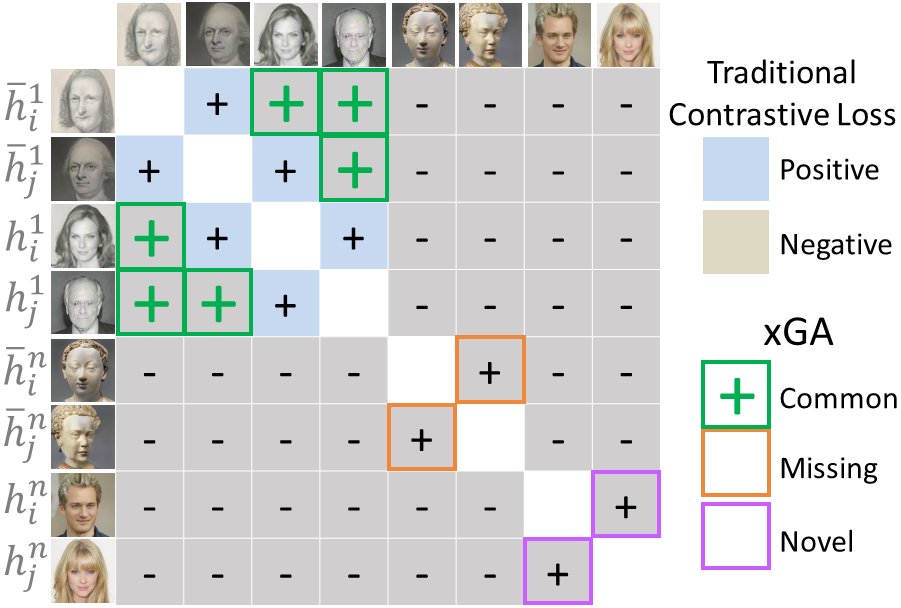}
        \caption{
        A table showing the proposed xGA modifications to typical contrastive loss with a simple two attribute model.  
        }
        \label{fig:pos_neg_table}
\end{figure}

The key idea of xGA is to audit a client model $\mathcal{G}_c$ via attribute (i.e., directions in the latent space) comparison to a reference model, in lieu of computing summary scores (e.g., FID, recall) from the synthesized images. In order to enable a fine-grained, yet interpretable, analysis of GANs, xGA performs automatic discovery and categorization of latent attributes: (i) \textit{common}: attributes that are shared by both the models; (ii) \textit{missing}: attributes that are captured by $\mathcal{G}_r$, but not $\mathcal{G}_c$; (iii) \textit{novel}: attributes that are encoded in $\mathcal{G}_c$ but not observed in the reference model. We express this new categorization scheme in figure \ref{fig:pos_neg_table}. Together, these latent attributes can provide a holistic characterization of GANs, while circumventing the need for customized metrics or human-centric analysis.

\noindent \textbf{Latent attributes}: Following state-of-the-art approaches such as LatentCLR \cite{yuksel2021latentclr}, we define attributes as direction vectors in the latent space of a GAN. For any sample $\mathrm{z} \in \mathcal{Z}_c$ and a direction vector $\mathrm{\delta}_n$, we can induce attribute-specific manipulation to the corresponding image as
\begin{equation}
\label{eq:direction}
    \mathcal{D}: (\mathrm{z}, \mathrm{\delta}_n)\rightarrow \mathrm{z} + \alpha \mathrm{\delta}_n,\mbox{ where }\mathrm{\delta}_n = \frac{\mathbf{M}_n\mathrm{z}}{\lVert \mathbf{M}_n\mathrm{z}\rVert},
\end{equation}for a scalar $\alpha$, and a learnable matrix $\mathbf{M}_n$. In other words, we consider the attribute change to be a linear model defined by the learnable direction $\mathrm{\delta}_n$. The manipulated image can then be obtained as $\mathcal{G}_c(\mathcal{D}(\mathrm{z},\mathrm{\delta}_n))$, or in shorter notation $\mathcal{G}_c(\mathrm{z},\mathrm{\delta}_n)$. Note that these latent attributes are not pre-specified and are discovered as part of the auditing process.

\subsection{Common Attribute Discovery}
Identifying common attributes between the client and reference GAN models is challenging, since it requires that the latent directions are \emph{aligned}, i.e., the exact same semantic change must be induced in unrelated latent spaces. When distilling from a parent model, i.e., training Toons from  Faces, attributes appear to align naturally, even under severe distribution shifts~\cite{wu2021stylealign}.
However, this does not hold true when the two models are trained independently, which requires us to  solve the joint problem of identifying the attributes as well as explicitly aligning them.

Formally, for a common attribute, we want the semantic change (in the generated images) induced by manipulating any sample $\mathrm{z} \in \mathcal{Z}_c$ along a direction $\delta$ in the client GAN's latent space to match the change in the direction $\bar{\delta}$ from the reference GAN's latent space for any $\bar{\mathrm{z}} \in \mathcal{Z}_r$. In other words, $\mathrm{S}(\mathcal{G}_c(\mathrm{z},\delta), \mathcal{G}_c(\mathrm{z})) \approx \mathrm{S}(\mathcal{G}_r(\bar{\mathrm{z}}, \bar{\delta}),\mathcal{G}_r(\bar{\mathrm{z}})), \forall~z \in \mathcal{Z}_c, \bar{\mathrm{z}} \in \mathcal{Z}_r$. Here, $\mathrm{S}$ denotes an \textit{oracle} detector (e.g., human subject test) which measures the semantic changes between the original sample and that obtained by manipulating the common attribute.

However, in practice, such a semantic change detector is not accessible and we need to construct a surrogate mechanism to quantify the alignment, i.e., 
\begin{multline}
\label{eq:alignment_1}
     \min_{\mathrm{\delta}_n, \bar{\mathrm{\delta}}_n} \mathcal{L}\bigg(\mathcal{G}_c(\mathrm{z},\mathrm{\delta}_n), \mathcal{G}_r(\bar{\mathrm{z}}, \bar{\mathrm{\delta}}_n)\bigg), \forall \mathrm{z}\in \mathcal{Z}_c, \forall \bar{\mathrm{z}}\in \mathcal{Z}_r,
\end{multline}for a common attribute pair $(\mathrm{\delta}_n,\bar{\mathrm{\delta}}_n)$. Any choice of the loss function $\mathcal{L}$ must satisfy two key requirements: (a) identify high-quality, latent 
directions within each of the latent spaces; 
(b) encourage cross-GAN alignment such that similar attributes end up being strongly correlated under the loss function. For example, in the case of a single GAN, the LatentCLR~\cite{yuksel2021latentclr} approach learns distinct directions using a contrastive objective that defines positive samples as those that have all been perturbed in the same direction, while manipulations in all other directions are considered negative\footnote{Other single GAN methods could be adapted, but LatentCLR's flexible loss requires less computation without the need to enforce orthogonality at every learning step.}. However, this approach is not suitable for our setting because of a key limitation -- alignment requires us to operate in a common feature space so that semantics across the two models are comparable. To address this, we first modify the objective  to operate in the latent space of an external, pre-trained feature extractor $\mathcal{F}$. In order to support alignment even in the scenario where $P_c(\mathrm{x}) \neq P_r(\mathrm{x})$, we can choose $\mathcal{F}$ that is robust to commonly occurring distributional shifts. 

Our approach works on mini-batches of size $B$ samples each, randomly drawn from $\mathcal{Z}_c$ and $\mathcal{Z}_r$ respectively. For the $i^{\text{th}}$ sample in a mini-batch from $\mathcal{Z}_c$, let us define the vector $\mathrm{h}_i^n$ as the divergence between the output of the GAN before and after perturbing along the $n^{\text{th}}$ latent direction, computed in the feature space of $\mathcal{F}$, \textit{i.e.}, 
$h_i^n = \mathcal{F}(\mathcal{G}_c(\mathrm{z}_i,\mathrm{\delta}_n)) - \mathcal{F}(\mathcal{G}_c(\mathrm{z}_i)).$ Similarly, we define the divergence $\bar{\mathrm{h}}_j^n = \mathcal{F}(\mathcal{G}_r(\bar{\mathrm{z}}_j,\bar{\mathrm{\delta}}_n)) - \mathcal{F}(\mathcal{G}_r(\bar{\mathrm{z}}_j))$ for the reference GAN. Next, we measure the semantic similarity between the divergence vectors as 
$g(\mathrm{h}_i^n,\bar{\mathrm{h}}_j^n) = \exp(\mathrm{cos}(\mathrm{h}_i^n,\bar{\mathrm{h}}_j^n)/\tau),$ where $\tau$ is the temperature parameter, and $\mathrm{cos}$ refers to cosine similarity. Now, the loss function for inferring a common attribute can be written as

\begin{equation}
\begin{split}
\label{eq:xent_1}
    \displaystyle &\mathcal{L}_{\text{xent}}(\mathrm{\delta}_n,\bar{\mathrm{\delta}}_n, \lambda_a) = \\ 
    & -\log \frac{\sum\limits_{i=1}^B \sum\limits_{j\neq i}^B 
    g(\mathrm{h}_i^n,\mathrm{h}_j^n) + g(\bar{\mathrm{h}}_i^n,\bar{\mathrm{h}}_j^n)  + \lambda_{\text{a}}g(\bar{\mathrm{h}}_i^n,\mathrm{h}_j^n)
    }{\sum\limits_{
    \substack{
       i=1 \\
       j=1
      }
    }^B\sum\limits_{l=1}^{N}\mathds{1}_{[l\neq n]}\bigg( 
    g(\mathrm{h}_i^l,\mathrm{h}_j^n) + g(\bar{\mathrm{h}}_i^l,\bar{\mathrm{h}}_j^n) + g(\bar{\mathrm{h}}_i^l,\mathrm{h}_j^n) \bigg)}
\end{split}
\end{equation}
Here $N$ denotes the total number of attributes. While the first two terms in the numerator are aimed at identifying distinct attributes from $\mathcal{G}_c$ and $\mathcal{G}_r$, the third term enforces the pair $(\delta_n, \bar{\delta}_n)$ to induce similar semantic change. When the $\lambda_a$ parameter is set to $0$, this optimization reinforces self-similarity of the attributes, without cross-similar semantics.  The terms in the denominator are based on the negative pairs (divergences from different latent directions) to enable contrastive training.

\subsection{Novel \& Missing Attribute Discovery}
A key component of our GAN auditing framework is the discovery of interpretable attributes that are unique to or missing from the client GAN's latent space. This allows practitioners to understand the novelty and limitations of a GAN model with respect to a well-established reference GAN. To this end, we exploit the key intuition that images synthesized by manipulating an attribute specific to the client model can manifest as out-of-distribution (OOD) to the reference model (and vice versa).

In order to characterize the OOD nature of such realizations, we define a likelihood score in the feature space from $\mathcal{F}$, which indicates whether a given sample is out of distribution. More specifically, we use the Density Ratio Estimation (DRE) \cite{sugiyama2012density,Nam15} method that seeks to approximate the ratio: $\mathrm{\gamma}(\mathrm{x}) = \frac{P(\mathrm{x})}{Q(\mathrm{x})}$ for any sample $\mathrm{x}$. When the ratio is low, it is likely that $\mathrm{x}$ is from the distribution $Q$ and hence OOD to $P$. We choose DRE, specifically the Kullbeck-Liebler Importance Estimation Procedure (KLIEP) \cite{sugiyama2008}, over other scoring functions because it is known to be highly effective at accurately detecting outliers \cite{menon16}.

\begin{figure}[t]
\centering
         
         \includegraphics[width=1.00\linewidth]{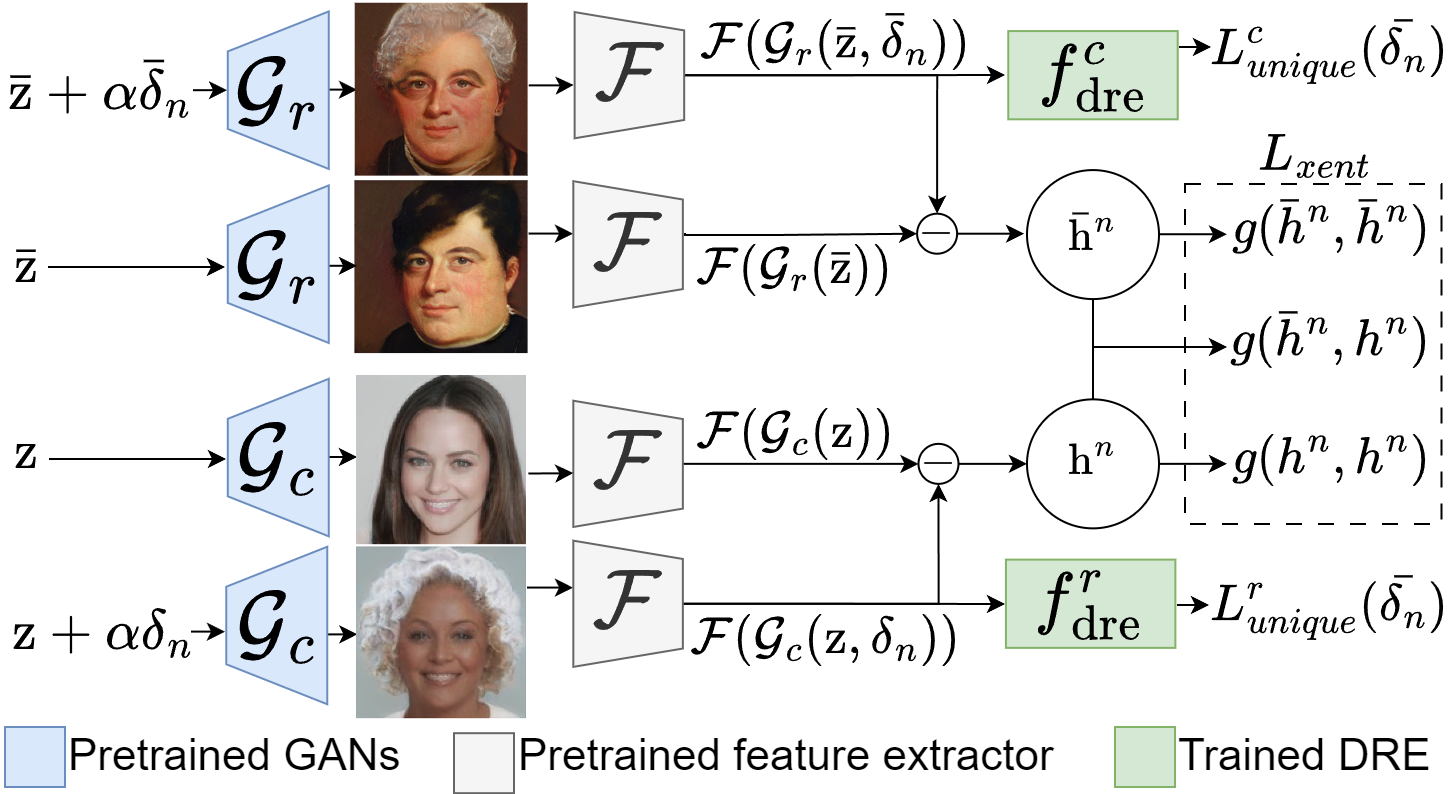}
        \caption{
        A diagram of our xGA model. $\mathcal{G}_{r}$, $\mathcal{G}_{c}$, and $\mathcal{F}$ are fixed pretrained models. $\delta_n$ and $\bar{\delta}_n$ are direction models trained to learn aligned attributes between the two Generators using the features of $\mathcal{F}$, and $f_{dre}$ are regularization models for unique attributes. 
        }
        \label{fig:latentCLR}
\end{figure}

We pre-train two separate DRE models to approximate $\mathrm{\gamma}_c(\mathrm{z})$, and $\mathcal{\gamma}_r(\bar{\mathrm{z}})$, wherein we treat data from $\mathcal{F}(\mathcal{G}_c(\mathrm{z}))$ as $P$ and $\mathcal{F}(\mathcal{G}_r(\bar{\mathrm{z}}))$ as $Q$ for the former, and vice versa for the latter. These DRE models are implemented as 2-layer MLP networks, $f^c_{\text{dre}}(.), f^r_{\text{dre}}(.)$, such that 
\begin{equation}
\label{eq:DRE_models}
    \hat{\mathrm{\gamma}}_c(\mathrm{z}) = f^c_{\text{dre}}(\mathcal{F}(\mathcal{G}_c(\mathrm{z}))) \mbox{ and }  \hat{\mathrm{\gamma}}_r(\bar{\mathrm{z}}) = f^r_{\text{dre}}(\mathcal{F}(\mathcal{G}_r(\bar{\mathrm{z}}))),
\end{equation}where $\mathcal{F}$ is the same feature extractor from \eqref{eq:xent_1}. We pass the output of the MLPs through a softplus ($\varphi(\mathrm{x}) = \log(1+e^{\mathrm{x}})$) function to ensure non-negativity. As stated previously, we use the KLIEP method to train DRE models. Using Section 4.1 of \cite{menon16}, the KLIEP loss used for training is defined as:
\begin{equation}
\label{eq:dre_train}
\mathcal{L}^c_{\text{KLIEP}} = \frac{1}{T_{2}} \sum_{j=1}^{T_{2}} \hat{\mathrm{\gamma}}_c\left(\mathrm{\bar{\mathrm{z}}}_j\right) - \frac{1}{T_{1}} \sum_{i=1}^{T_{1}} \ln \hat{\mathrm{\gamma}}_c(\mathrm{z}_i),
\end{equation}where $\bar{\mathrm{z}}_j$ and $\mathrm{z}_i$ are random samples drawn from the latent spaces $\mathcal{Z}_r$ and $\mathcal{Z}_c$ respectively (with $T_1$ and $T_2$ total samples). Similarly, we can define the KLIEP loss term for the reference model as:
\begin{equation}
\label{eq:dre_train2}
\mathcal{L}^r_{\text{KLIEP}} = \frac{1}{T_{1}} \sum_{i=1}^{T_{1}} \hat{\mathrm{\gamma}}_r\left(\mathrm{\mathrm{z}}_i\right) - \frac{1}{T_{2}} \sum_{j=1}^{T_{2}} \ln \hat{\mathrm{\gamma}}_r(\bar{\mathrm{z}}_j).
\end{equation}

We also investigated using log-loss functions to train the DRE model, but found it to be consistently inferior to the KLIEP losses (see supplement for details). Finally, we use the pre-trained DRE models from the client and reference GAN data to identify novel and missing attributes, where for a given attribute $n$ in the reference GAN, we can enforce its uniqueness by utilizing the client DRE model to give us 
$ \mathcal{L}_{\text{Unique}}^r(\mathrm{\delta}_n) = \hat{\mathrm{\gamma_c}}(\mathrm{z}, \mathrm{\delta}_n ) $ 
and similarly for the client GAN we can use the reference DRE model 
$\mathcal{L}_{\text{Unique}}^c(\mathrm{\bar{\delta}}_n) = \hat{\mathrm{\gamma_r}}(\mathrm{\bar{z}}, \mathrm{\bar{\delta}}_n ) $ 
Note, we interpret the novel attributes from the reference GAN as the missing attributes for the client GAN. 

\subsection{Overall Objective}
We now present the overall objective of xGA to identify $N_c$ common, $N_n$ novel and $N_m$ missing attributes simultaneously. Denoting the total number of attributes $N = N_c + \text{max}(N_n, N_m)$, the total loss can be written as:
\begin{equation}
\begin{split}
\label{eq:full_xga}
    \nonumber \mathcal{L}_{\text{xGA}} = & \sum_{n=1}^{N} \mathcal{L}_{\text{xent}}(\delta_n, \bar{\delta}_n, \mathds{1}_{[n\leq N_c]} \lambda_a) \\
        \nonumber & + \lambda_b \bigg[ 
        \sum_{p=N_c+1}^{N_c + N_n} 
        \mathcal{L}_{\text{Unique}}^c(\mathrm{\bar{\delta}}_p)  
          + \sum_{q=N_c + 1}^{N_c + N_m} \mathcal{L}_{\text{Unique}}^r(\mathrm{\delta}_q)  
          \bigg] 
\end{split}
\end{equation}
Here, the hyper-parameter $\lambda_b$ is the penalty for enforcing the attributes between the two latent spaces to be disparate (missing/novel). And we set $g( . , . )=0$ in $L_{xent}$ if one of the directions vectors does not exist (i.e. when $N_n \neq N_m$).

\section{Experiments}
In order to systematically evaluate the efficacy of our proposed GAN audit approach, we consider a suite of GAN models trained using several benchmark datasets. In this section, we present both qualitative and quantitative assessments of xGA, and additional results are included in the Supplementary Material.

\subsection{Datasets and GAN Models}
\label{subsec:data_gans}
For most experiments, we used a StyleGANv2~\cite{karras2019style} trained on the CelebA \cite{liu2015faceattributes} dataset as our reference GAN model. This choice is motivated both by its wide-spread use as well as the availability of fine-grained, ground truth attributes for each of the face images in CelebA, and to ensure that this model is fully independent from other client GANs (e.g., ToonGAN is finetuned from FFHQ GAN). In one experiment for the AFHQ dataset, we used a StyleGANv2 trained using only \textit{cat} images from AFHQ as the reference. Also, we considered FFHQ-trained StyleGANv3~\cite{karras2017progressive} and non-StyleGAN architectures such as GANformer~\cite{hudson2021ganformer2} for defining the reference (see supplement).

In our empirical study, we constructed a variety of (StyleGANv2) client models and performed xGA: (i) $5$ trained with different CelebA subsets constructed by excluding images specific to a chosen attribute (hat, glasses, male, female and beard); (ii) $2$ trained with CelebA subsets constructed by excluding images containing any of a chosen set of attributes (beards$\mid$hats, smiles$\mid$glasses$\mid$ties); (iv) $3$ transferred GANs for Met Faces, cartoons~\cite{cartoonStyleGan22}, and Disney images \cite{cartoonStyleGan22} respectively.

\subsection{Training Settings}
In all our experiments, xGA training is carried out for $10,000$ iterations with random samples drawn from $\mathcal{Z}_c$ and $\mathcal{Z}_r$. We fixed the desired number of attributes to be $N_c = 12$, $N_n = 4$ and $N_m = 4$. Note, this choice was to enable training xGA on a single 15GB Tesla T4 GPU. With the StyleGAN2 models, our optimization takes $4$ hours; StyleGAN3 takes $12$ hours due to gradient check-pointing. For all latent directions $\{\delta_n\}$ and $\{\bar{\delta}_n\}$, we set $\alpha=3$ and this controls how far we manipulate each sample in a given direction. In each iteration, the effective batch size was $10$, wherein $2$ samples were used to construct a positive pair and a subset of $5$ directions were randomly chosen for updating (enforced due to memory constraints).  We used the Adam \cite{kingma2014adam} optimizer with learning rate $0.001$ to update the latent direction parameters. Note, all other model parameters (generators, feature extractor, DRE models) were fixed and never updated. Following common practice with StyleGANs, the attributes are modeled in the style space and the generator's outputs are appropriately resized to fit the size requirements of the chosen feature extractor.

For our optimization objective, we set the hyper-parameter $\lambda_a = 0.1$ in $\mathcal{L}_{\text{xGA}}$. To perform $\text{DRE}$ training, we used 2-layer MLPs trained via the Adam optimizer for $1000$ iterations to minimize the KLIEP losses specified in \eqref{eq:dre_train} and \eqref{eq:dre_train2}. At each step, we constructed batches of $32$ samples from both reference and client GANs, and projected them into the feature space of $\mathcal{F}$. Lastly, we set $\lambda_b = 1.0$; we explore tuning this parameter in the supplement, finding it to be relatively insensitive.

\subsection{Evaluation: Common Attribute Discovery}
We begin by evaluating the ability of xGA in recovering common attributes across reference and client models. As mentioned earlier, for effective alignment, the choice of the feature extractor is critical. More specifically, $\mathcal{F}$ must be sufficiently expressive to uncover aligned attributes from both client and reference models. Furthermore, it is important to handle potential distribution shifts across the datasets used to train the GAN models. Hence, a feature extractor that can be robust to commonly occurring distribution shifts is expected to achieve effective alignment via \eqref{eq:xent_1}. In fact, we make an interesting observation that performing attribute discovery in such an external feature space leads to improved disentanglement in the inferred latent directions. For all results reported here, we used a robust variant of ResNet that was trained to be adversarially robust to style variations~\cite{shu2021encoding}. Please refer to the ablation in Section \ref{sec:subsec_ablation} for a comparison of different choices.

\begin{figure}[!htbp]
    \centering
    \includegraphics[width=1.0\linewidth]{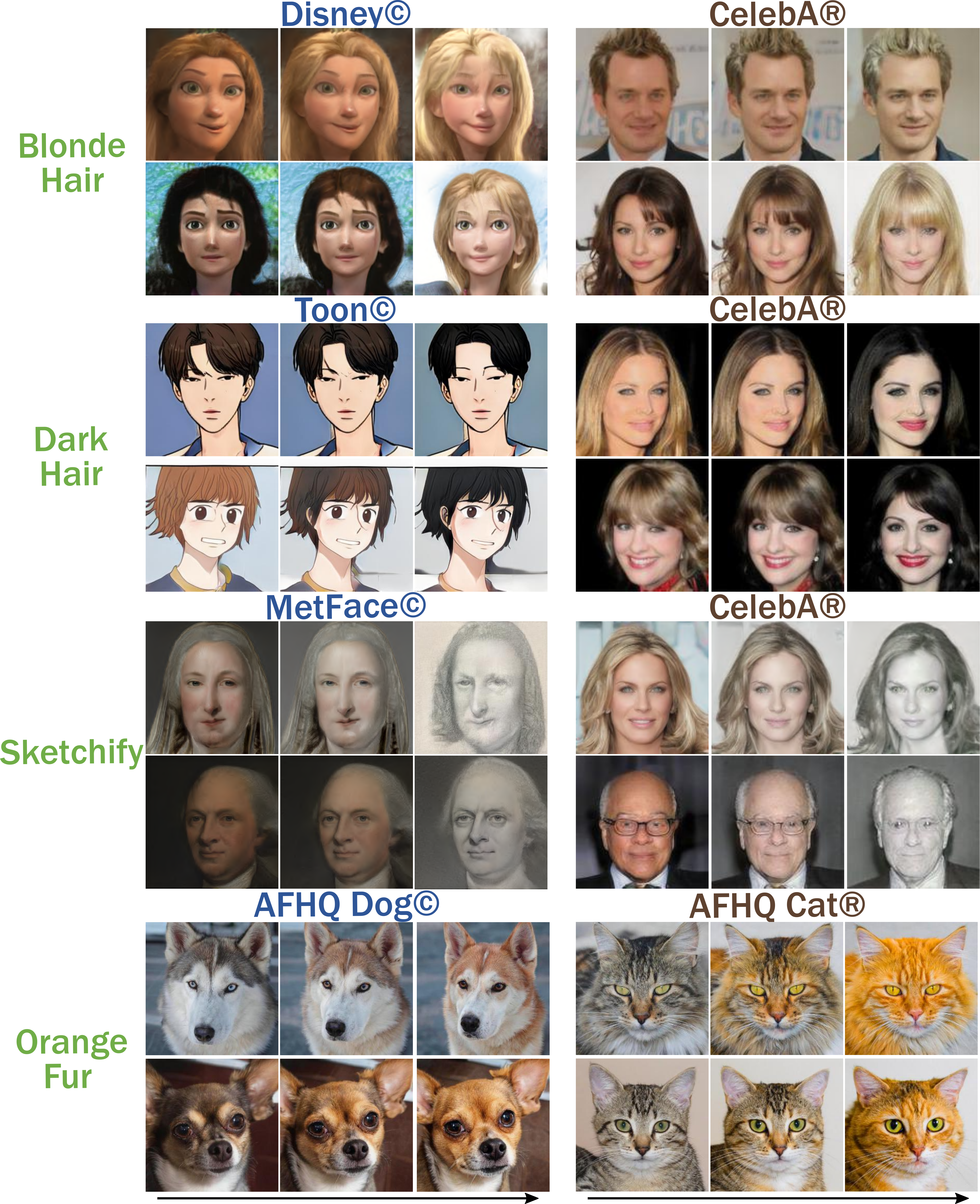}
    \caption{Visualizing common attributes discovered using xGA for different client-reference GAN pairs. For each case, we illustrate one common attribute (indicated by our description in green) with two random samples from the GAN latent space. } 
    \label{fig:metface}
\end{figure}

\myparagraph{Qualitative results } In Figure \ref{fig:metface}, we show several examples of common attributes identified by xGA for different client-reference pairs, we observe that xGA finds non-trivial attributes. For example, the ``sketchify'' attribute which naturally occurs in Met Faces (a  dataset of paintings), is surprisingly encoded even in the reference CelebA GAN (which only consists of photos of people). We also show examples of other interesting attributes such as ``orange fur'' in the case of dog-GAN $\times$ cat-GAN or ``blonde hair'' in the case of Disney-GAN $\times$ CelebA-GAN.
These results indicate that our proposed alignment objective, when coupled with a robust feature space, can effectively reveal common semantic directions across the client and reference models. We include several additional examples in the supplement. 

\myparagraph{Quantitative results }
To perform more rigorous quantitative comparisons, we setup a controlled experiment using $7$ client models corresponding to different CelebA subsets (obtained by excluding images pertinent to specific characteristics). As discussed earlier, we use a standard CelebA StyleGANv2 as the common reference model across all $7$ experiments. Next, we introduce a score of merit for common attribute discovery based on the intuition that images perturbed along the same attribute will result in similar prediction changes, when measured through an ``oracle" attribute classifier \cite{liu2015faceattributes}. 

We first generate a batch of random samples from the latent spaces of client and reference GANs, and manipulate them along a common attribute direction $(\delta_n, \bar{\delta}_n)$ inferred using xGA. In other words, we synthesize pairs of original and attribute-manipulated images from the two GANs and for each pair, we measure the discrepancy in the predictions from an ``oracle' attribute classifier. Mathematically, this can be expressed as $\mathrm{a}^n_i = |\mathcal{C}(\mathcal{G}_c(\mathrm{z}_i, \delta_n)) - \mathcal{C}(\mathcal{G}_c(\mathrm{z}_i))|$ and $\bar{\mathrm{a}}^n_j = |\mathcal{C}(\mathcal{G}_r(\bar{\mathrm{z}}_j, \bar{\delta}_n)) - \mathcal{C}(\mathcal{G}_r(\bar{\mathrm{z}}_j))|$, where $\mathcal{C}$ is the attribute classifier trained using the labeled CelebA dataset. Finally, we define an alignment score that compares the expected prediction discrepancy across the two GANs using cosine similarity (higher value indicates alignment).
\begin{align}
\label{eq:cocosine}
\mathcal{A}_{\text{score}} = & \mathbb{E}_n \bigg[\cos\bigg(\mathbb{E}_i[\mathrm{a}^n_i], \mathbb{E}_j [\bar{\mathrm{a}}^n_j] \bigg)\bigg],
\end{align}where the inner expectations are w.r.t. the batch of samples and the outer expectation is w.r.t. the $N_c$ common attributes.

We implement $5$ baseline approaches that apply state-of-the-art attribute discovery methods to the client and reference GANs (independently), and subsequently peform greedy, post-hoc alignment. In particular, we consider SeFa \cite{shen2021closed}, Voynov \cite{voynov2020unsupervised}, LatentCLR \cite{yuksel2021latentclr}, Jacobian \cite{wei2021jacobian}, and Hessian \cite{peebles2020hessian} methods for attribute discovery. Given the attributes for the two GANs, we use predictions from the ``oracle'' attribute classifier to measure the degree of alignment between every pair of directions. For example, the pair with the highest cosine similarity score is selected as the first common attribute. Next, we use the remaining latent directions to greedily pick the next attribute, and this process is repeated until we obtain $N_c=12$ attributes. We compute the alignment score from \eqref{eq:cocosine} for all the methods and report results from the $7$ controlled experiments in Table \ref{tbl:celeba_metrics_common_novel}. Interestingly, we find that, despite using the ``oracle'' classifier for alignment, the performance of the baseline methods is significantly inferior to xGA. This clearly evidences the efficacy of our optimization strategy.

\begin{table}[!tb]
\centering
\begin{tabular}{rcc}
Method & $\mathcal{A}_{\text{score}}$ ($\uparrow$)& $\mathcal{R}_{\text{score}}$ ($\uparrow$) \\ \hline
SeFa + G. S        & $0.382 \pm{0.042}$           &  $0.167	\pm{0.165}$ \\
Voynov + G. S      & $0.544	\pm{0.033}$           & $0.254	\pm{0.246}$ \\
LatentCLR + G. S   & $0.543	\pm{0.031}$           & $0.297	\pm{0.326}$ \\
Hessian + G. S     & $0.567	\pm{0.065}$           & $0.224	\pm{0.273}$ \\
Jacobian + G. S    & $0.502	\pm{0.024}$           & $0.233	\pm{0.201}$ \\ \midrule
\ours & $\mathbf{0.660} \pm{0.147}$               & $\bm{0.411} \pm{0.193}$ \\
\end{tabular}
\caption{\textbf{Common and Missing attribute discovery}. The average alignment scores from the $7$ controlled CelebA experiments. Note, we report both the mean and standard deviations ($\pm{\text{ std}}$) for each case, and ``G. S'' refers to the greedy strategy that we use for alignment.} 
\label{tbl:celeba_metrics_common_novel}
\end{table}

\begin{figure}[!tb]
    \centering
    \includegraphics[width=1.0\linewidth]{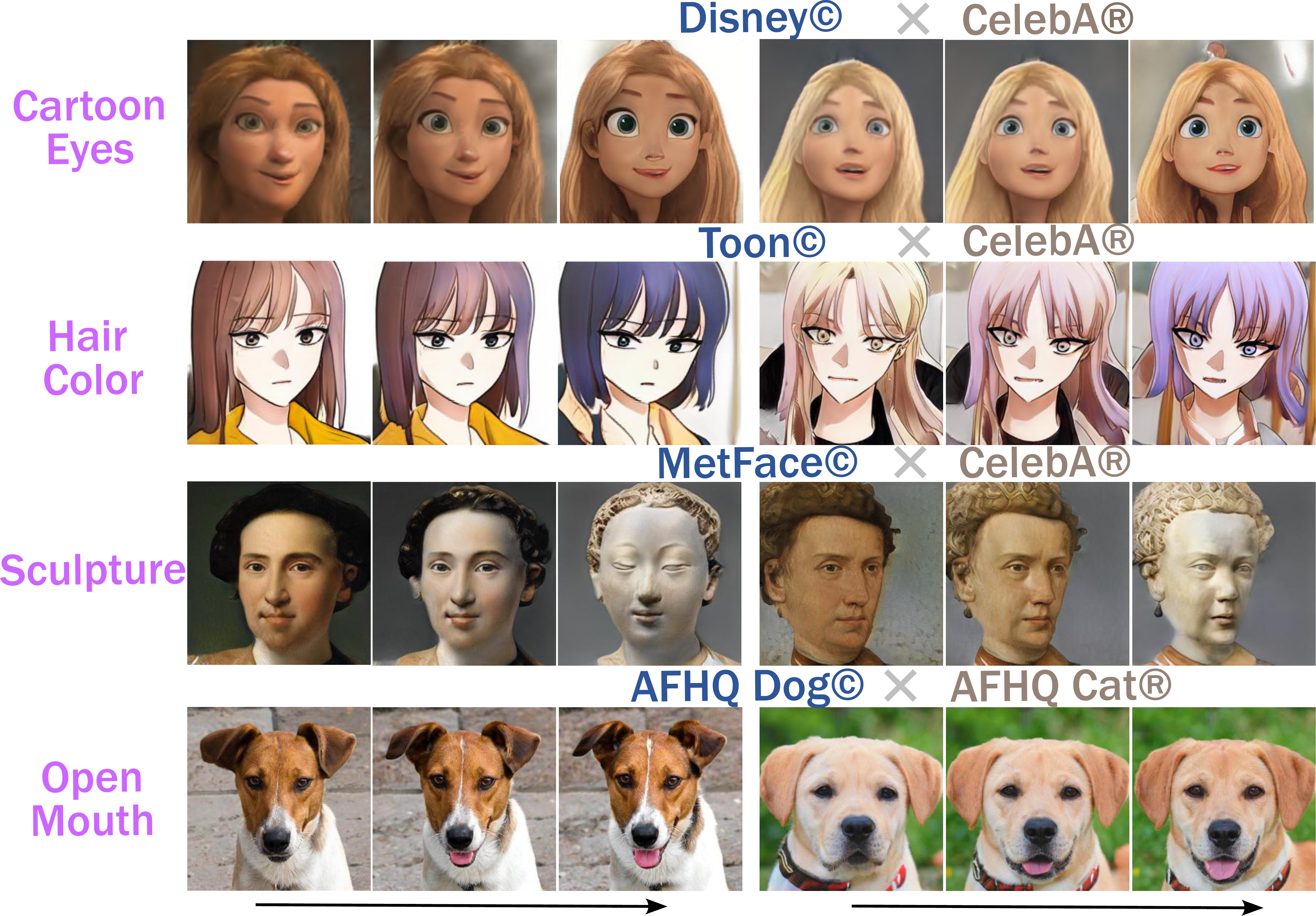}
    \caption{
    Visualizing novel attributes in different client GANs characterized by challenging distribution shifts with respect to the reference GAN (CelebA or AFHQ Cat-GANs). In each case, we show image manipulation in the attribute direction for two random sample from the latent space. 
    }
    \vspace{-2mm}
    \label{fig:cats_vs_dogs}
\end{figure}

\subsection{Evaluation: Novel/Missing Attribute Discovery}
In this section, we study the effectiveness of xGA in discovering novel (only present in the client) and missing (only present in the reference model) attributes. 

\begin{figure}[!t]
    \centering
    \includegraphics[width=0.99\linewidth]{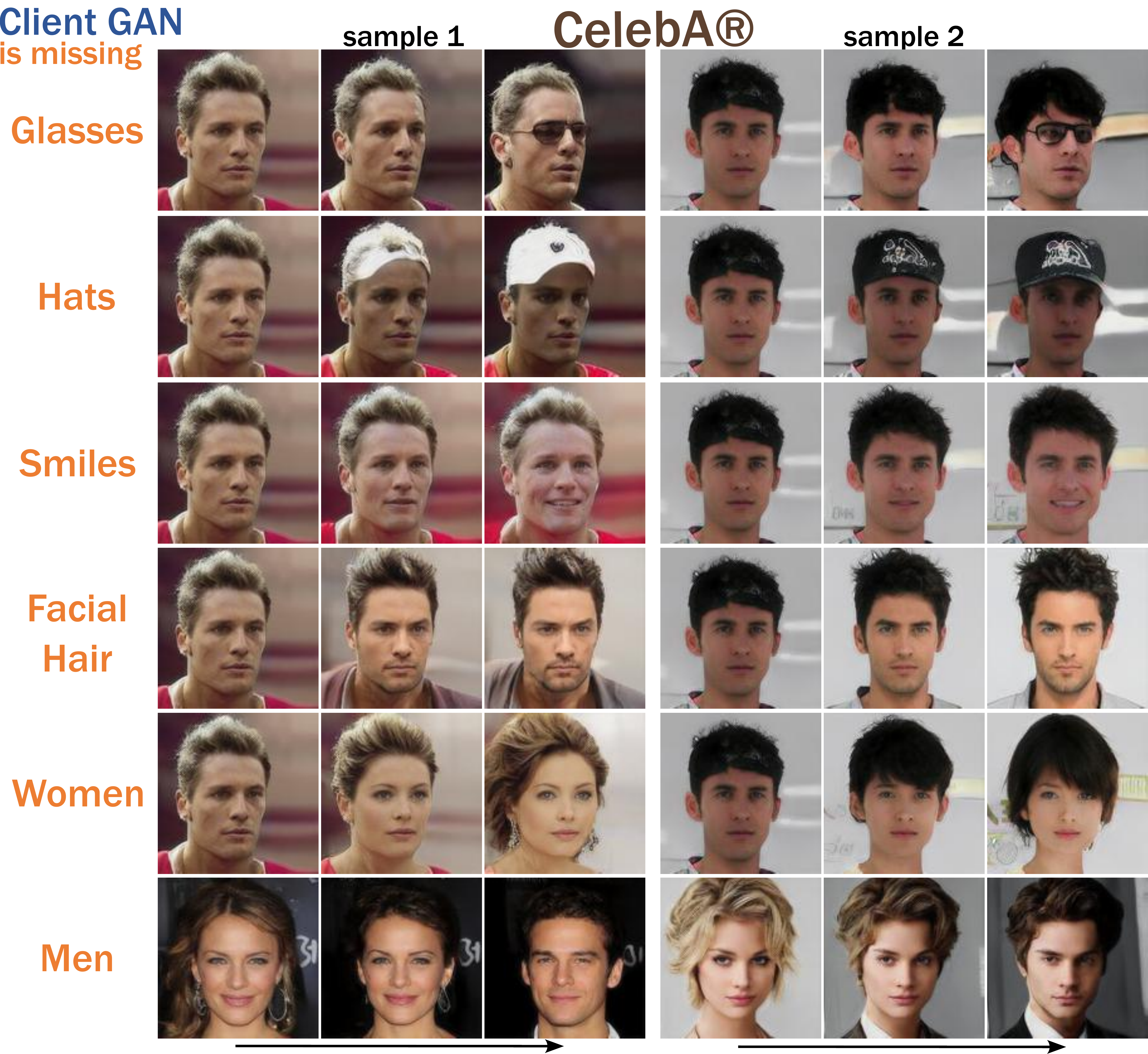}
    \caption{Using multiple clients trained with different subsets of CelebA data (one of the face attributes explicitly dropped), we find that, in all cases, xGA accurately recovers the missing attribute.}
    \vspace{-4mm}
    \label{fig:leave_out_celeba}
\end{figure}

\myparagraph{Qualitative results} 
We first show results for novel attribute discovery for different client GANs in Figure \ref{fig:cats_vs_dogs}. xGA produces highly intuitive results by identifying attributes that are unlikele to occur in the reference GAN. For example, ``cartoon eyes'' and ``sculptures'' are found to be unique to Disney and Met Faces GANs, when compared to CelebA. Next, we performed missing attribute discovery from the controlled CelebA experiments, where we know precisely which attribute is not encoded by the client GAN w.r.t the reference (standard CelebA StyleGANv2). As described earlier, the client models are always trained on a subset of data used by the reference model and by design, there are no novel attributes. Figure \ref{fig:leave_out_celeba} shows examples for the different missing attributes. We find that xGA successfully reveals each of the missing client attributes, even though the data distributions $P_c(\mathrm{x})$ and $P_r(\mathrm{x})$ are highly similar (except for a specific missing attribute).

\myparagraph{Quantitative results } 
To benchmark xGA in missing attribute discovery, we use the $7$ controlled CelebA client models and audit with respect to the reference CelebA GAN. We denote the set of attributes (one or more) which are explicitly excluded in each client model by ${\mathcal{M}}$. In order to evaluate how well xGA identifies the excluded attributes, we introduce a metric based on mean reciprocal rank (MRR) \cite{radev2002evaluating,voorhees1999trec}. For each of the $N_m$ missing attributes from xGA, we compute the average semantic discrepancy from the ``oracle'' attribute classifier as, 
$$\mathrm{a}^n = \mathbb{E}_i[|\mathcal{C}(\mathcal{G}_c(\mathrm{z}_i, \delta_n)) - \mathcal{C}(\mathcal{G}_c(\mathrm{z}_i))|].$$Denoting the rank of a missing attribute $m \in \mathcal{M}$ in the difference vector $\mathrm{a}^n$ as $\text{rank}(m,\mathrm{a}^n)$, we can define the attribute recovery (for both missing/novelty) score as:
\begin{equation}
\mathcal{R}_{\text{Score}} = \mathbb{E}_m  \bigg[\underset{n} {\max} \left( \frac{1}{\text{rank}(m,\mathrm{a}^n)} \right)\bigg]
\label{eq:score_unique}
\end{equation}In Table \ref{tbl:celeba_metrics_common_novel}, we show results for missing attribute discovery based on this score. We observe that xGA significantly outperforms all baselines in identifying the missing attribute across the suite of client GANs.

\subsection{Analysis}
\label{sec:subsec_ablation}
In this section, we examine the key components of \ours{} to understand its behavior better. 

\myparagraph{Impact of the Choice of $\mathcal{F}$ } We start by studying the choice of the external, feature space used to perform attribute discovery. For this analysis, we consider the case where we assume $\mathcal{G}_r = \mathcal{G}_c$, wherein xGA simplifies to the standard setting of attribute discovery with a single GAN model (set $\lambda_b = 0$), such as SeFA and latentCLR. We make an interesting observation that, using a robust latent space  leads to improved diversity and disentanglement in the inferred attributes, when compared to the native latent space of StyleGAN. To quantify this behavior we consider two evaluation metrics based on the predictions for a batch of synthesized images $\mathcal{G}_c(\mathrm{z}, \delta_n)$ from the ``oracle'' attribute classifier. First, for each latent direction $\delta_n$, the average prediction entropy $\mathcal{H}_{\text{score}}$~\cite{liu2015faceattributes} is defined as:

\begin{equation}
\label{eq:h_score}
\mathcal{H}_{\text{score}} =  \mathbb{E}_n \bigg[\mathbb{E}_i \bigg[\texttt{Entropy}(\mathcal{C}(\mathcal{G}_c(\mathrm{z}_i, \delta_n)))\bigg]\bigg] 
\end{equation}

Second, the deviation in the predictions across all latent directions $\mathcal{D}_{\text{score}}$ is defined in \eqref{eq:d_score}, where $K$ is the total number of attributes in the ``oracle'' classifier $\mathcal{C}$:

\begin{equation}
\label{eq:d_score}
\mathcal{D}_{\text{score}} =  \sum_{k=1}^K \texttt{Variance}\bigg[ \bigg\{\mathbb{E}_i[\mathcal{C}(\mathcal{G}_c(\mathrm{z}_i, \delta_n)]\bigg\}_{n=1}^N \bigg]_k
\end{equation}

When the entropy is low, it indicates that the semantic manipulation is concentrated to a specific attribute, and hence disentangled. On the other hand, when the deviation is high, it is reflective of the high diversity in the inferred latent directions.
\begin{table}[!tb]
\centering
\resizebox{1.0 \columnwidth}{!}{
\begin{tabular}{rll}
Method            & $\mathcal{H}_{\text{score}}$ ($\downarrow$) & $\mathcal{D}_{\text{score}}$ ($\uparrow$) \\ \hline
SeFa  \cite{shen2021closed}                          & $4.006 \pm{0.259}$ & $1.031 \pm{0.077}$  \\
	
LatentCLR \cite{yuksel2021latentclr}                  & $2.348 \pm{0.203}$ &   $0.749 \pm{0.929}$   \\ 
Voynov \cite{voynov2020unsupervised}                    & $2.508 \pm{0.069}$ &  $0.585 \pm{0.725}$    \\ 
Hessian \cite{peebles2020hessian}                    & $2.707 \pm{0.145}$ &  $0.642 \pm{0.795}$    \\ 
Jacobian \cite{wei2021jacobian}                    & $2.675 \pm{0.070}$ &  $0.661 \pm{0.826}$    \\ \midrule
\ours~(ViT)           &   $1.988 \pm{0.068}$                        &   $3.072 \pm{3.845}$   \\ 
\ours~(MAE ViT)     &   $2.102 \pm{0.035}$                          &   $3.103 \pm{3.814}$   \\ 
\ours~(CLIP ViT)     &   $2.091 \pm{0.041}$                         &   $3.135 \pm{3.901}$   \\
\ours~(ResNet-50)                  & $1.901 \pm{0.060}$             &   $3.111 \pm{3.852}$  \\ 
\ours~(Clip ResNet-50)               & $2.033 \pm{0.038}$           &   $3.121 \pm{3.863}$  \\ 
\ours~(advBN ResNet-50)           & $\mathbf{1.881} \pm{0.057}$     &  $\mathbf{3.153} \pm{3.904}$    \\ \hline
\end{tabular}
}
\caption{ 
\textbf{Choice of the feature space for attribute discovery}. Using an external feature space is superior to GAN's native style space, in terms of both entropy ($\times 100$) and deviation metrics. In this experiment, we set $\mathcal{G}_r = \mathcal{G}_c$, and aggregate the metrics from the set of controlled CelebA StyleGANs.
}
\label{tbl:entropy_results}
\end{table}

For this analysis, we considered the following feature extractors for implementing xGA: (i) vanilla ResNet-50 trained on ImageNet \cite{he2016deep}; (ii) robust variant of ResNet-50 trained with advBN\cite{shu2021encoding}; (iii) ResNet-50 trained via CLIP \cite{radford2021learning}. Table \ref{tbl:entropy_results} shows the performance of the three feature extractors on attribute discovery with our $7$ CelebA GANs trained using different data subsets. Note, we scale all entropy and diversity scores by $100$ for ease of readability. We make a striking finding that, in terms both the entropy and deviation scores, performing attribute discovery in an external feature space is significantly superior to carrying out the optimization in the native style space (all baselines). As expected, LatentCLR produces the most disentangled attributes among the baselines, and regardless of the choice of $\mathcal{F}$, xGA leads to significant improvements. More importantly, the key benefit of xGA becomes more apparent from the improvements in the deviation score over the baselines. In the supplement, we include examples for the attributes inferred using all the methods. Finally, among the different choices for $\mathcal{F}$, the advBN ResNet-50 performs the best in terms of both metrics and hence it was used in all our experiments.

\begin{figure}[!tb]
     \centering
     \includegraphics[width=0.99\linewidth]{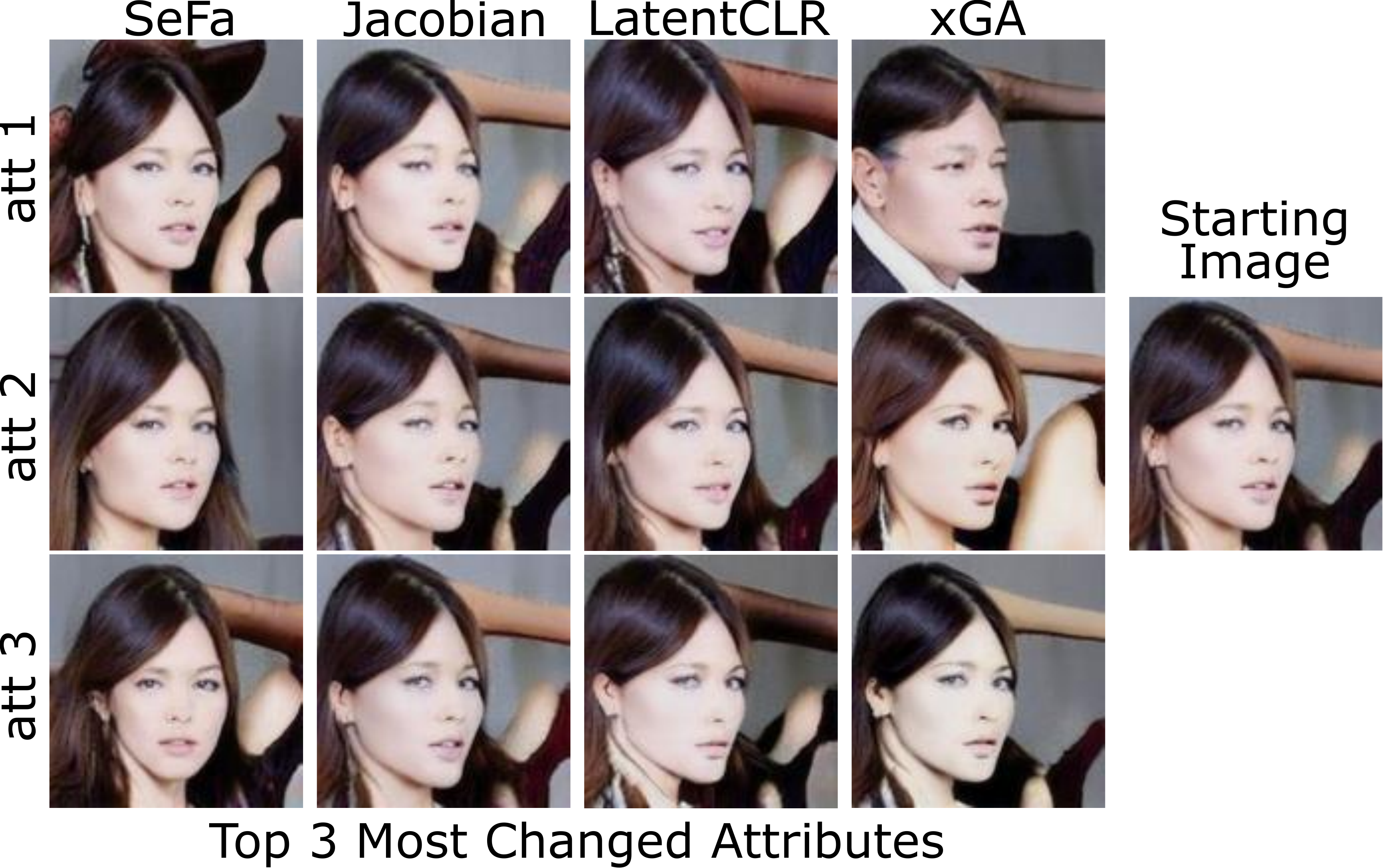}
     \caption{Comparing xGA on single GAN attribute discovery with existing approaches, we find that more diverse and novel attributes can be found simply by using an external feature space. We exploit this for effective alignment across two GAN models. 
     }
     \label{fig:1gan_examples}
 \end{figure}
 
\myparagraph{Single GAN Qualitative Results}
Figure \ref{fig:1gan_examples} visualizes a shortened example of the top $3$ attributes (induce most changes in the ``oracle'' classifier predictions). This example show a clear improvement by using a pretrained feature extractor, as xGA identifies the most diverse semantic changes. Complete results, all discovered attributes for all methods, are shown in the supplement.

\begin{figure}[!tb]
    \centering
    \includegraphics[width=0.99\linewidth]{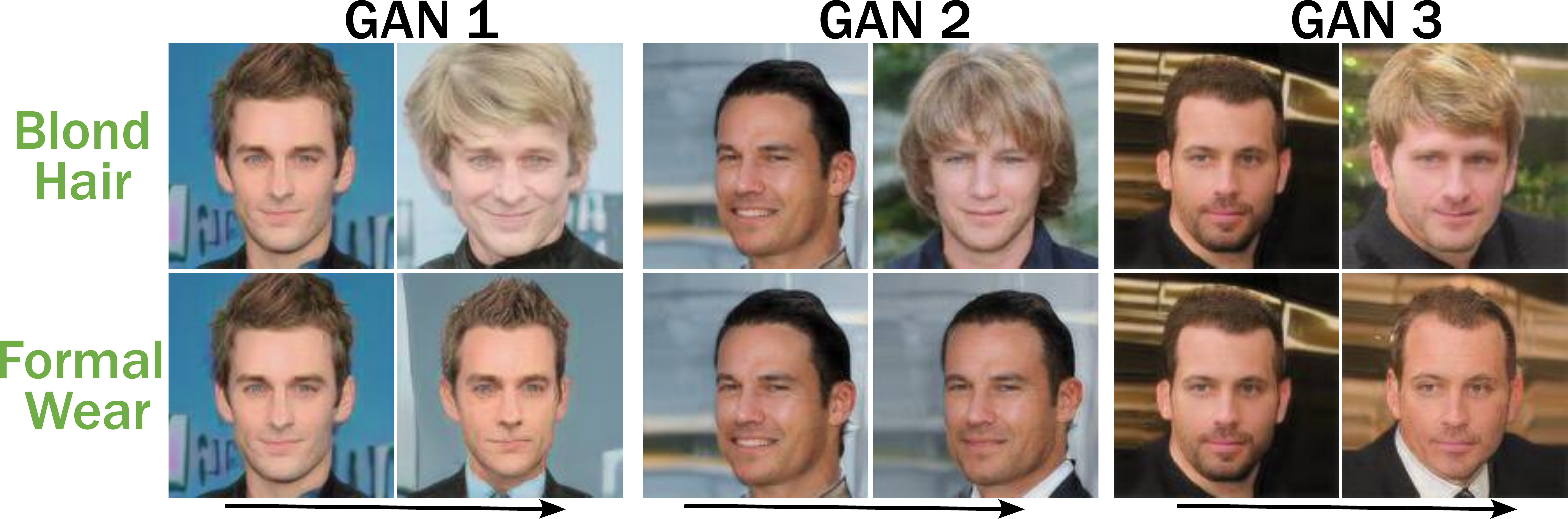}
    \caption{Common attributes identified using xGA with three different StyleGANs.} 
    \vspace{-2mm}
     \label{fig:3gan}
\end{figure}

\myparagraph{Extending xGA to compare multiple GANs } Though all our experiments used a client model w.r.t a reference, our method can be readily extended to perform comparative analysis of multiple GANs, with the only constraint arising from GPU memory since all generators need to be loaded into memory for optimization. We performed a proof-of-concept experiment by discovering common attributes across $3$ different independently trained StyleGANs as shown in Figure \ref{fig:3gan}. For this setup, we expanded the cost function outlined in \eqref{eq:xent_1} to include $3$ pairwise alignment terms from the $3$ GANs to perform contrastive training, in addition to an extra independent term from the third model. While beyond scope for the current work, scaling xGA is an important direction for future work. 

\section{Discussion}
We introduced the first cross-GAN auditing framework, which utilizes a novel optimization technique to jointly infer common, novel and missing attributes for a client GAN w.r.t any reference GAN. Through a large suite of datasets and GAN models, we demonstrate that the proposed method (i) consistently leads to higher quality (disentangled \& diverse) attributes, (ii) effectively reveals shared attributes even across challenging distribution shifts, and (iii) accurately identifies the novel/missing attributes in our controlled experiments (i.e., known ground truth). 

\noindent \textbf{Limitations } 
First, similar to other optimization-based attribute discovery approaches \cite{voynov2020unsupervised}, \cite{yuksel2021latentclr}, there is no guarantee that all prevalent factors are captured, though our controlled empirical studies clearly demonstrate the efficacy of xGA over existing approaches. Second, while using an external feature space enhances the performance of attribute discovery, this becomes an additional component that must be tuned. While we found advBN ResNet-50 to be a reasonable choice for a variety of face datasets (and AFHQ), a more systematic solution will expand the utility of our approach to other applications.

\subsection*{Acknowledgements}
This work performed under the auspices of the U.S. Department of Energy by Lawrence Livermore National Laboratory under Contract DE-AC52-07NA27344. The project is directly supported by LDRD 22-ERD-006, and DOE HRRL. The manuscript is reviewed and released under LLNL-PROC-832985.

{\small
\bibliographystyle{ieee_fullname}
\bibliography{00-references}

\begin{thebibliography}{10}\itemsep=-1pt

\bibitem{aws}
Amazon web services.
\newblock \url{https://aws.amazon.com/}.
\newblock Accessed: 2023-03-01.

\bibitem{gcp}
Google cloud: Cloud computing services.
\newblock \url{https://cloud.google.com/}.
\newblock Accessed: 2023-03-01.

\bibitem{azure}
Microsoft azure: Cloud computing services.
\newblock \url{https://azure.microsoft.com/en-us}.
\newblock Accessed: 2023-03-01.

\bibitem{alaa2022faithful}
Ahmed Alaa, Boris Van~Breugel, Evgeny~S Saveliev, and Mihaela van~der Schaar.
\newblock How faithful is your synthetic data? sample-level metrics for
  evaluating and auditing generative models.
\newblock In {\em International Conference on Machine Learning}, pages
  290--306. PMLR, 2022.

\bibitem{cartoonStyleGan22}
Jihye Back.
\newblock Fine-tuning stylegan2 for cartoon face generation.
\newblock {\em CoRR}, abs/2106.12445, 2021.

\bibitem{bagal2021molgpt}
Viraj Bagal, Rishal Aggarwal, PK Vinod, and U~Deva Priyakumar.
\newblock Molgpt: Molecular generation using a transformer-decoder model.
\newblock {\em Journal of Chemical Information and Modeling}, 2021.

\bibitem{bau2019seeing}
David Bau, Jun-Yan Zhu, Jonas Wulff, William Peebles, Hendrik Strobelt, Bolei
  Zhou, and Antonio Torralba.
\newblock Seeing what a gan cannot generate.
\newblock In {\em Proceedings of the IEEE/CVF International Conference on
  Computer Vision}, pages 4502--4511, 2019.

\bibitem{beaulieu2019privacy}
Brett~K Beaulieu-Jones, Zhiwei~Steven Wu, Chris Williams, Ran Lee, Sanjeev~P
  Bhavnani, James~Brian Byrd, and Casey~S Greene.
\newblock Privacy-preserving generative deep neural networks support clinical
  data sharing.
\newblock {\em Circulation: Cardiovascular Quality and Outcomes},
  12(7):e005122, 2019.

\bibitem{bian2021generative}
Yuemin Bian and Xiang-Qun Xie.
\newblock Generative chemistry: drug discovery with deep learning generative
  models.
\newblock {\em Journal of Molecular Modeling}, 27(3):1--18, 2021.

\bibitem{chen2021deepfakes}
Jimmy~S Chen, Aaron~S Coyner, RV~Paul Chan, M~Elizabeth Hartnett, Darius~M
  Moshfeghi, Leah~A Owen, Jayashree Kalpathy-Cramer, Michael~F Chiang, and
  J~Peter Campbell.
\newblock Deepfakes in ophthalmology: Applications and realism of synthetic
  retinal images from generative adversarial networks.
\newblock {\em Ophthalmology Science}, 1(4):100079, 2021.

\bibitem{chen2020simple}
Ting Chen, Simon Kornblith, Mohammad Norouzi, and Geoffrey Hinton.
\newblock A simple framework for contrastive learning of visual
  representations.
\newblock In {\em International conference on machine learning}, pages
  1597--1607. PMLR, 2020.

\bibitem{goodfellow2014generative}
Ian Goodfellow, Jean Pouget-Abadie, Mehdi Mirza, Bing Xu, David Warde-Farley,
  Sherjil Ozair, Aaron Courville, and Yoshua Bengio.
\newblock Generative adversarial nets.
\newblock {\em Advances in neural information processing systems}, 27, 2014.

\bibitem{gupta2020multi}
Harshit Gupta, Thong~H Phan, Jaejun Yoo, and Michael Unser.
\newblock Multi-cryogan: Reconstruction of continuous conformations in cryo-em
  using generative adversarial networks.
\newblock In {\em European Conference on Computer Vision}, pages 429--444.
  Springer, 2020.

\bibitem{harkonen2020ganspace}
Erik H{\"a}rk{\"o}nen, Aaron Hertzmann, Jaakko Lehtinen, and Sylvain Paris.
\newblock G{AN}space: Discovering interpretable {GAN} controls.
\newblock {\em arXiv preprint arXiv:2004.02546}, 2020.

\bibitem{he2016deep}
Kaiming He, Xiangyu Zhang, Shaoqing Ren, and Jian Sun.
\newblock Deep residual learning for image recognition.
\newblock In {\em Proceedings of the IEEE conference on computer vision and
  pattern recognition}, pages 770--778, 2016.

\bibitem{Heusel2017FID}
Martin Heusel, Hubert Ramsauer, Thomas Unterthiner, Bernhard Nessler, and Sepp
  Hochreiter.
\newblock Gans trained by a two time-scale update rule converge to a local nash
  equilibrium.
\newblock In I. Guyon, U.~Von Luxburg, S. Bengio, H. Wallach, R. Fergus, S.
  Vishwanathan, and R. Garnett, editors, {\em Advances in Neural Information
  Processing Systems}, volume~30. Curran Associates, Inc., 2017.

\bibitem{hudson2021ganformer2}
Drew~A Hudson and C.~Lawrence Zitnick.
\newblock Compositional transformers for scene generation.
\newblock {\em Advances in Neural Information Processing Systems {NeurIPS}
  2021}, 2021.

\bibitem{karras2017progressive}
Tero Karras, Timo Aila, Samuli Laine, and Jaakko Lehtinen.
\newblock Progressive growing of {GAN}s for improved quality, stability, and
  variation.
\newblock In {\em International Conference on Learning Representations}, 2018.

\bibitem{karras2021alias}
Tero Karras, Miika Aittala, Samuli Laine, Erik H{\"a}rk{\"o}nen, Janne
  Hellsten, Jaakko Lehtinen, and Timo Aila.
\newblock Alias-free generative adversarial networks.
\newblock {\em Advances in Neural Information Processing Systems}, 34, 2021.

\bibitem{karras2019style}
Tero Karras, Samuli Laine, and Timo Aila.
\newblock A style-based generator architecture for generative adversarial
  networks.
\newblock In {\em Proceedings of the IEEE/CVF Conference on Computer Vision and
  Pattern Recognition}, pages 4401--4410, 2019.

\bibitem{karras2020analyzing}
Tero Karras, Samuli Laine, Miika Aittala, Janne Hellsten, Jaakko Lehtinen, and
  Timo Aila.
\newblock Analyzing and improving the image quality of style{GAN}.
\newblock In {\em Proceedings of the IEEE/CVF Conference on Computer Vision and
  Pattern Recognition}, pages 8110--8119, 2020.

\bibitem{kingma2014adam}
Diederik~P Kingma and Jimmy Ba.
\newblock Adam: A method for stochastic optimization.
\newblock {\em arXiv preprint arXiv:1412.6980}, 2014.

\bibitem{liu2015faceattributes}
Ziwei Liu, Ping Luo, Xiaogang Wang, and Xiaoou Tang.
\newblock Deep learning face attributes in the wild.
\newblock In {\em Proceedings of International Conference on Computer Vision
  (ICCV)}, December 2015.

\bibitem{menon16}
A. Menon and C.~S. Ong.
\newblock Linking losses for density ratio and class-probability estimation.
\newblock In {\em Proceedings of the 33rd International Conference on Machine
  Learning}, page 304–313, 2016.

\bibitem{Nam15}
Hyunha Nam and Masashi Sugiyama.
\newblock Direct density ratio estimation with convolutional neural networks
  with application in outlier detection.
\newblock {\em IEICE Transactions on Information and Systems},
  E98.D(5):1073--1079, 2015.

\bibitem{olson2021unsupervised}
Matthew~Lyle Olson, Shusen Liu, Rushil Anirudh, Jayaraman~J Thiagarajan,
  Weng-Keen Wong, and Peer-Timo Bremer.
\newblock Unsupervised attribute alignment for characterizing distribution
  shift.
\newblock In {\em NeurIPS 2021 Workshop on Distribution Shifts: Connecting
  Methods and Applications}, 2021.

\bibitem{olson2021contrastive}
Matthew~L Olson, Thuy-Vy Nguyen, Gaurav Dixit, Neale Ratzlaff, Weng-Keen Wong,
  and Minsuk Kahng.
\newblock Contrastive identification of covariate shift in image data.
\newblock In {\em 2021 IEEE Visualization Conference (VIS)}, pages 36--40.
  IEEE, 2021.

\bibitem{peebles2020hessian}
William Peebles, John Peebles, Jun-Yan Zhu, Alexei Efros, and Antonio Torralba.
\newblock The hessian penalty: A weak prior for unsupervised disentanglement.
\newblock In {\em European Conference on Computer Vision}, pages 581--597.
  Springer, 2020.

\bibitem{pinkney2020resolution}
Justin~NM Pinkney and Doron Adler.
\newblock Resolution dependent gan interpolation for controllable image
  synthesis between domains.
\newblock {\em arXiv preprint arXiv:2010.05334}, 2020.

\bibitem{radev2002evaluating}
Dragomir~R Radev, Hong Qi, Harris Wu, and Weiguo Fan.
\newblock Evaluating web-based question answering systems.
\newblock In {\em LREC}. Citeseer, 2002.

\bibitem{radford2021learning}
Alec Radford, Jong~Wook Kim, Chris Hallacy, Aditya Ramesh, Gabriel Goh,
  Sandhini Agarwal, Girish Sastry, Amanda Askell, Pamela Mishkin, Jack Clark,
  et~al.
\newblock Learning transferable visual models from natural language
  supervision.
\newblock In {\em International Conference on Machine Learning}, pages
  8748--8763. PMLR, 2021.

\bibitem{raji2020closing}
Inioluwa~Deborah Raji, Andrew Smart, Rebecca~N White, Margaret Mitchell, Timnit
  Gebru, Ben Hutchinson, Jamila Smith-Loud, Daniel Theron, and Parker Barnes.
\newblock Closing the ai accountability gap: Defining an end-to-end framework
  for internal algorithmic auditing.
\newblock In {\em Proceedings of the 2020 conference on fairness,
  accountability, and transparency}, pages 33--44, 2020.

\bibitem{shen2020interfacegan}
Yujun Shen, Ceyuan Yang, Xiaoou Tang, and Bolei Zhou.
\newblock Interface{GAN}: Interpreting the disentangled face representation
  learned by {GAN}s.
\newblock {\em IEEE transactions on pattern analysis and machine intelligence},
  2020.

\bibitem{shen2021closed}
Yujun Shen and Bolei Zhou.
\newblock Closed-form factorization of latent semantics in {GAN}s.
\newblock In {\em Proceedings of the IEEE/CVF Conference on Computer Vision and
  Pattern Recognition}, pages 1532--1540, 2021.

\bibitem{shu2021encoding}
Manli Shu, Zuxuan Wu, Micah Goldblum, and Tom Goldstein.
\newblock Encoding robustness to image style via adversarial feature
  perturbations.
\newblock {\em Advances in Neural Information Processing Systems}, 34, 2021.

\bibitem{sugiyama2012density}
Masashi Sugiyama, Taiji Suzuki, and Takafumi Kanamori.
\newblock {\em Density ratio estimation in machine learning}.
\newblock Cambridge University Press, 2012.

\bibitem{sugiyama2008}
Masashi Sugiyama, Taiji Suzuki, Shinichi Nakajima, Hisashi Kashima, Paul von
  B\"{u}nau, and Motoaki Kawanabe.
\newblock Direct importance estimation for covariate shift adaptation.
\newblock {\em Annals of the Institute of Statistical Mathematics},
  60(4):699–746, 2008.

\bibitem{voorhees1999trec}
Ellen~M Voorhees et~al.
\newblock The trec-8 question answering track report.
\newblock In {\em Trec}, volume~99, pages 77--82, 1999.

\bibitem{voynov2020unsupervised}
Andrey Voynov and Artem Babenko.
\newblock Unsupervised discovery of interpretable directions in the gan latent
  space.
\newblock In {\em International conference on machine learning}, pages
  9786--9796. PMLR, 2020.

\bibitem{wei2021jacobian}
Yuxiang Wei, Yupeng Shi, Xiao Liu, Zhilong Ji, Yuan Gao, Zhongqin Wu, and
  Wangmeng Zuo.
\newblock Orthogonal jacobian regularization for unsupervised disentanglement
  in image generation.
\newblock In {\em Proceedings of the IEEE/CVF International Conference on
  Computer Vision}, pages 6721--6730, 2021.

\bibitem{wu2021stylealign}
Zongze Wu, Yotam Nitzan, Eli Shechtman, and Dani Lischinski.
\newblock Stylealign: Analysis and applications of aligned stylegan models.
\newblock {\em arXiv preprint arXiv:2110.11323}, 2021.

\bibitem{yan22c-fairness-auditing}
Tom Yan and Chicheng Zhang.
\newblock Active fairness auditing.
\newblock In {\em Proceedings of the 39th International Conference on Machine
  Learning}, volume 162 of {\em Proceedings of Machine Learning Research},
  pages 24929--24962. PMLR, 17--23 Jul 2022.

\bibitem{yuksel2021latentclr}
O{\u{g}}uz~Kaan Y{\"u}ksel, Enis Simsar, Ezgi~G{\"u}lperi Er, and Pinar
  Yanardag.
\newblock Latentclr: A contrastive learning approach for unsupervised discovery
  of interpretable directions.
\newblock In {\em Proceedings of the IEEE/CVF International Conference on
  Computer Vision}, pages 14263--14272, 2021.

\end{thebibliography}
}
\newpage
\appendix

\section{Ablation study}
First, we investigate the effect of the $\lambda_b$ parameter on KLIEP loss that allows us to discover novel attributes. In addition to KLIEP loss presented in the main text, we analyze a model trained with simple log loss used to train binary classifiers to predict the likelihood of a given sample. 

\subsection{Log Loss Model}
 This model, we denote as LOG, is nearly identical to the DRE models except instead of a softplus final activation, it uses a sigmoid function $\sigma(x) = \frac{1}{1 + e^{-x}}$. 
These models are used to classify whether a given feature belongs to $\mathcal{G}_c(z)$ or $\mathcal{G}_r(\bar{z})$.  We pre-train two separate LOG models to approximate $\hat{\gamma}_c(x) = \hat{p}_c(x)$, and $\hat{\gamma}_r(x) =  \hat{p}_r(x)$, where we treat $\mathcal{G}_c$ as $P(\mathrm{x} | Y = 1)$ and  $\mathcal{G}_r$ as $P(\mathrm{x} | Y = 0)$. These LOG models are learned using simple 2-layer MLPs, $f^c_{LOG}(~), f^r_{LOG}(~)$, such that 
\begin{equation}
\label{eq:log_models}
    \hat{\mathrm{\gamma}}_c(\mathrm{z}) = f^c_{\text{LOG}}(\mathcal{F}(\mathcal{G}_c(\mathrm{z}))) \mbox{ and }  \hat{\mathrm{\gamma}}_r(\bar{\mathrm{z}}) = f^r_{\text{LOG}}(\mathcal{F}(\mathcal{G}_r(\bar{\mathrm{z}}))),
\end{equation} where $\mathcal{F}$ is the same Encoder model used in main paper's equation 3.

The loss used for training the LOG models is defined as follows:
\begin{equation}
\mathcal{L}^c_{\text{Log}} = \frac{1}{T_{2}} \sum_{j=1}^{T_{2}} - \log (1 - \hat{\gamma}_c\left(\mathrm{\bar{z}}_j\right) )
                           + \frac{1}{T_{1}} \sum_{i=1}^{T_{1}} - \log (\hat{\gamma}_c\left(\mathrm{z  }_i)\right)
\end{equation}

where $\mathrm{\bar{z}}_j$ and $\mathrm{z}_i$ are random samples drawn from the latent space of each generator. The loss term for the second model LOG model is 
\begin{equation}
\mathcal{L}^r_{\text{Log}} = \frac{1}{T_{1}} \sum_{j=1}^{T_{1}} - \log (1 - \hat{\gamma}_r\left(\mathrm{\bar{z}}_j\right) )
                           + \frac{1}{T_{2}} \sum_{i=1}^{T_{2}} - \log (\hat{\gamma}_r\left(\mathrm{z  }_i)\right)
\end{equation}

The LOG models $f^1_{LOG}(~), f^2_{LOG}(~)$ are  trained to minimize $\mathcal{L}^1_{\text{Log}},\mathcal{L}^2_{\text{Log}}$ respectively. 

Finally, the trained LOG models are used to minimize the loss in equation 7 (rather than DRE models); the objective in equation 7 remains the same.

\begin{table}[!tb]
\centering
\small{
\begin{tabular}{ccc}
$\lambda$ & $\mathcal{R}_{\text{Score}}$ (DRE loss) ($\uparrow$) & $\mathcal{R}_{\text{Score}}$ (Log loss )($\uparrow$)\\ \hline
0                                          & $0.42  \pm{0.38}$           & $0.42 \pm{0.38}$     \\
0.1                                       & $\bm{0.61} \pm{0.35}$       & $0.37 \pm{0.41}$     \\
0.2                                       & $0.54 \pm{0.33}$             & $0.44 \pm{0.39}$     \\
0.5                                       & $0.57 \pm{0.40}$            & $0.45 \pm{0.38}$     \\
1                                         & $\bm{0.61} \pm{0.33}$       & $0.40 \pm{0.40}$     \\
5                                         & $0.57 \pm{0.39}$            & $0.34 \pm{0.32}$    \\                     \hline
\end{tabular}
\caption{ The effect on the unique direction score when modifying the regularization $\lambda$ on the average $\mathcal{R}_{\text{Score}}$ ($\pm{\text{ std}}$) for the the 7 CelebA pairwise leave-attribute-out experiments using a Robust ResNet-50 encoder.} 
\label{tbl:dre_lambda_abl}
}
\end{table}

\subsection{Missing attribute ablation study results} 

Table \ref{tbl:dre_lambda_abl} illustrates the missing attribute discovery score for each CelebA split versus full CelebA. With $\lambda=0$ (i.e. ignoring the DRE loss), the attribute discovery process has difficulty capturing some missing attributes. When using a regularization model trained with Log-loss, the results are consistently worse than DRE, sometimes even worse than with $\lambda=0$. The KLIEP loss model, on the other hand, performs consistently better for all lambda values $>0$.

\section{Same dataset, different architecture}
\begin{figure}[!tb]
    \centering
    \includegraphics[width=1.0\linewidth]{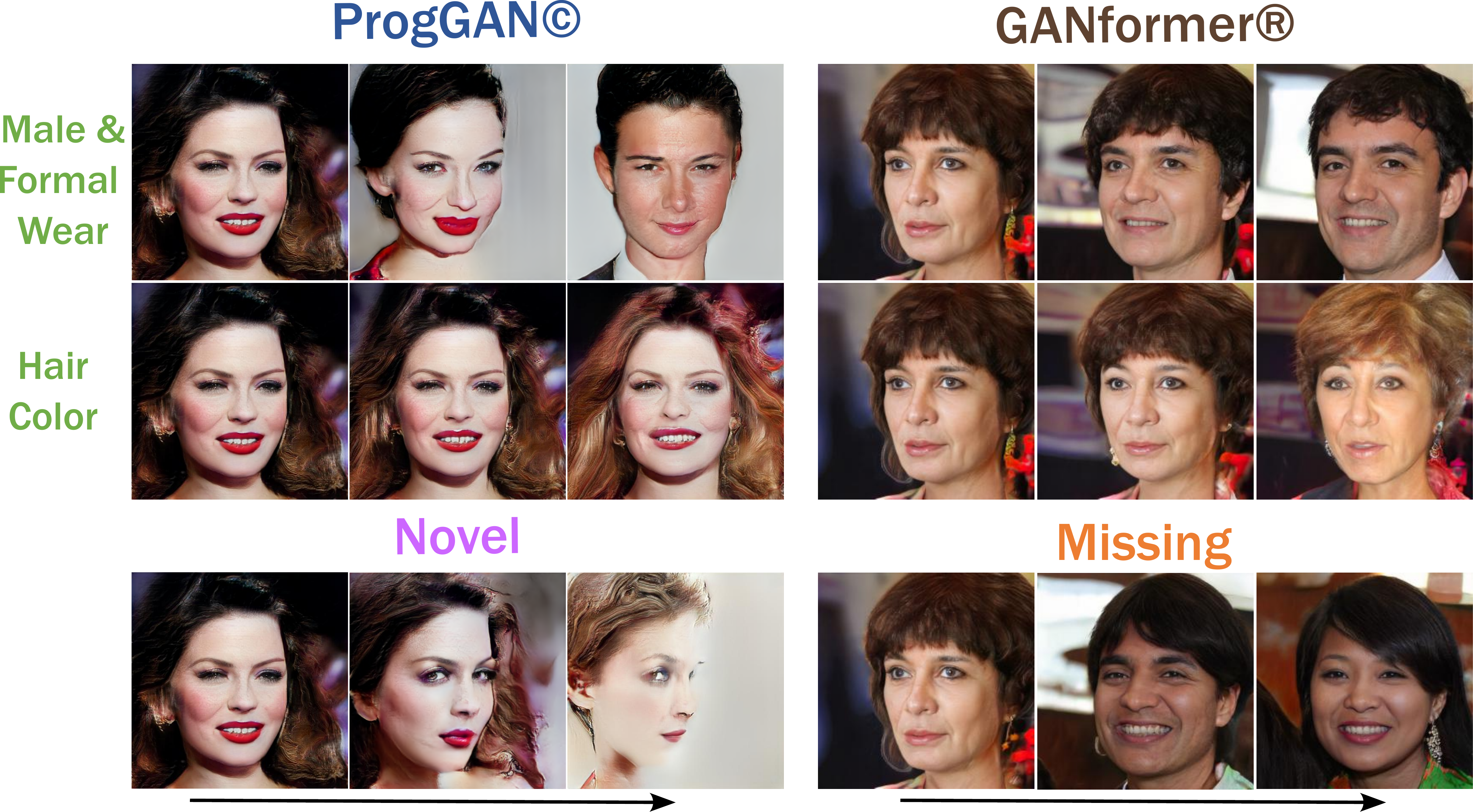}
    \caption{An example of applying our method to two generative models trained on the same dataset (FFHQ). We find ProgGAN and GANformer are able to find some alignment, and that the newer model (GANformer) is better at capturing the full data distribution of FFHQ (Missing) whereas ProgGAN is prone to generating non-realistic images (Novel).}
     \label{fig:proggan_v_gansformer}
\end{figure}

\begin{figure}[!tb]
    \centering
    \includegraphics[width=1.0\linewidth]{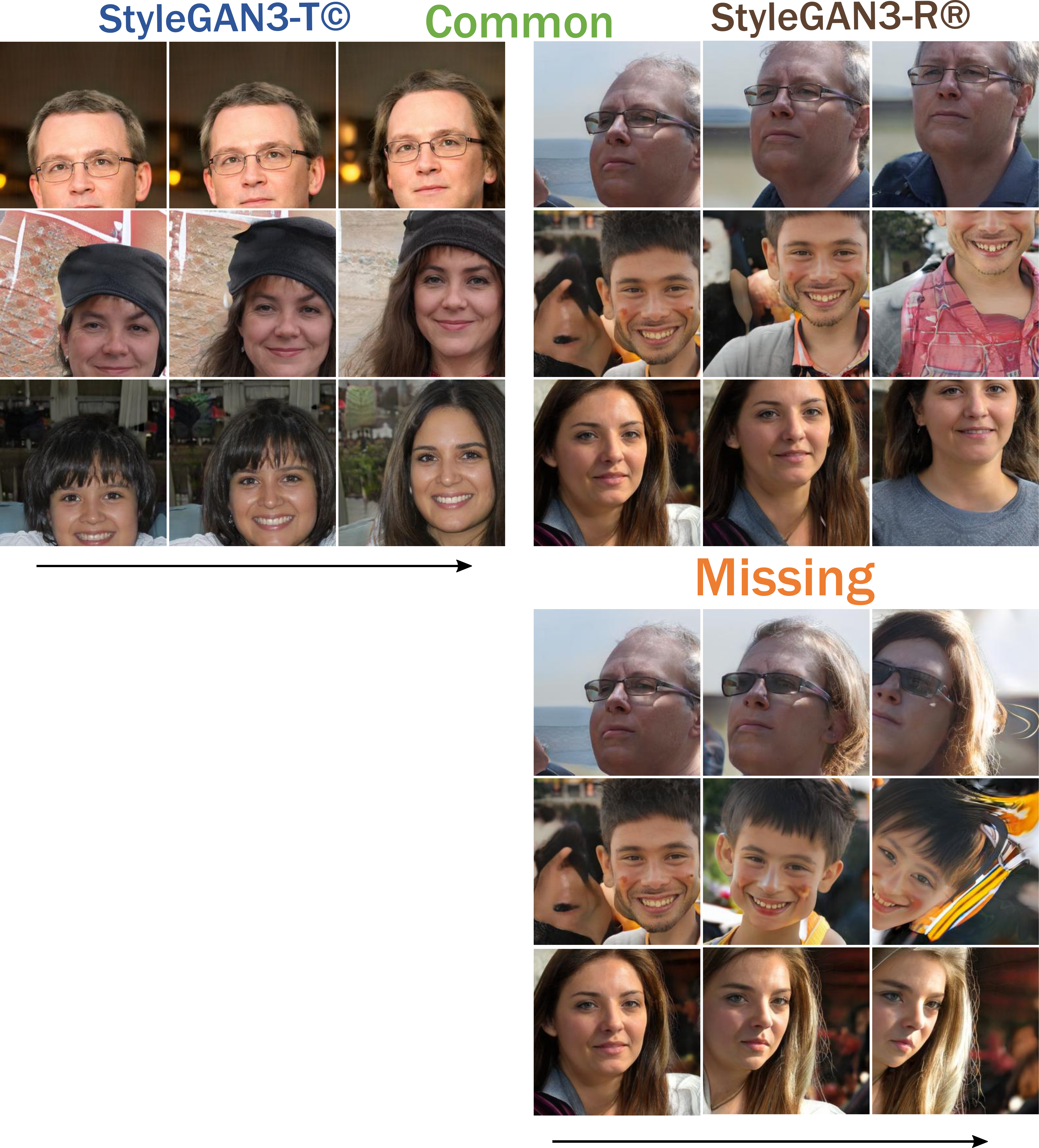}
    \caption{A few examples from our experiment applying xGA between two StyleGAN3 models, both trained on FFHQ, but with different model configurations. As expected both models having translation equivarience, and the rotation equivariance is missing from the translation model. }
     \label{fig:sg3}
\end{figure}

To verify the effectiveness of xGA at comparing models trained on the same dataset with different configurations, we perform two sets of experiments. We use Prog-GAN \cite{karras2017progressive} (client) and GANformer \cite{hudson2021ganformer2} (reference) trained on the FFHQ dataset. Figure \ref{fig:proggan_v_gansformer} shows an example of how these two GANs can be aligned, and how the novel/missing attribute reflects each GAN's capacity to learn the data distribution. 
We also use two configurations of a StyleGAN3 \cite{karras2021alias} trained on FFHQ. Figure \ref{fig:sg3} shows how translation equivarience is preserved in both models, whereas only the StyleGAN3-r model is rotationally equivarient.

\begin{figure}[!tb]
     \centering
     \includegraphics[width=0.99\linewidth]{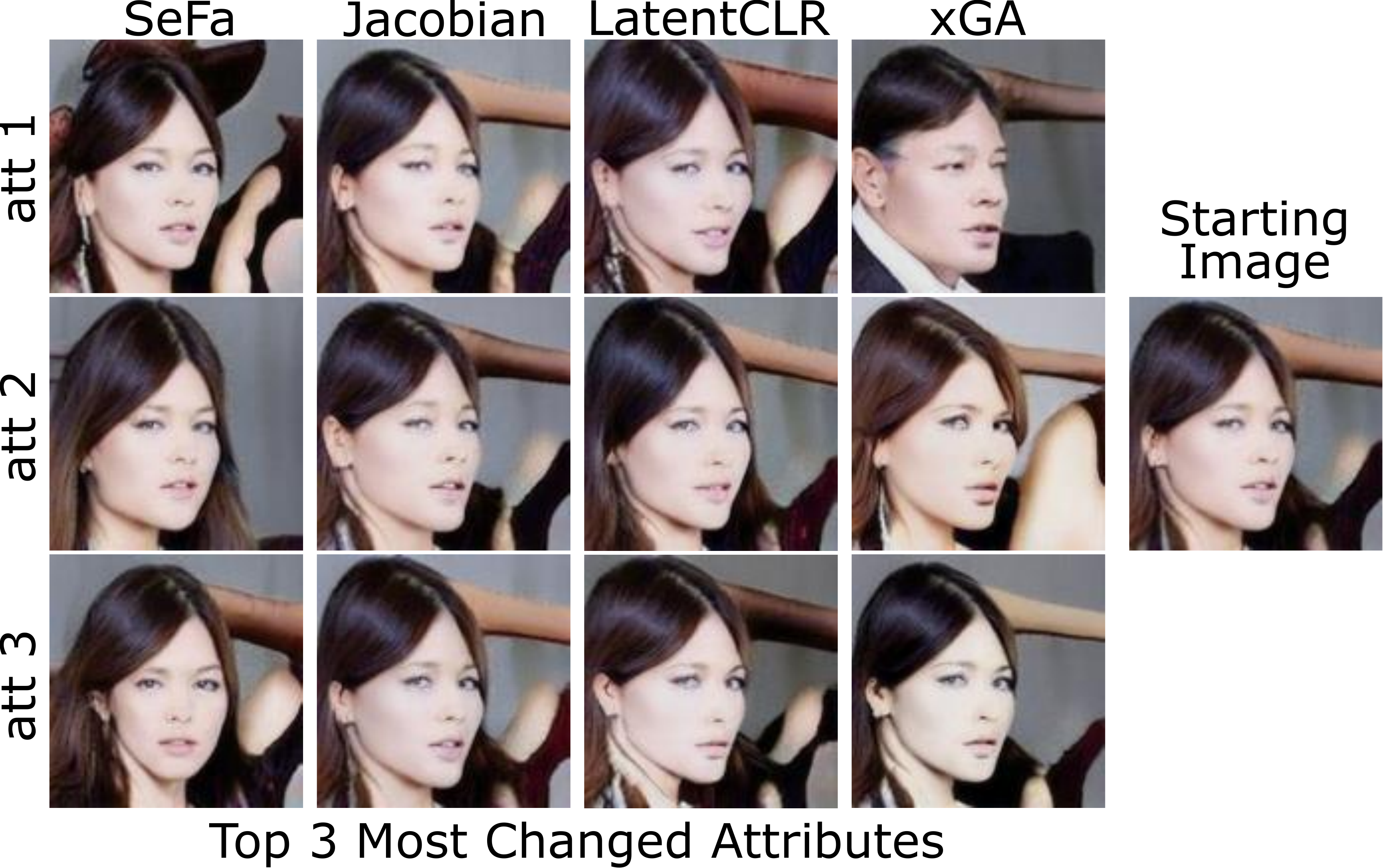}
     \caption{Comparing xGA on single GAN attribute discovery with existing approaches, we find that more diverse and novel attributes can be found simply by using an external feature space. We exploit this for effective alignment across two GAN models. Complete examples for all methods are provided below. 
     }
     \label{fig:1gan_examples}
 \end{figure}

\section{Single GAN results}
\begin{figure*}
\centering
         \includegraphics[width=0.89\linewidth]{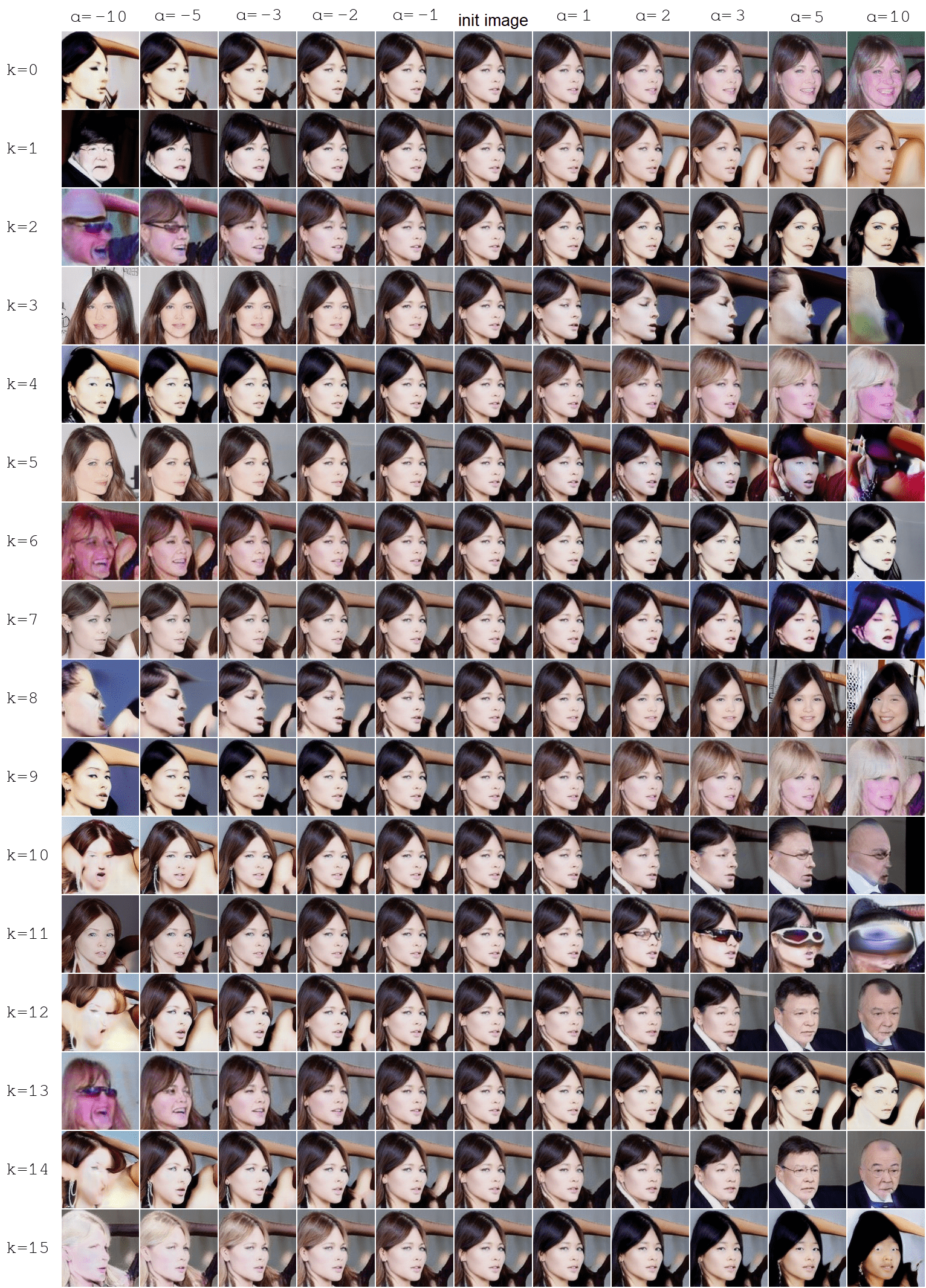}
         
        \caption{All 16 learned attributes of a Vanilla ResNet model.}
        \label{fig:supp-singlegan-resnet}
\end{figure*}

\begin{figure*}
\centering
         \includegraphics[width=0.89\linewidth]{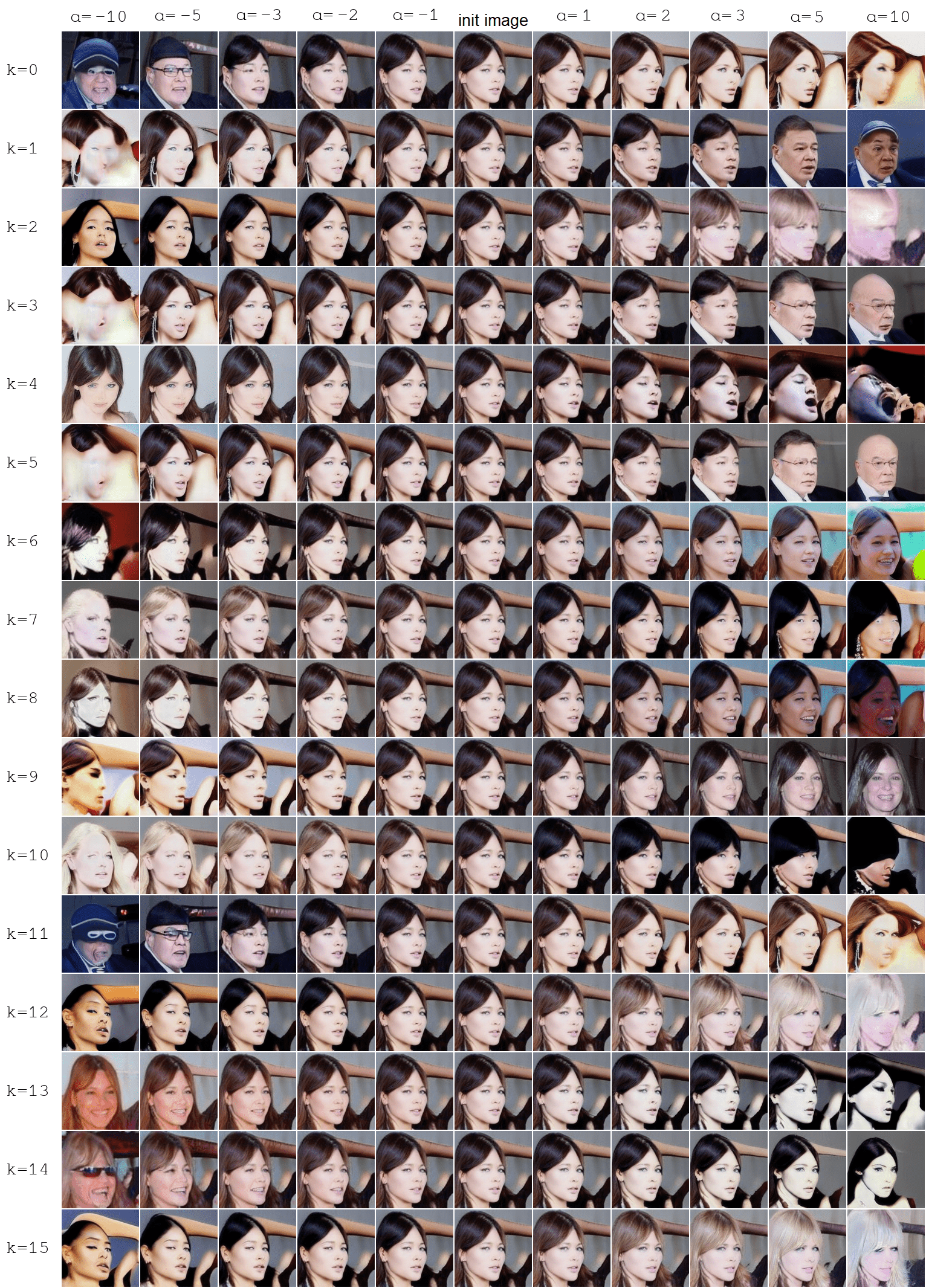}
         
        \caption{All 16 learned attributes of a robust ResNet model.}
        \label{fig:supp-singlegan-advbn}
\end{figure*}

\begin{figure*}
\centering
         \includegraphics[width=0.89\linewidth]{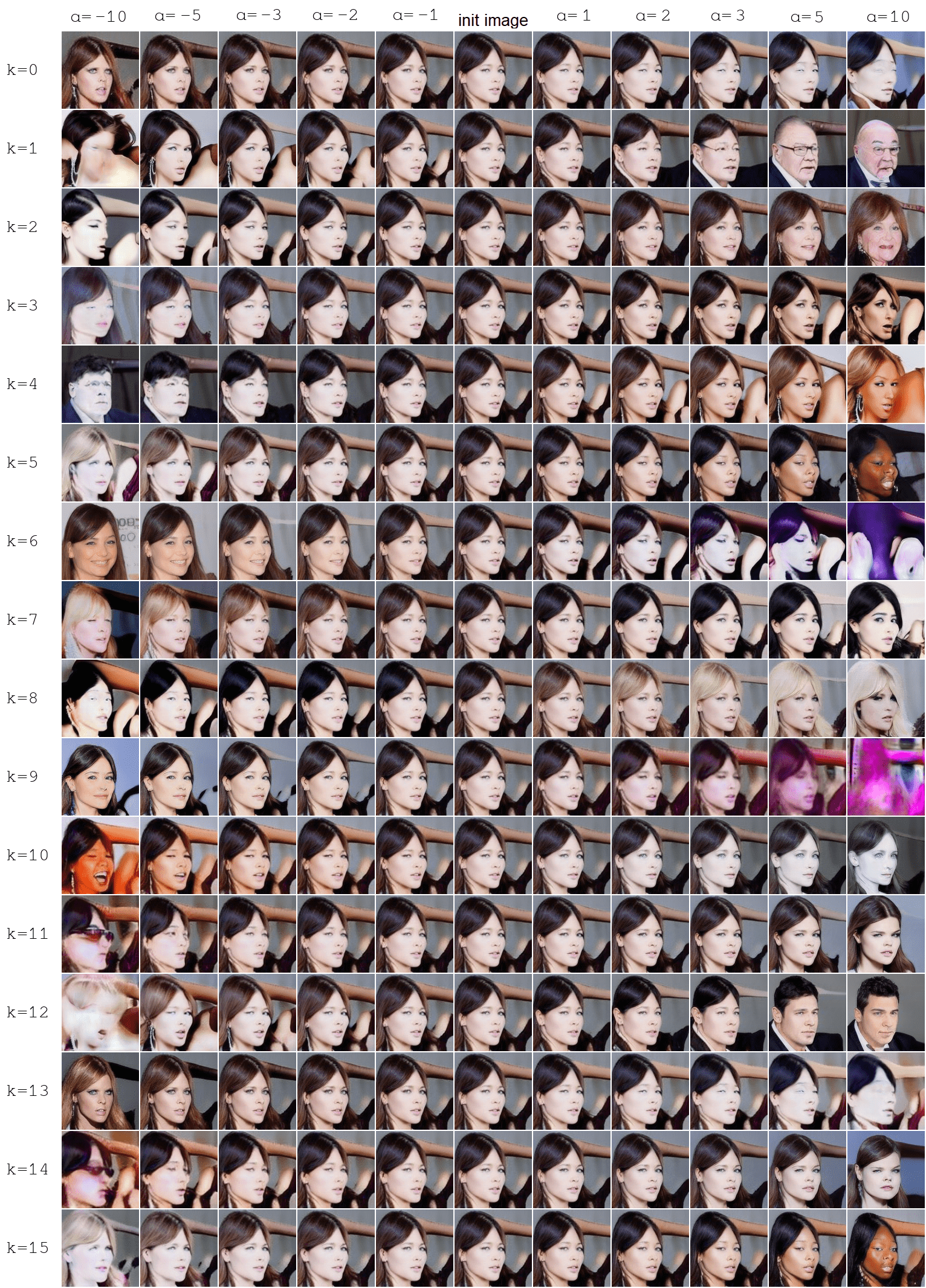}
         
        \caption{All 16 learned attributes of a clip ResNet model.}
        \label{fig:supp-singlegan-clip}
\end{figure*}

\begin{figure*}
\centering
         \includegraphics[width=0.89\linewidth]{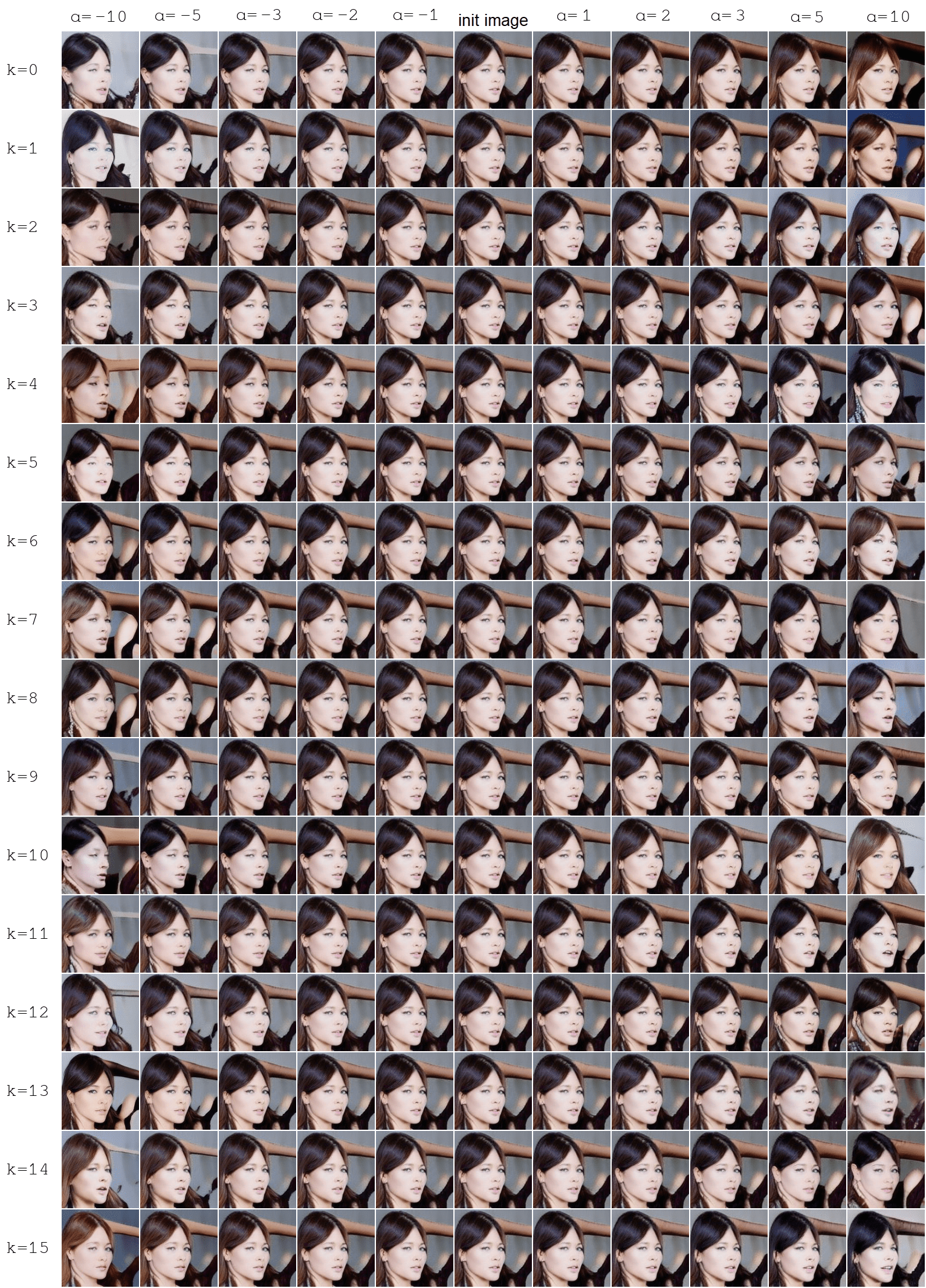}
         
        \caption{All 16 learned attributes of the Hessian method.}
        \label{fig:supp-singlegan-hessian}
\end{figure*}

\begin{figure*}
\centering
         \includegraphics[width=0.89\linewidth]{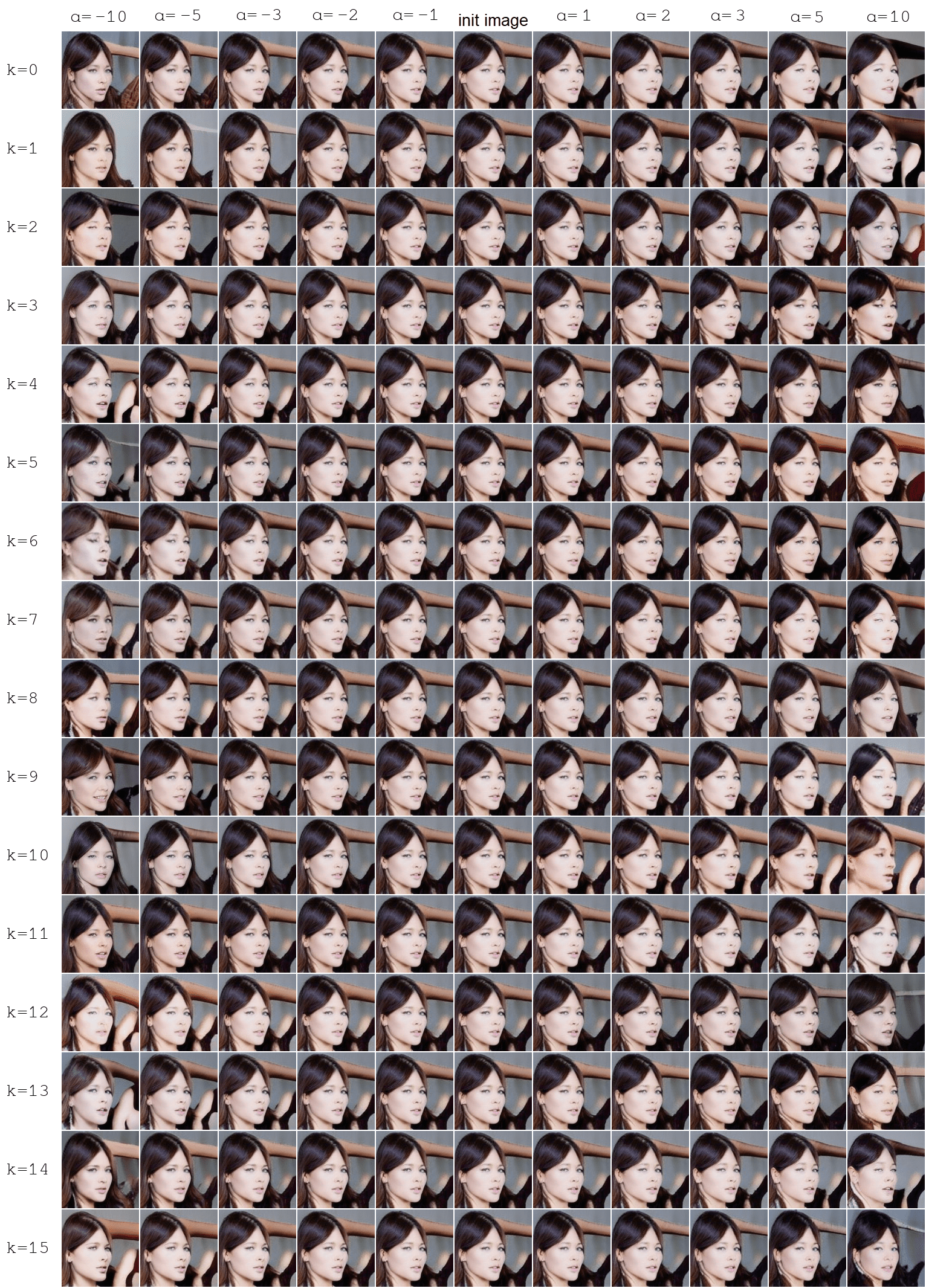}
         
        \caption{All 16 learned attributes of the Jacobian method.}
        \label{fig:supp-singlegan-jacobian}
\end{figure*}

\begin{figure*}
\centering
         \includegraphics[width=0.89\linewidth]{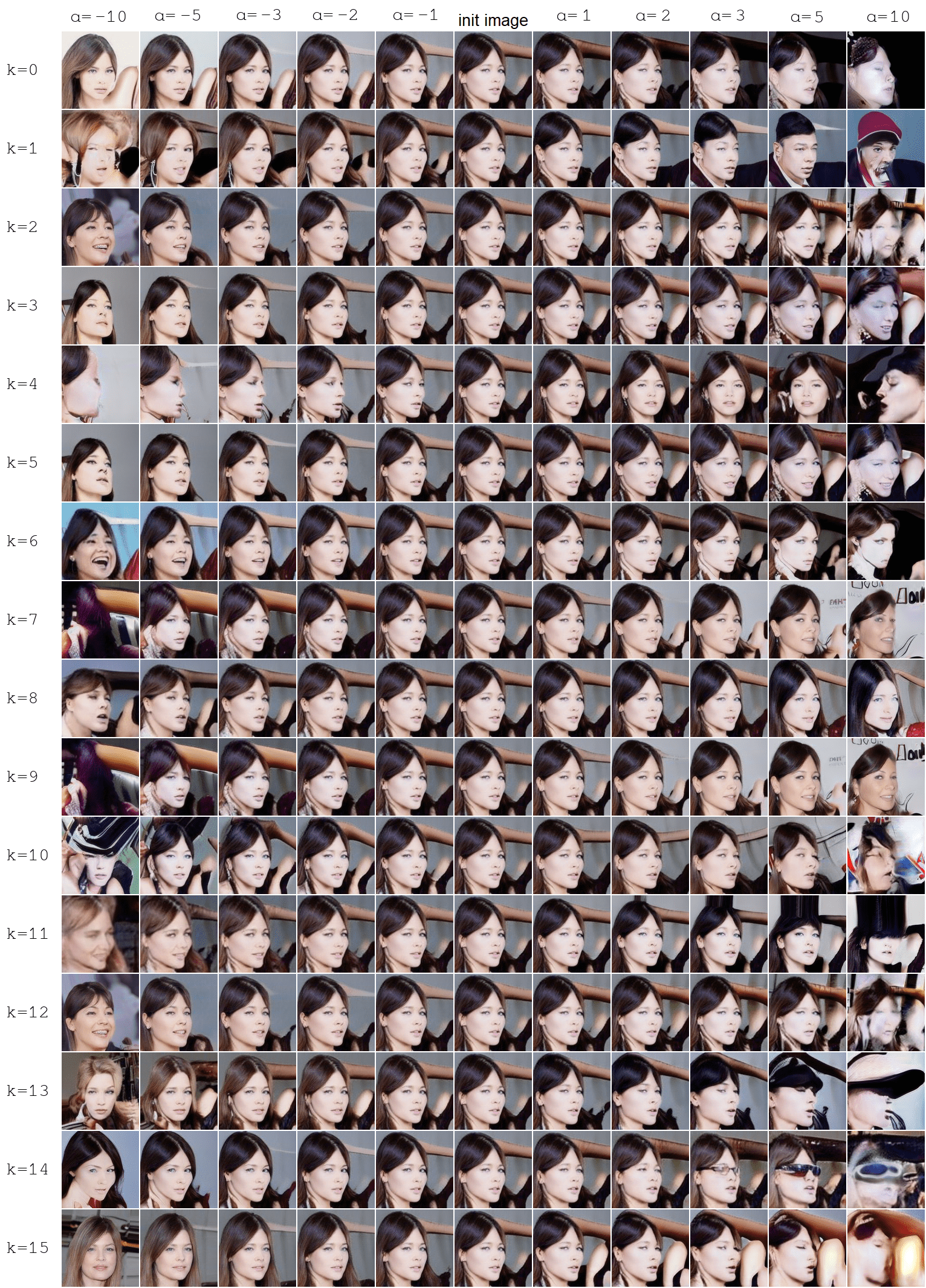}
         
        \caption{All 16 learned attributes of the original LatentCLR model with conditional directions.}
        \label{fig:supp-singlegan-latentclr}
\end{figure*}

\begin{figure*}
\centering
         \includegraphics[width=0.89\linewidth]{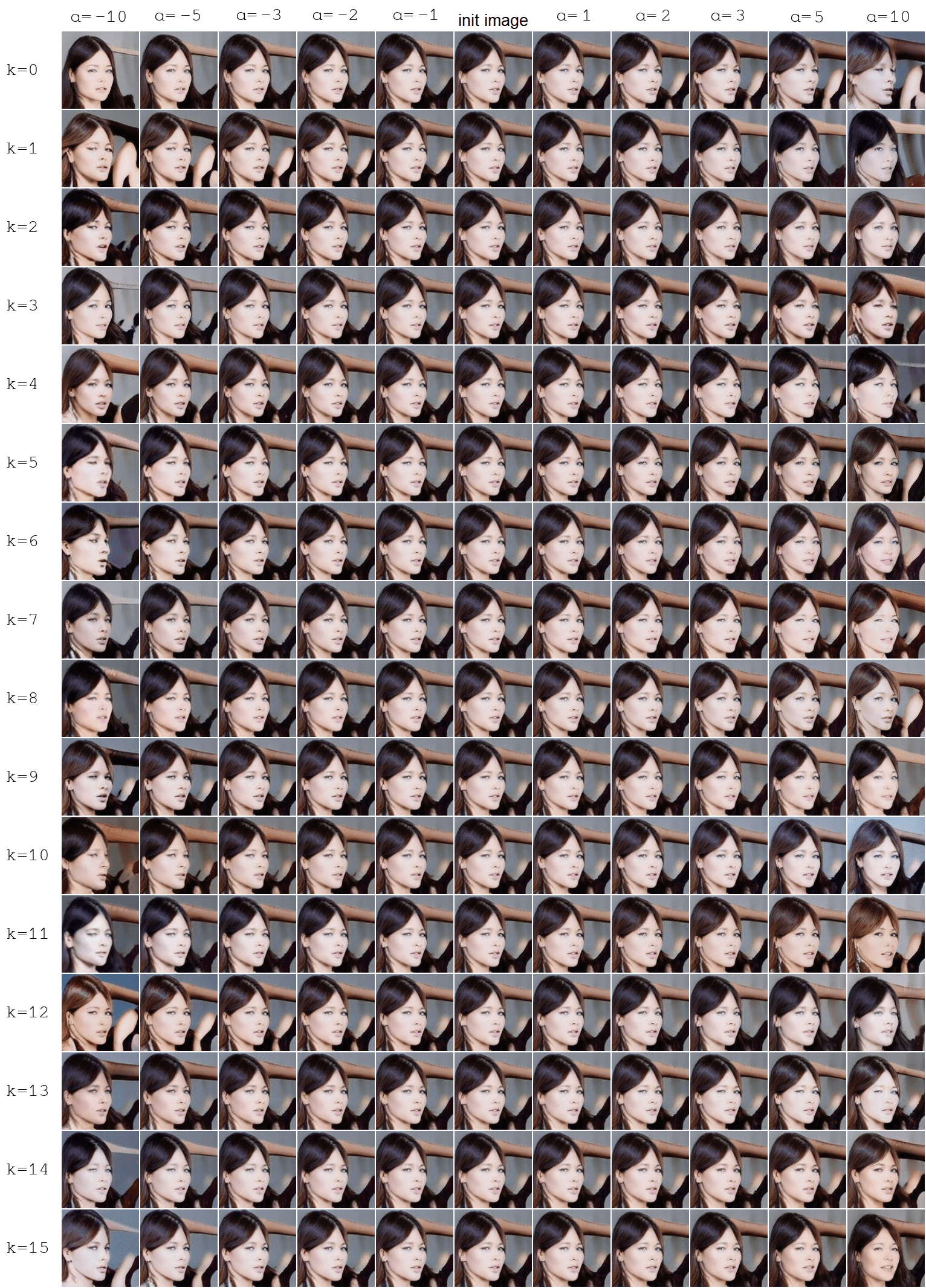}
         
        \caption{All 16 learned attributes of the Voynov method.}
        \label{fig:supp-singlegan-voynov}
\end{figure*}

\begin{figure*}
\centering
         \includegraphics[width=0.89\linewidth]{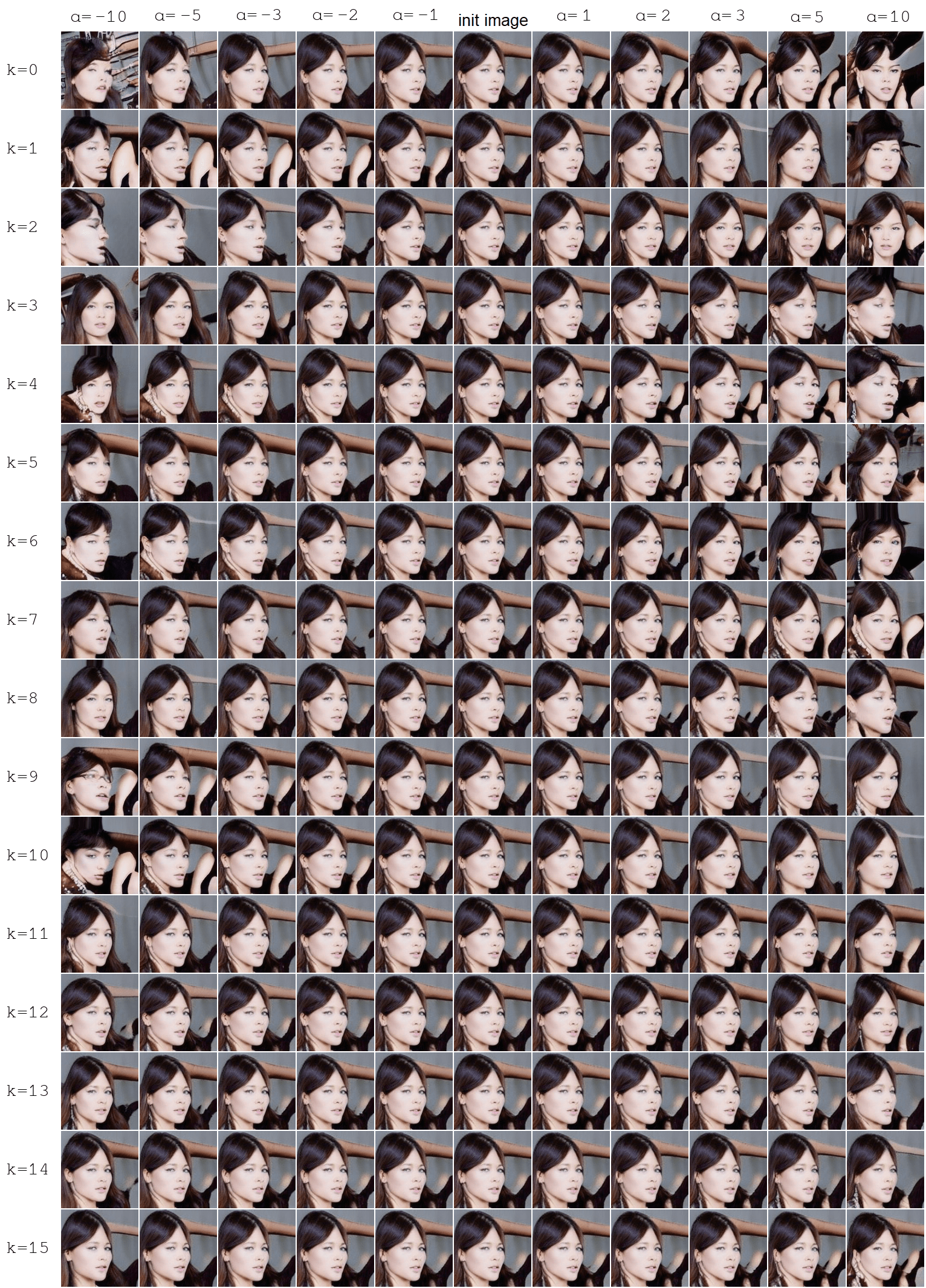}
         
        \caption{The top 16 learned attributes of the SeFa method.}
        \label{fig:supp-singlegan-sefa}
\end{figure*}

Here we present the full training of all learned directions for each of our methods using the same starting point from CelebA GAN.
Figure \ref{fig:1gan_examples} visualizes a shortened example of the top $3$ attributes (induce most changes in the ``oracle'' classifier predictions) and it is clear that xGA identifies the most diverse semantic changes. Complete results can be seen as follows:
\begin{enumerate}
    \item Sefa \cite{shen2021closed}: Figure \ref{fig:supp-singlegan-sefa}
    \item LatentCLR \cite{yuksel2021latentclr}: Figure \ref{fig:supp-singlegan-latentclr}
    \item Voynov \cite{voynov2020unsupervised}: Figure \ref{fig:supp-singlegan-voynov}
    \item Hessian \cite{peebles2020hessian}: Figure \ref{fig:supp-singlegan-hessian}
    \item Jacobian \cite{wei2021jacobian}: Figure \ref{fig:supp-singlegan-jacobian}
    \item xGA (ImageNet ResNet-50): figure \ref{fig:supp-singlegan-resnet}
    \item xGA (advBN ResNet-50): figure \ref{fig:supp-singlegan-advbn}
    \item xGA (CLIP ResNet-50): figure  \ref{fig:supp-singlegan-clip}
\end{enumerate}
We visualize both positive and negative directions for every model. Even though xGA and LatentCLR are not directly trained for negative directions, we find these attributes to be semantically meaningful and interesting.

\begin{figure*}
\centering
         \includegraphics[width=0.89\linewidth]{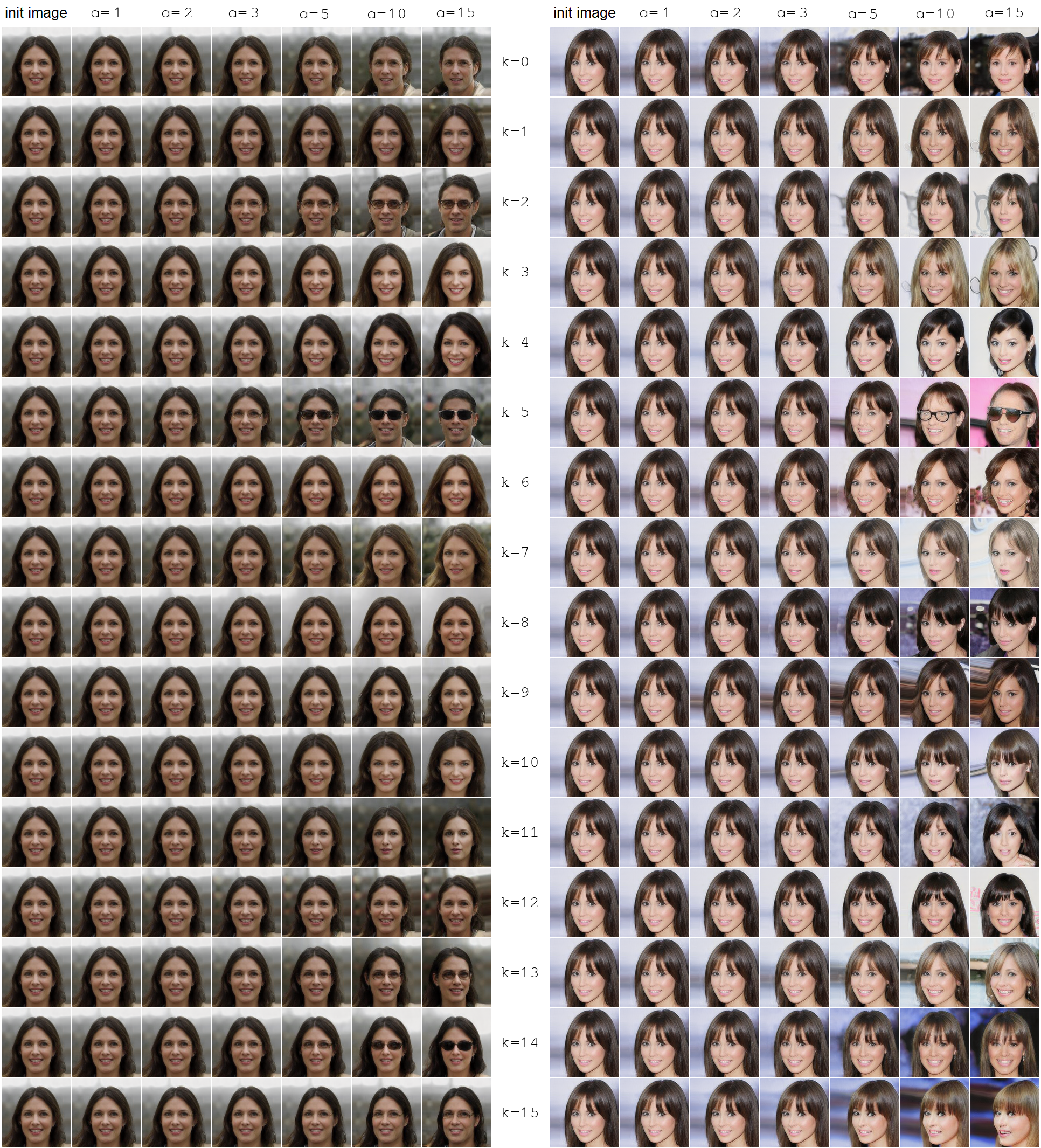}
        \caption{All 16 directions of two single GAN LatentCLR models trained on different GANs where the reference GAN is CelebA and client GAN is CelebA without Hats. The attributes are sorted by most similar ($k=0$) to least similar ($k=15$). While there are some major similarities (short hair $k=0$, eyeglasses $k=5$), the lack of a dedicated constraint for finding hats shows the flaws with this approach. This example is a clear demonstration that without a dedicated cross-model constraint (i.e., using the DRE models), finding missing attributes is difficult.}
        \label{fig:supp-2singleGANs-sorted}
\end{figure*}

Next we present an example where we compare two LatentCLR models trained on different GANs where the reference GAN is CelebA and client GAN is CelebA without Hats. We sort all the directions by most similar (as described in the main paper) and show an example of the results in Figure \ref{fig:supp-2singleGANs-sorted}, finding many similarities, but no dedicated Hat attribute in the reference GAN. Showing how without the dedicated constraint of the DRE models, finding missing attributes is difficult.

\section{Expanded Qualitative Results}
\begin{figure*}[]
\centering
         \includegraphics[width=0.66\linewidth]{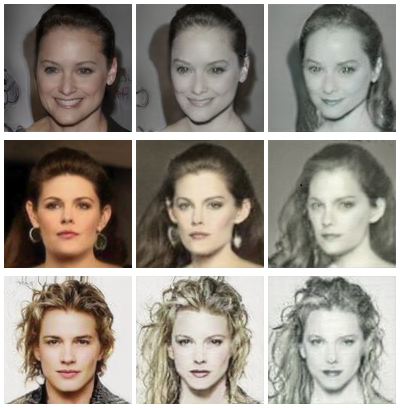}
         
         \includegraphics[width=0.66\linewidth]{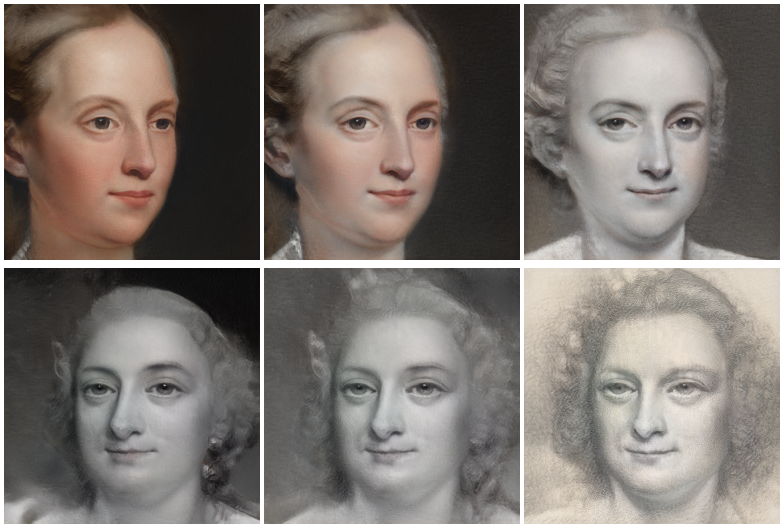}

        \caption{Examples of common sketch attribute between CelebA (Top) and Metfaces (Bottom).}
        \label{fig:supp-metface_sketch}
\end{figure*}

\begin{figure*}[]
\centering
         \includegraphics[width=0.66\linewidth]{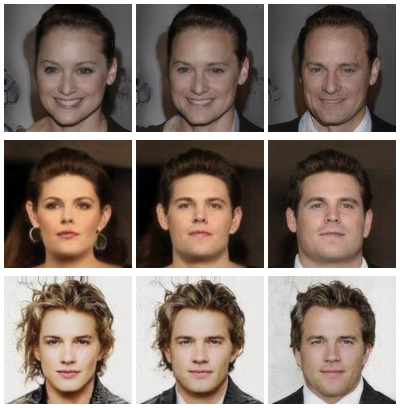}
         \includegraphics[width=0.66\linewidth]{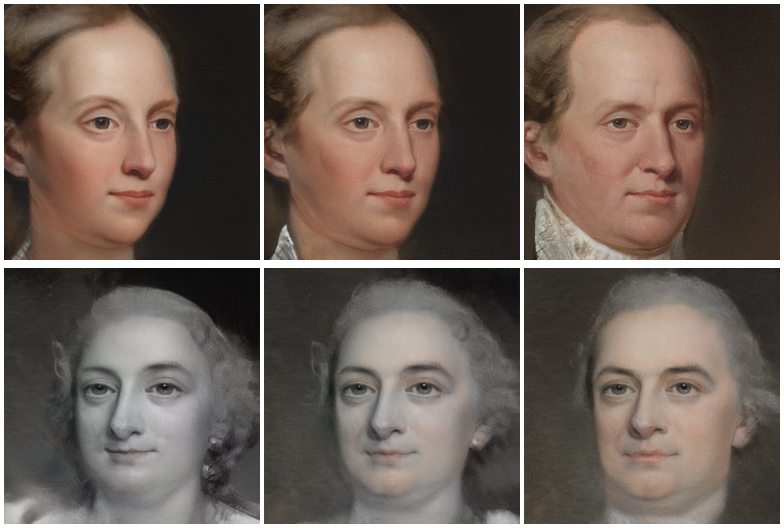}

        \caption{Examples of common formal-wear attribute between CelebA (Top) and Metfaces (Bottom). }
         \label{fig:supp-metface_formal}
\end{figure*}

\begin{figure*}[]
\centering
         \includegraphics[width=0.66\linewidth]{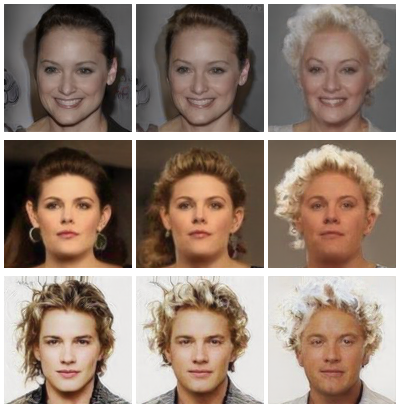}
         \includegraphics[width=0.66\linewidth]{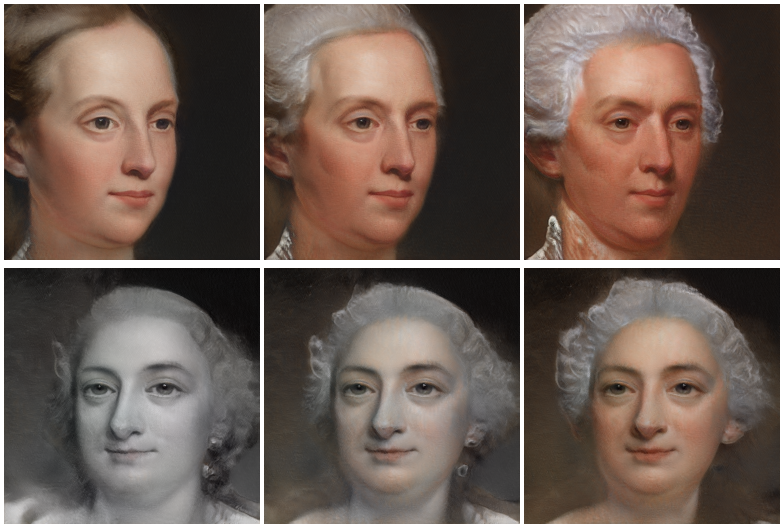}

        \caption{Examples of common  white, curly hair attribute between CelebA (Top) and Metfaces (Bottom).}
        \label{fig:supp-metface_curly}
\end{figure*}

\begin{figure*}[]
\centering
         \includegraphics[width=0.66\linewidth]{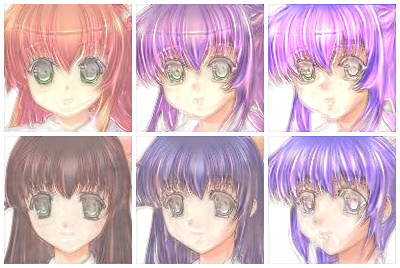}
         \includegraphics[width=0.66\linewidth]{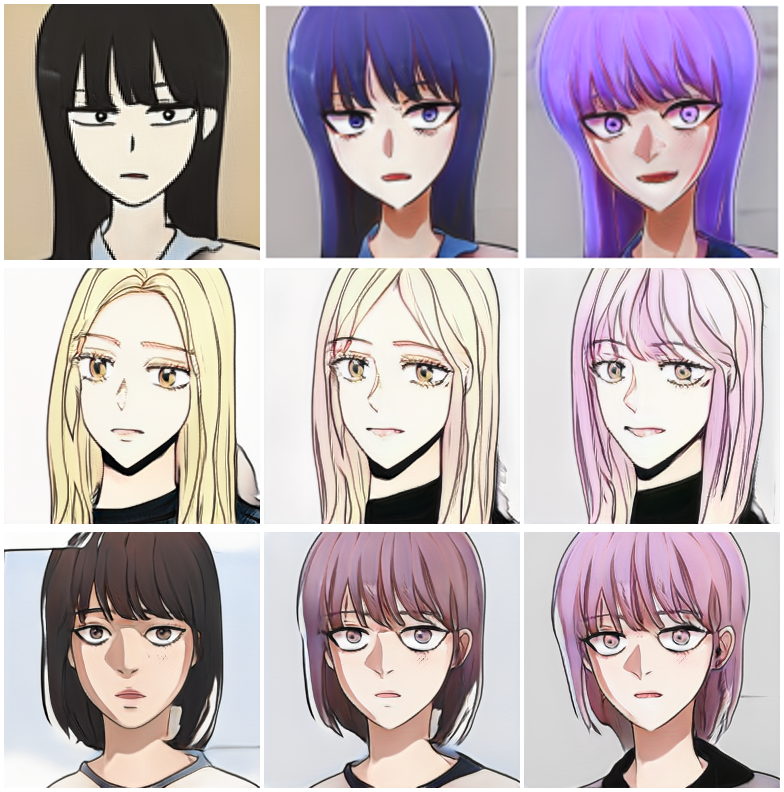}

        \caption{Examples of common purple hair attribute between Anime (Top) and Toon (Bottom).}
        \label{fig:supp-anime_purplehair}
\end{figure*}

\begin{figure*}[]
\centering
         \includegraphics[width=0.66\linewidth]{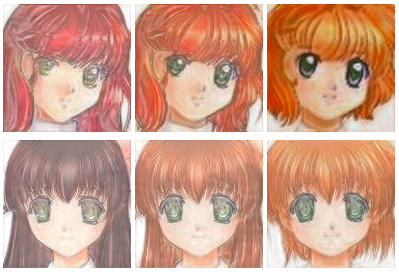}
         
         \includegraphics[width=0.66\linewidth]{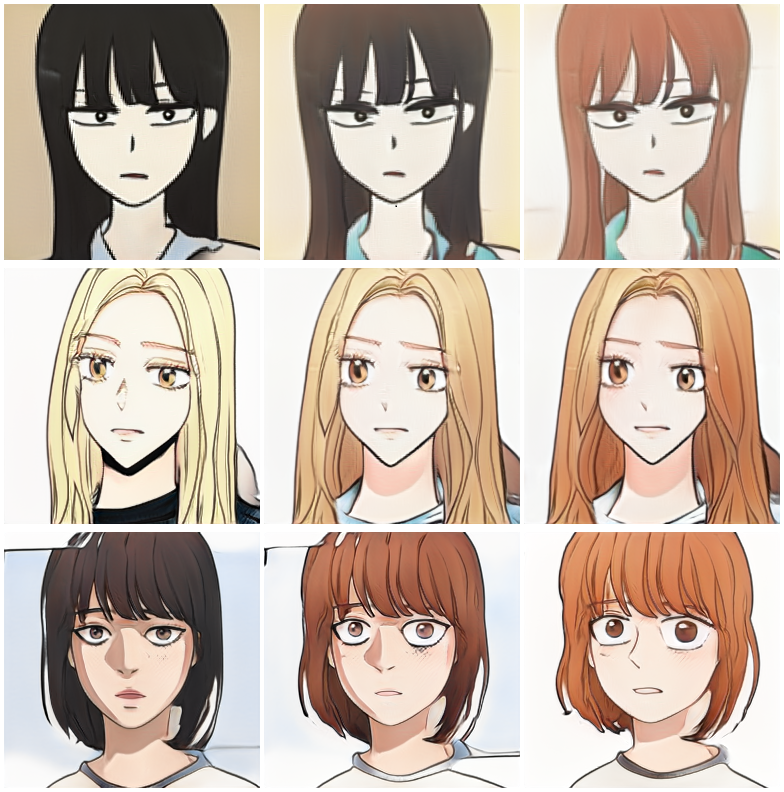}

        \caption{Examples of common orange/brown hair attribute between Anime (Top) and Toon (Bottom).}
        \label{fig:supp-anime_orangehair}
\end{figure*}

\begin{figure*}[]
\centering
         \includegraphics[width=0.66\linewidth]{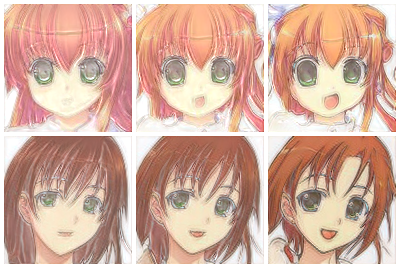}
         
         \includegraphics[width=0.66\linewidth]{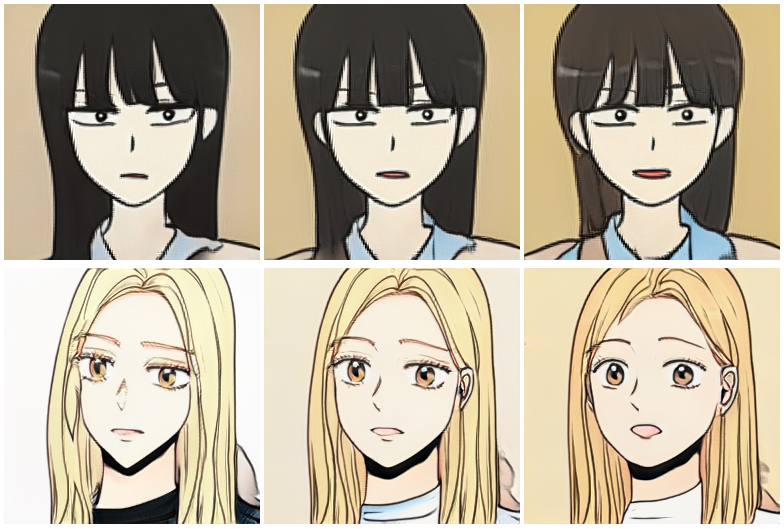}

        \caption{Examples of common open-mouth attribute between Anime (Top) and Toon (Bottom).}
        \label{fig:supp-anime_mouths}
\end{figure*}

\begin{figure*}[]
\centering
         \includegraphics[width=0.66\linewidth]{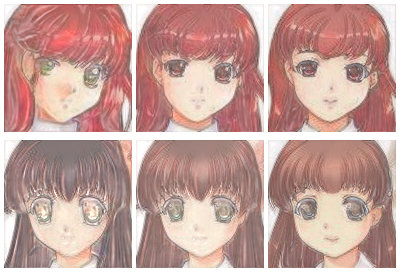}
         
         \includegraphics[width=0.66\linewidth]{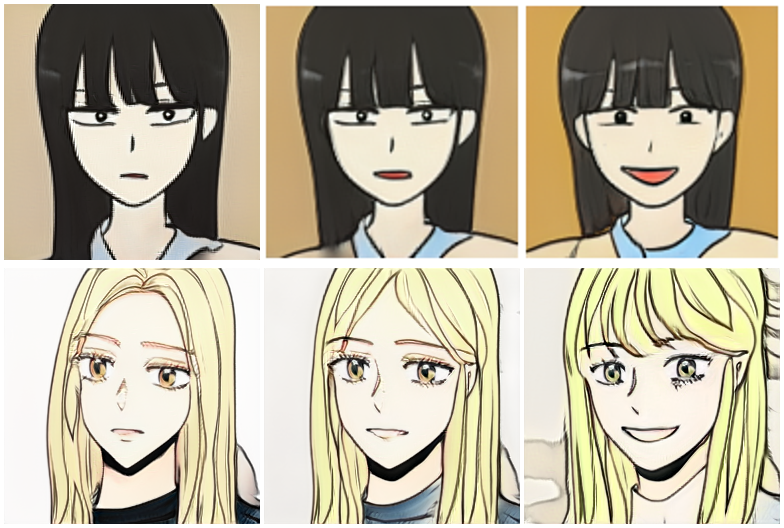}

        \caption{Examples of common smiling attribute between Anime (Top) and Toon (Bottom).}
        \label{fig:supp-anime_smiles}
\end{figure*}

\begin{figure*}[]
\centering
         \includegraphics[width=0.66\linewidth]{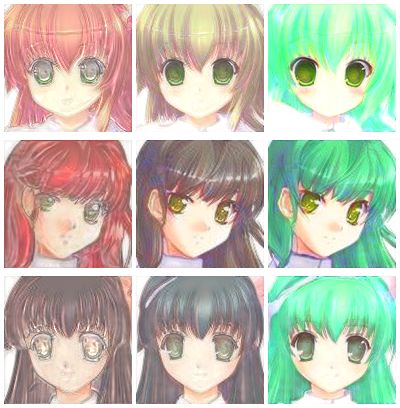}

         \includegraphics[width=0.66\linewidth]{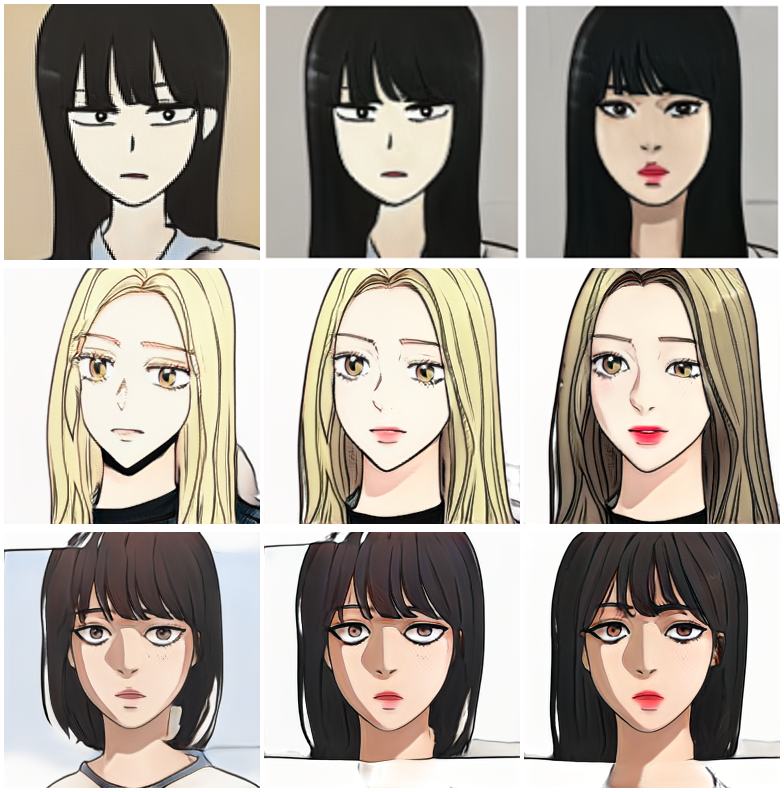}

        \caption{Examples of novel green hair attribute from Anime (Top) and the missing lipstick attribute from Toon (Bottom).}
        \label{fig:supp-anime_uniques}
\end{figure*}

\begin{figure*}[]
\centering
         \includegraphics[width=0.66\linewidth]{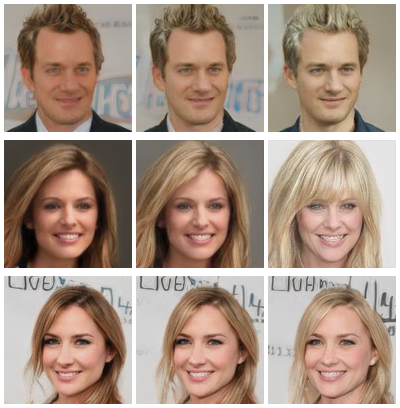}

         \includegraphics[width=0.66\linewidth]{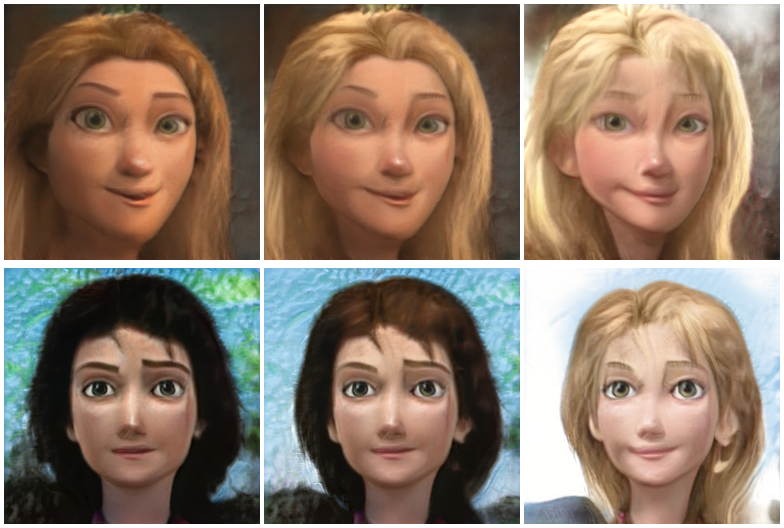}

        \caption{Examples of common blonde attribute between CelebA (Top) and Disney (Bottom). }
         \label{fig:supp-disney_blonde}
\end{figure*}

\begin{figure*}[]
\centering
         \includegraphics[width=0.66\linewidth]{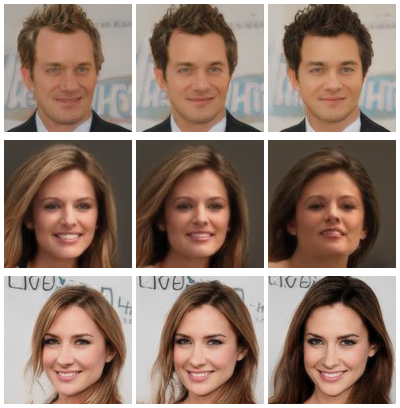}

         \includegraphics[width=0.66\linewidth]{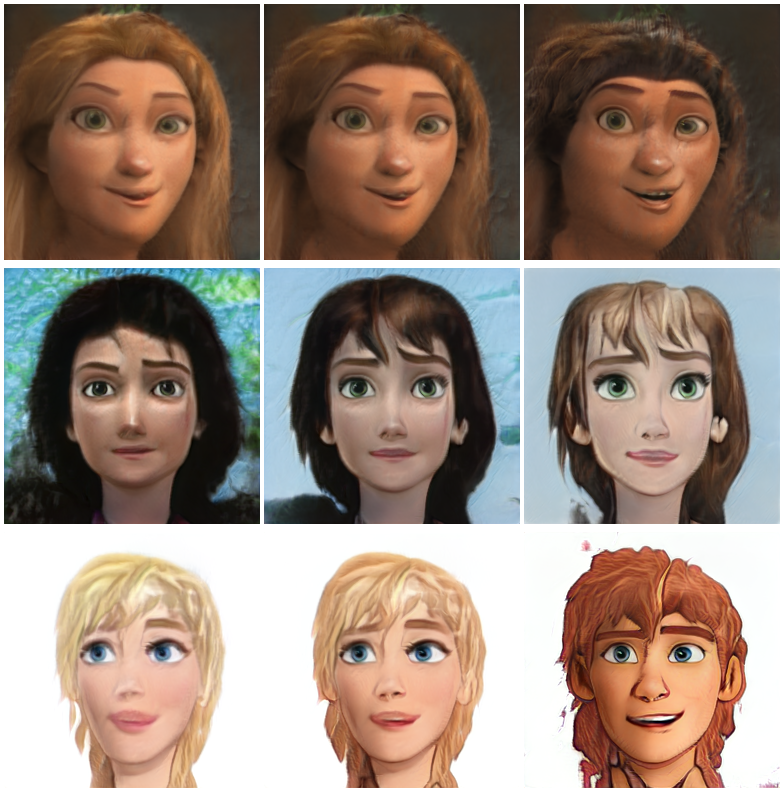}

        \caption{Examples of common brown hair attribute between CelebA (Top) and Disney (Bottom). }
         \label{fig:supp-disney_brown}
\end{figure*}

\begin{figure*}[]
\centering
         \includegraphics[width=0.66\linewidth]{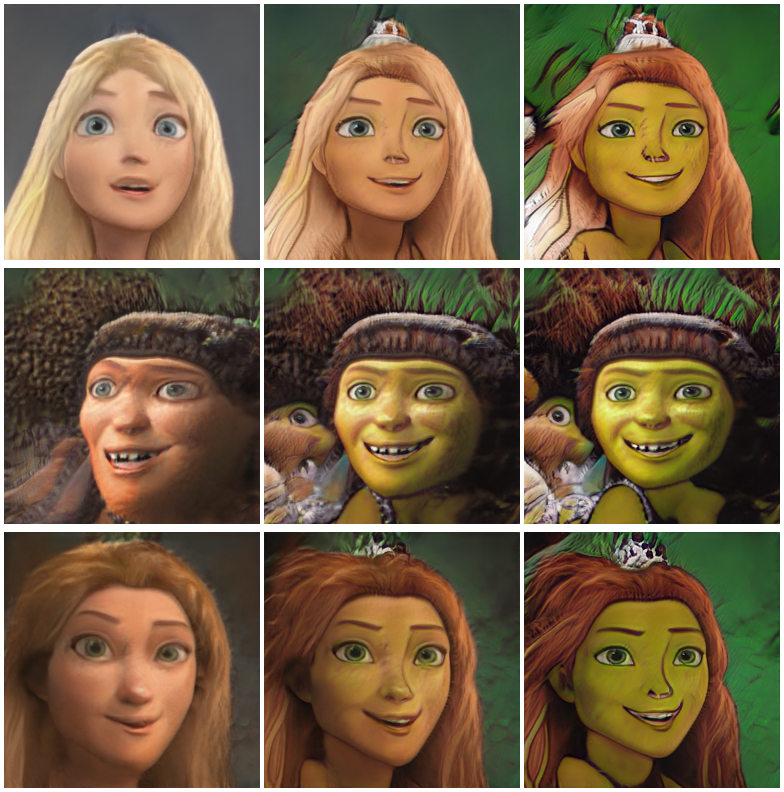}

         \includegraphics[width=0.66\linewidth]{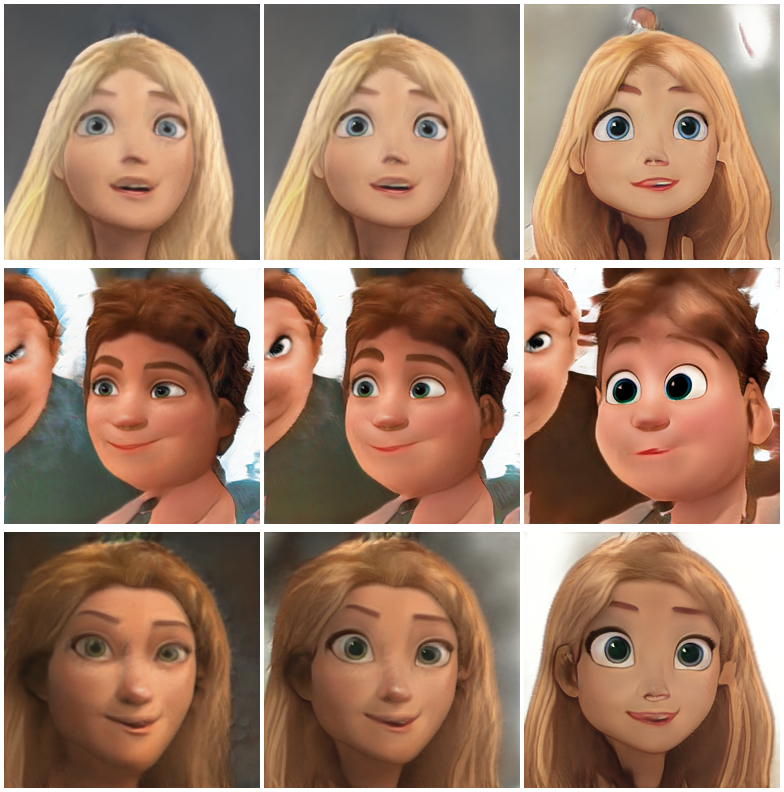}

        \caption{Two examples of novel Disney attributes: making princesses ogre-like and large cartoonish eyes.}
         \label{fig:supp-disney_uniques}
\end{figure*}

\begin{figure*}[]
\centering
         \includegraphics[width=0.33\linewidth]{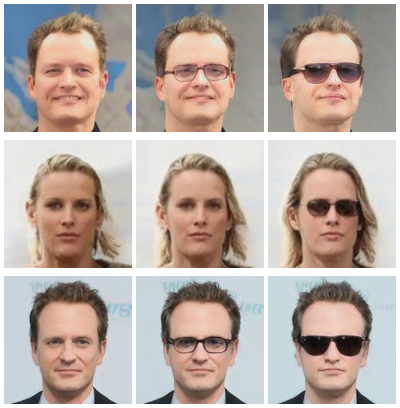}
         \vspace{1mm}
         \includegraphics[width=0.33\linewidth]{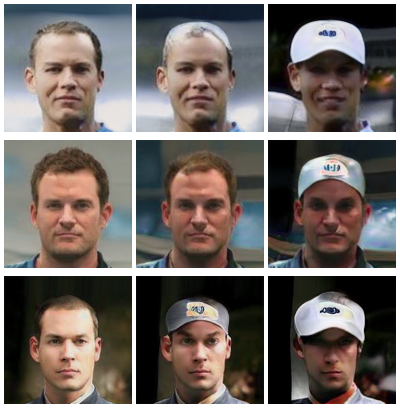}
         \vspace{1mm}
         \includegraphics[width=0.33\linewidth]{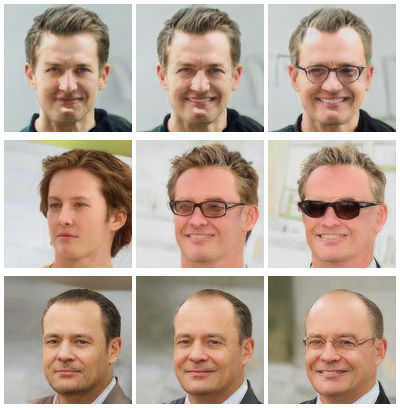}
        \caption{Examples of various missing attributes from full CelebA GAN against three different attribute splits: \textbf{(Left 3)} the missing eyeglass attribute, \textbf{(Middle 3)} the missing hats attribute and \textbf{(Right 3)} the missing eyeglass, smiling, and attempting to identify the necktie attribute.}
        \label{fig:supp-celeba_uniques}
\end{figure*}

Here we present many additional examples of shared directions between two GANs,
and novel/missing directions from a few different GAN pairs that contain subset of the CelebA dataset. 
We introduce a new GAN (anime), as it produces interesting common, missing, and novel attributes, though the GAN itself produces lower quality images than other models, and as such we leave it here in the supplement.
The figures are arranged as follows:
\begin{enumerate}
    \item Common attributes: CelebA (reference) and Metface (client) sketch (\ref{fig:supp-metface_sketch}), formal (\ref{fig:supp-metface_formal}), and curly hair  (\ref{fig:supp-metface_curly})
    \item Common attributes: Anime (client) and Toon (reference) purple hair (\ref{fig:supp-anime_purplehair}), orange/brown hair (\ref{fig:supp-anime_orangehair}), open mouths (\ref{fig:supp-anime_mouths}), and smiling (\ref{fig:supp-anime_smiles}); missing attributes of green hair / lipstick (\ref{fig:supp-anime_uniques})
    \item Common attributes: CelebA (reference) and Disney (client) blonde hair (\ref{fig:supp-disney_blonde}), and brown hair (\ref{fig:supp-disney_brown}); novel Disney attributes of turning green / cartoonish eyes (\ref{fig:supp-disney_uniques})
    \item Additional missing attributes from different CelebA client GANs, with CelebA reference GAN (\ref{fig:supp-celeba_uniques})
\end{enumerate}

\section{Expanded Quantitative Results}

First we present the results for using ViT-based feature extractors in table \ref{tbl:supp-vit-experiments}. We include 3 different pretrained models: one original trained on ImageNet, CLIP, and MAE. While ViT does well for entropy metric, it performs poorly for cross model based experiments.

Next, we present the entire results for our missing attribute quantitative experiments. To recap these experiments, we use the $7$ controlled CelebA models which are missing one or more attributes (hat, glasses, male, female, beard, beards|hats, and smiles|glasses|ties) and treat them as the client model; we audit these models with respect to the reference CelebA GAN. The $7$ missing attribute experiments are shown in table \ref{tbl:supp-full-recovery}, where we can see xGA performs well (e.g., easily finding the missing glasses attribute). The $7$ attribute alignment experiments are shown in table \ref{tbl:supp-full-alignment}, where again we see xGA with a robust resnet performs well, especially when the client GAN is missing multiple attributes (e.g., client GAN is missing beards and hats).

For completion's sake, we run pairwise experiments between each GAN, treating each GAN as reference versus the other $7$ GANs, which results in a total of $56$ client/reference paired experiments. We report these comprehensive results in the following tables (where rows are reference GAN and columns are the client): \ref{tbl:supp-unique-voynov}, \ref{tbl:supp-unique-latentclr}, \ref{tbl:supp-unique-hessian}, \ref{tbl:supp-unique-jacobian}, \ref{tbl:supp-unique_vanilla} , \ref{tbl:supp-unique_att} ,  \ref{tbl:supp-unique_robust} ,  \ref{tbl:supp-unique_rnclip} ,  \ref{tbl:supp-unique_vit} ,  \ref{tbl:supp-unique_clipvit} , and  \ref{tbl:supp-unique_mae}. 
We also compute the the common attribute results experiments in the following tables: \ref{tbl:supp-cosine-voynov},\ref{tbl:supp-cosine-latentclr},\ref{tbl:supp-cosine-hessian},\ref{tbl:supp-cosine-jacobian},  \ref{tbl:supp-cosine-vanilla}, \ref{tbl:supp-cosine-att}, \ref{tbl:supp-cosine-robust}, \ref{tbl:supp-cosine-clipRN}, \ref{tbl:supp-cosine-vit}, \ref{tbl:supp-cosine-clipvit}, and \ref{tbl:supp-cosine-mae}.

\begin{table*}[!tb]\centering
\begin{tabular}{llllllll}
                       & Female & Male  & No Hats & No Glasses & No Beards & \begin{tabular}[c]{@{}l@{}}No Beard\\ No Hats\end{tabular} & \begin{tabular}[c]{@{}l@{}}No Glasses\\ No Smiles\\ No Ties\end{tabular} \\ \hline
SeFa                    & 0.143  & 0.143 & 0.045   & 0.111      & 0.063     & 0.278       & 0.189  \\
jacobian                & 0.478  & 0.536 & 0.086   & 0.390      & 0.388     & 0.120       & 0.287  \\
Hessian                 & 1.000  & 1.000 & 0.056   & 0.167      & 0.167     & 0.096       & 0.407  \\
LatentCLR               & 1.000  & 1.000 & 0.250   & 0.333      & 0.200     & 0.153       & 0.537  \\
Voynov                  & 0.500  & 1.000 & 0.333   & 0.050      & 0.500     & 0.153       & 0.259  \\
xGA (ResNet-50)         & 1.000  & 1.000 & 0.333   & 0.500      & 0.200     & 0.167       & 0.465  \\
xGA (Clip ResNet-50)    & 1.000  & 1.000 & 1.000   & 0.200      & 0.200     & 0.108       & 0.383  \\
xGA (advBN ResNet-50)   & 1.000  & 1.000 & 0.250   & 1.000      & 0.063     & 0.183       & 0.401  \\ \hline
\end{tabular}
\caption{The full results for the recovery scores ($\mathcal{R}_{\text{score}}$), where CelebA GAN is the reference.}
 \label{tbl:supp-full-recovery}
\end{table*}

\begin{table*}[!tb] \centering
\begin{tabular}{llllllll}
              & Female & Male  & No Hats & No Glasses & No Beards & \begin{tabular}[c]{@{}l@{}}No Beard\\ No Hats\end{tabular} & \begin{tabular}[c]{@{}l@{}}No Glasses\\ No Smiles\\ No Ties\end{tabular} \\ \hline
SeFa                    & 0.413  & 0.458 & 0.372   & 0.374      & 0.314     & 0.387    & 0.355  \\
Hessian                 & 0.475  & 0.489 & 0.652   & 0.598      & 0.618     & 0.615    & 0.525  \\
LatentCLR               & 0.519  & 0.511 & 0.556   & 0.533      & 0.512     & 0.593    & 0.579  \\
Voynov                  & 0.566  & 0.477 & 0.567   & 0.555      & 0.570     & 0.562    & 0.513  \\
Jacobian                & 0.523  & 0.452 & 0.505   & 0.528      & 0.519     & 0.491    & 0.495  \\
xGA (ResNet-50)         & 0.457  & 0.403 & 0.740   & 0.792      & 0.461     & 0.643    & 0.489  \\
xGA (Clip ResNet-50)    & 0.753  & 0.451 & 0.772   & 0.791      & 0.894     & 0.580    & 0.656  \\
xGA (advBN ResNet-50)   & 0.615  & 0.357 & 0.750   & 0.825      & 0.619     & 0.803    & 0.649  \\ \hline
\end{tabular}
\caption{The full results for the alignment scores ($\mathcal{A}_{\text{score}}$), where CelebA GAN is the reference}
 \label{tbl:supp-full-alignment}
\end{table*}

\begin{table*}[!tb]
\centering
\begin{tabular}{llll}
Method / Model      &   $\mathcal{H}_{\text{score}}$ ($\downarrow$)    &  $\mathcal{A}_{\text{score}}$ ($\uparrow$) & $\mathcal{R}_{\text{score}}$ ($\uparrow$) \\ \hline
xGA + ViT         &   $1.988 \pm{0.068}$    & $0.377 \pm{0.090} $    & $0.249 \pm{0.217}$          \\
xGA + ViT + MAE   &   $2.102 \pm{0.035}$    & $0.349 \pm{0.089}$          & $0.194 \pm{0.197}$         \\ 
xGA + ViT + Clip  &   $2.091 \pm{0.041}$    & $0.397 \pm{0.122}$          & $0.268 \pm{0.195}$          \\ 
\hline
\end{tabular}
\caption{ ViT-based extractors results. The average entropy scores for all 8 CelebA experiments, the average alignment scores ($\mathcal{A}_{\text{score}}$) for the CelebA pairwise experiments, and the average recovery scores ($\mathcal{R}_{\text{score}}$) for the CelebA pairwise leave-attribute-out experiments ($\pm{\text{ std}}$)} 
\label{tbl:supp-vit-experiments}
\end{table*}

\begin{figure*}
\centering
         \includegraphics[width=0.89\linewidth]{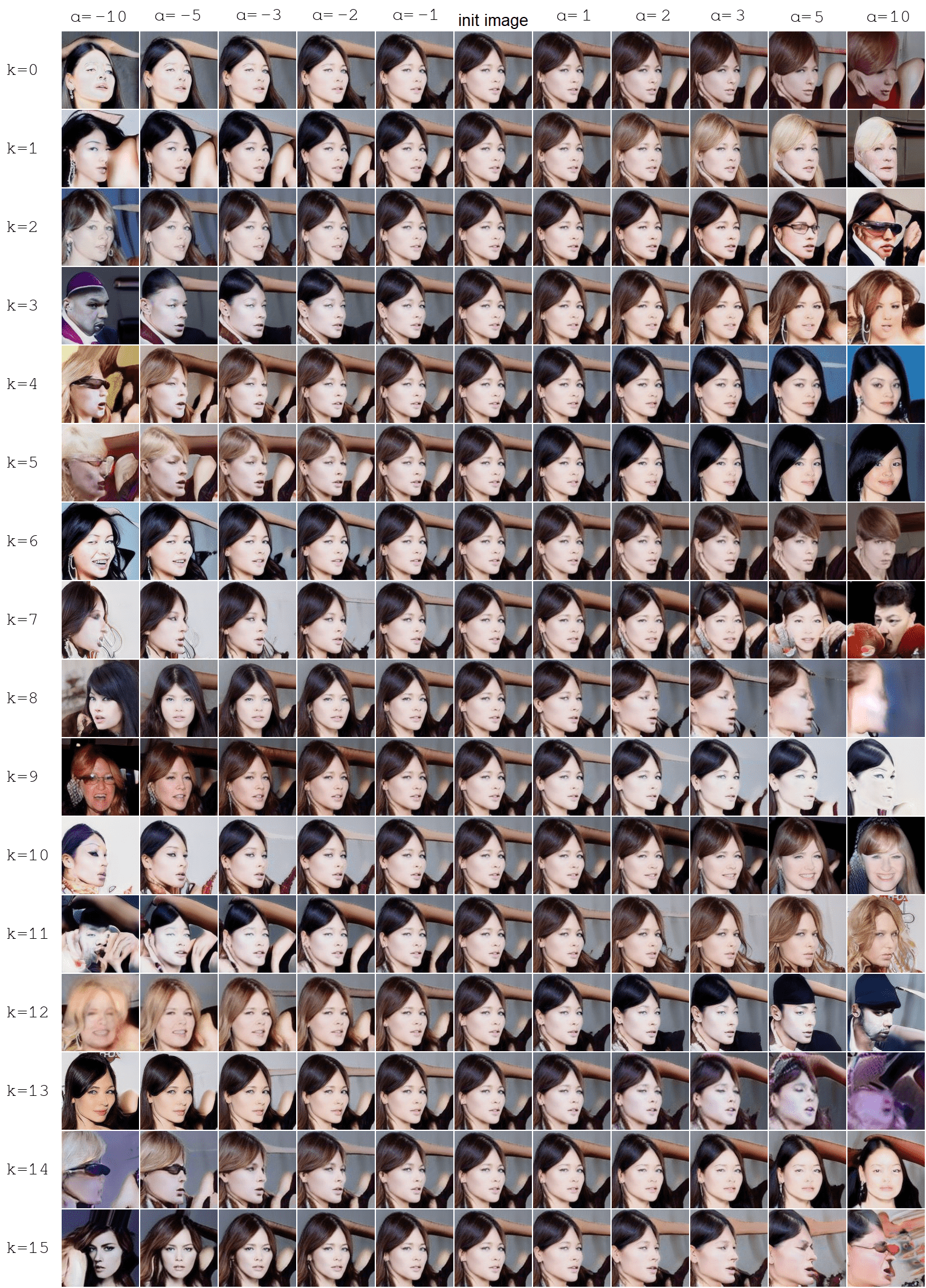}
         
        \caption{All 16 learned attributes of a Attribute Classifier ResNet model.}
        \label{fig:supp-singlegan-att}
\end{figure*}

\begin{figure*}
\centering
         \includegraphics[width=0.89\linewidth]{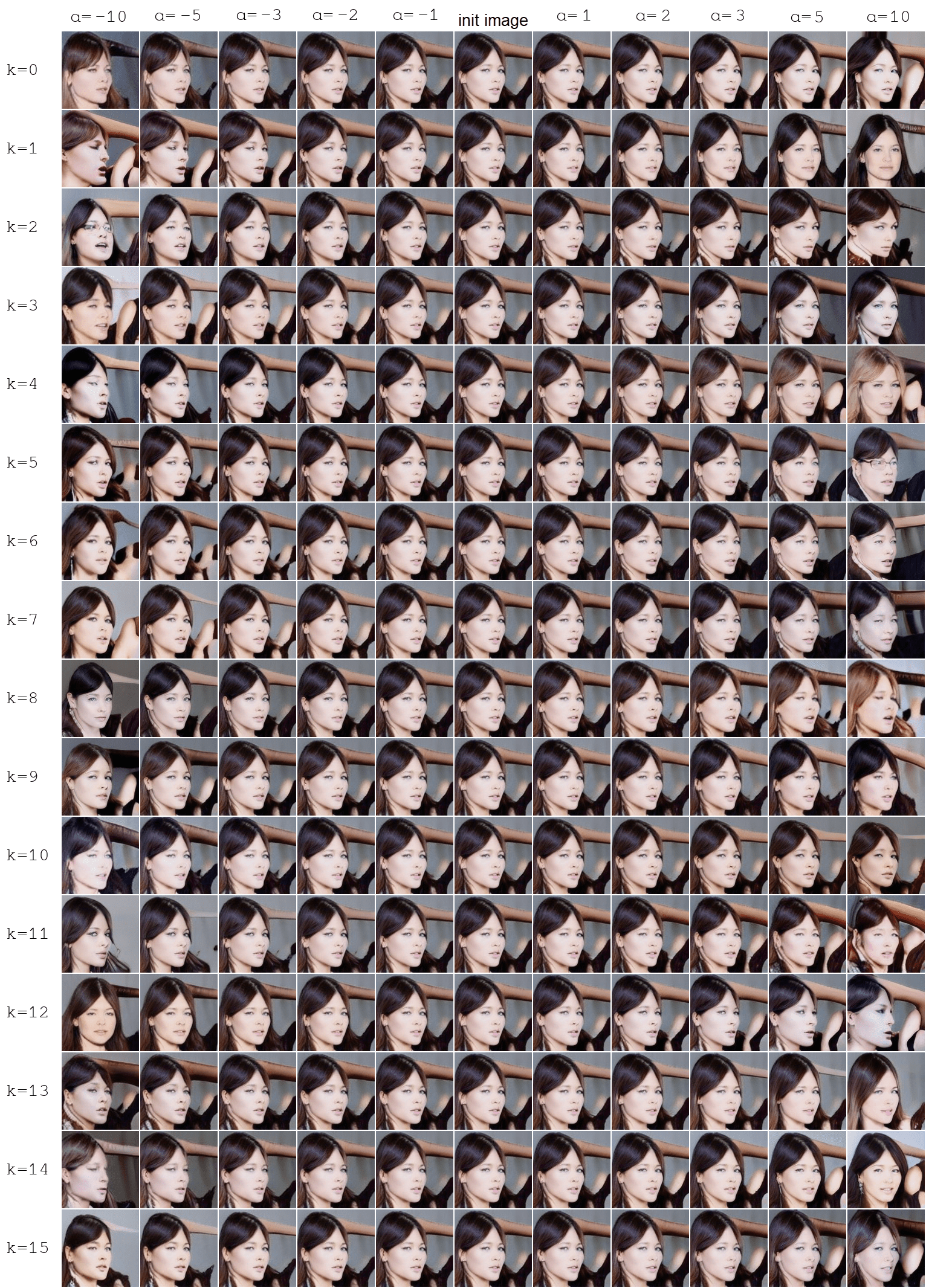}
         
        \caption{All 16 learned attributes of the original LatentCLR model (using global directions, rather than conditional).}
        \label{fig:supp-singlegan-globalclr}
\end{figure*}

\newpage
\newpage
\begin{table*}[]\centering
\vspace{25mm}
\begin{tabular}{lllllllll}
                       & Full CelebA & Female & Male  & No Hats & No Glasses & No Beards & \begin{tabular}[c]{@{}l@{}}No Beard\\ No Hats\end{tabular} & \begin{tabular}[c]{@{}l@{}}No Glasses\\ No Smiles\\ No Ties\end{tabular} \\ \hline
Full CelebA              &       & 0.143  & 0.143 & 0.045 & 0.111   & 0.063   & 0.278        & 0.189               \\
Female              & 0.000 &        & 0.167 & 1.000 & 0.167   & 0.200   & 0.170        & 0.417               \\
Male                & 0.000 & 0.059  &       & 0.250 & 0.111   & 0.250   & 0.188        & 0.303               \\
No Hats               & 0.000 & 0.056  & 0.111 &       & 0.083   & 0.067   & 0.188        & 0.267               \\
No Glasses            & 0.000 & 0.059  & 0.125 & 0.333 &         & 0.042   & 0.306        & 0.203               \\
No Beards             & 0.000 & 0.333  & 0.083 & 0.043 & 0.063   &         & 0.185        & 0.107               \\
\begin{tabular}[c]{@{}l@{}}No Beard\\ No Hats\end{tabular}         & 0.000 & 0.143  & 0.100 & 0.125 & 0.143   & 0.125   &              & 0.511               \\
\begin{tabular}[c]{@{}l@{}}No Glasses\\ No Smiles\\ No Ties\end{tabular} & 0.000 & 0.053  & 0.333 & 0.250 & 0.500   & 0.059   & 0.096        &     \\ \hline               
\end{tabular}
\caption{The full results for the recovery scores ($\mathcal{R}_{\text{score}}$) of the SeFa method.}
 \label{tbl:supp-unique-sefa}
\end{table*}

\begin{table*}[]\centering
\begin{tabular}{lllllllll}
                       & Full CelebA & Female & Male  & No Hats & No Glasses & No Beards & \begin{tabular}[c]{@{}l@{}}No Beard\\ No Hats\end{tabular} & \begin{tabular}[c]{@{}l@{}}No Glasses\\ No Smiles\\ No Ties\end{tabular} \\ \hline
Full CelebA            & -           & 0.478  & 0.536 & 0.086   & 0.390      & 0.388     & 0.120                                                      & 0.287                                                                    \\
Female                 & 0           & -      & 0.246 & 0.048   & 0.050      & 0.046     & 0.048                                                      & 0.279                                                                    \\
Male                   & 0           & 0.586  & -     & 0.124   & 0.333      & 0.733     & 0.384                                                      & 0.405                                                                    \\
No Hats                & 0           & 0.251  & 0.240 & -       & 0.097      & 0.180     & 0.108                                                      & 0.313                                                                    \\
No Glasses             & 0           & 0.585  & 0.542 & 0.048   & -          & 0.114     & 0.100                                                      & 0.218                                                                    \\
No Beards              & 0           & 0.290  & 0.583 & 0.069   & 0.080      & -         & 0.066                                                      & 0.271                                                                    \\
\begin{tabular}[c]{@{}l@{}}No Beard\\ No Hats\end{tabular}          & 0           & 0.373  & 0.396 & 0.034   & 0.189      & 0.049     & -                                                          & 0.297                                                                    \\
\begin{tabular}[c]{@{}l@{}}No Glasses\\ No Smiles\\ No Ties\end{tabular} & 0           & 0.448  & 0.667 & 0.055   & 0.033      & 0.537     & 0.269                                                      & -  \\  \hline                                                                    
\end{tabular}
\caption{The full results for the recovery scores ($\mathcal{R}_{\text{score}}$) of the Jacobian loss.}
 \label{tbl:supp-unique-jacobian}
\end{table*}

\begin{table*}[]\centering
\begin{tabular}{lllllllll}
                       & Full CelebA & Female & Male  & No Hats & No Glasses & No Beards & \begin{tabular}[c]{@{}l@{}}No Beard\\ No Hats\end{tabular} & \begin{tabular}[c]{@{}l@{}}No Glasses\\ No Smiles\\ No Ties\end{tabular} \\ \hline
Full CelebA            & -           & 1.000  & 1.000 & 0.056   & 0.167      & 0.167     & 0.096                                                      & 0.407                                                                    \\
Female                 & 0           & -      & 0.200 & 0.056   & 0.045      & 0.036     & 0.057                                                      & 0.364                                                                    \\
Male                   & 0           & 0.333  & -     & 0.077   & 1.000      & 1.000     & 0.300                                                      & 0.300                                                                    \\
No Hats                & 0           & 0.500  & 0.250 & -       & 0.083      & 0.071     & 0.042                                                      & 0.370                                                                    \\
No Glasses             & 0           & 0.333  & 0.250 & 0.083   & -          & 0.111     & 0.071                                                      & 0.150                                                                    \\
No Beards              & 0           & 0.125  & 0.167 & 0.071   & 0.063      & -         & 0.046                                                      & 0.218                                                                    \\
\begin{tabular}[c]{@{}l@{}}No Beard\\ No Hats\end{tabular}          & 0           & 0.167  & 0.333 & 0.032   & 0.071      & 0.053     & -                                                          & 0.375                                                                    \\
\begin{tabular}[c]{@{}l@{}}No Glasses\\ No Smiles\\ No Ties\end{tabular} & 0           & 1.000  & 0.333 & 0.200   & 0.048      & 0.143     & 0.102                                                      & -    \\  \hline                                                                   
\end{tabular}
\caption{The full results for the recovery scores ($\mathcal{R}_{\text{score}}$) of the Hessian loss.}
 \label{tbl:supp-unique-hessian}
\end{table*}

\begin{table*}[]\centering
\begin{tabular}{lllllllll}
                       & Full CelebA & Female & Male  & No Hats & No Glasses & No Beards & \begin{tabular}[c]{@{}l@{}}No Beard\\ No Hats\end{tabular} & \begin{tabular}[c]{@{}l@{}}No Glasses\\ No Smiles\\ No Ties\end{tabular} \\ \hline
Full CelebA            & -           & 1.000  & 1.000 & 0.250   & 0.333      & 0.200     & 0.153                                                      & 0.537                                                                    \\
Female                 & 0           & -      & 0.250 & 0.048   & 0.034      & 0.050     & 0.044                                                      & 0.137                                                                    \\
Male                   & 0           & 1.000  & -     & 0.143   & 0.500      & 0.333     & 0.267                                                      & 0.242                                                                    \\
No Hats                & 0           & 0.250  & 1.000 & -       & 0.125      & 0.167     & 0.077                                                      & 0.158                                                                    \\
No Glasses             & 0           & 0.333  & 1.000 & 0.038   & -          & 0.250     & 0.525                                                      & 0.153                                                                    \\
No Beards              & 0           & 0.125  & 0.500 & 0.045   & 0.063      & -         & 0.049                                                      & 0.381                                                                    \\
\begin{tabular}[c]{@{}l@{}}No Beard\\ No Hats\end{tabular}          & 0           & 1.000  & 1.000 & 0.043   & 0.167      & 0.091     & -                                                          & 0.389                                                                    \\
\begin{tabular}[c]{@{}l@{}}No Glasses\\ No Smiles\\ No Ties\end{tabular} & 0           & 1.000  & 0.500 & 0.067   & 0.033      & 0.125     & 0.072                                                      & - \\  \hline                                                                      
\end{tabular}
\caption{The full results for the recovery scores ($\mathcal{R}_{\text{score}}$) of the LatentCLR loss.}
 \label{tbl:supp-unique-latentclr}
\end{table*}

\begin{table*}[]\centering
\begin{tabular}{lllllllll}
                       & Full CelebA & Female & Male  & No Hats & No Glasses & No Beards & \begin{tabular}[c]{@{}l@{}}No Beard\\ No Hats\end{tabular} & \begin{tabular}[c]{@{}l@{}}No Glasses\\ No Smiles\\ No Ties\end{tabular} \\ \hline
Full CelebA            & -           & 0.500  & 1.000 & 0.333   & 0.050      & 0.500     & 0.153                                                      & 0.259                                                                    \\
Female                 & 0           & -      & 0.143 & 0.045   & 0.167      & 0.038     & 0.061                                                      & 0.410                                                                    \\
Male                   & 0           & 0.500  & -     & 0.250   & 0.333      & 0.333     & 0.306                                                      & 0.256                                                                    \\
No Hats                & 0           & 0.500  & 0.143 & -       & 0.091      & 0.500     & 0.139                                                      & 0.511                                                                    \\
No Glasses             & 0           & 0.333  & 1.000 & 0.050   & -          & 0.167     & 0.094                                                      & 0.377                                                                    \\
No Beards              & 0           & 0.167  & 0.250 & 0.167   & 0.091      & -         & 0.052                                                      & 0.370                                                                    \\
\begin{tabular}[c]{@{}l@{}}No Beard\\ No Hats\end{tabular}          & 0           & 0.500  & 0.500 & 0.034   & 0.050      & 0.034     & -                                                          & 0.386                                                                    \\
\begin{tabular}[c]{@{}l@{}}No Glasses\\ No Smiles\\ No Ties\end{tabular} & 0           & 0.143  & 0.500 & 0.143   & 0.029      & 1.000     & 0.286                                                      & - \\  \hline                                                                      
\end{tabular}
\caption{The full results for the recovery scores ($\mathcal{R}_{\text{score}}$) of the voynov loss.}
 \label{tbl:supp-unique-voynov}
\end{table*}

\begin{table*}[] \centering
\begin{tabular}{lllllllll}
                       & Full CelebA & Female & Male  & No Hats & No Glasses & No Beards & \begin{tabular}[c]{@{}l@{}}No Beard\\ No Hats\end{tabular} & \begin{tabular}[c]{@{}l@{}}No Glasses\\ No Smiles\\ No Ties\end{tabular} \\ \hline
Full CelebA            & -           & 1.000  & 1.000 & 0.333   & 0.500      & 0.200     & 0.167         & 0.465                  \\
Female                 & 0           & -      & 0.333 & 0.053   & 0.067      & 0.034     & 0.046         & 0.381                  \\
Male                   & 0           & 1      & -     & 0.059   & 0.250      & 0.083     & 0.082         & 0.521                  \\
No Hats                & 0           & 1      & 0.5   & -       & 0.143      & 0.071     & 0.062         & 0.460                  \\
No Glasses             & 0           & 1      & 0.5   & 0.059   & -          & 0.091     & 0.081         & 0.377                  \\
No Beards              & 0           & 1      & 1     & 1       & 0.091      & -         & 0.154         & 0.378                  \\
\begin{tabular}[c]{@{}l@{}}No Beard\\ No Hats\end{tabular}          & 0           & 1      & 1     & 0.033   & 0.05       & 0.042     & -             & 0.382                  \\
\begin{tabular}[c]{@{}l@{}}No Glasses\\ No Smiles\\ No Ties\end{tabular} & 0           & 1      & 1     & 0.083   & 0.029      & 0.333     & 0.122         & -                      \\ \hline
\end{tabular}
\caption{The full results for the recovery scores ($\mathcal{R}_{\text{score}}$) of the ImageNet ResNet.}
 \label{tbl:supp-unique_vanilla}
\end{table*}

\begin{table*}[] \centering
\begin{tabular}{lllllllll}
                       & Full CelebA & Female & Male  & No Hats & No Glasses & No Beards & \begin{tabular}[c]{@{}l@{}}No Beard\\ No Hats\end{tabular} & \begin{tabular}[c]{@{}l@{}}No Glasses\\ No Smiles\\ No Ties\end{tabular} \\ \hline
Full CelebA            & -           & 1.000  & 0.333 & 1.000   & 0.091      & 1.000     & 0.525         & 0.390                  \\
Female                 & 0           & -      & 0.200 & 0.042   & 0.059      & 0.038     & 0.036         & 0.198                  \\
Male                   & 0           & 1      & -     & 0.063   & 0.143      & 0.056     & 0.306         & 0.492                  \\
No Hats                & 0           & 1      & 0.333 & -       & 0.333      & 1.000     & 0.516         & 0.365                  \\
No Glasses             & 0           & 1      & 0.333 & 0.1     & -          & 0.500     & 0.270         & 0.357                  \\
No Beards              & 0           & 1      & 1     & 0.053   & 0.111      & -         & 0.042         & 0.211                  \\
\begin{tabular}[c]{@{}l@{}}No Beard\\ No Hats\end{tabular}          & 0           & 1      & 1     & 0.033   & 0.071      & 0.048     & -             & 0.212                  \\
\begin{tabular}[c]{@{}l@{}}No Glasses\\ No Smiles\\ No Ties\end{tabular} & 0           & 1      & 1     & 1       & 0.03       & 0.5       & 0.274         & -                      \\ \hline
\end{tabular}
\caption{The full results for the recovery scores ($\mathcal{R}_{\text{score}}$) of the Attribute Classifier ResNet.}
 \label{tbl:supp-unique_att}
\end{table*}

\begin{table*}[] \centering
\begin{tabular}{lllllllll}
                       & Full CelebA & Female & Male  & No Hats & No Glasses & No Beards & \begin{tabular}[c]{@{}l@{}}No Beard\\ No Hats\end{tabular} & \begin{tabular}[c]{@{}l@{}}No Glasses\\ No Smiles\\ No Ties\end{tabular} \\ \hline
Full CelebA            & -           & 1.000  & 1.000 & 0.250   & 1.000      & 0.063     & 0.183         & 0.401                  \\
Female                 & 0           & -      & 0.333 & 0.500   & 0.042      & 0.048     & 0.047         & 0.363                  \\
Male                   & 0           & 0.5    & -     & 0.077   & 0.500      & 0.111     & 0.563         & 0.419                  \\
No Hats                & 0           & 1      & 1     & -       & 1.000      & 0.143     & 0.517         & 0.423                  \\
No Glasses             & 0           & 1      & 0.333 & 0.034   & -          & 0.333     & 0.200         & 0.361                  \\
No Beards              & 0           & 1      & 1     & 0.143   & 0.045      & -         & 0.113         & 0.370                  \\
\begin{tabular}[c]{@{}l@{}}No Beard\\ No Hats\end{tabular}          & 0           & 1      & 1     & 0.028   & 0.5        & 0.038     & -             & 0.376                  \\
\begin{tabular}[c]{@{}l@{}}No Glasses\\ No Smiles\\ No Ties\end{tabular} & 0           & 1      & 1     & 0.056   & 0.033      & 1         & 0.556         & -                      \\ \hline
\end{tabular}
\caption{The full results for the recovery scores ($\mathcal{R}_{\text{score}}$) of the Robust ResNet.}
 \label{tbl:supp-unique_robust}
\end{table*}

\begin{table*}[] \centering
\begin{tabular}{lllllllll}
                       & Full CelebA & Female & Male  & No Hats & No Glasses & No Beards & \begin{tabular}[c]{@{}l@{}}No Beard\\ No Hats\end{tabular} & \begin{tabular}[c]{@{}l@{}}No Glasses\\ No Smiles\\ No Ties\end{tabular} \\ \hline
Full CelebA            & -           & 1.000  & 1.000 & 1.000   & 0.200      & 0.200     & 0.108         & 0.383                  \\
Female                 & 0           & -      & 0.250 & 0.028   & 0.200      & 0.063     & 0.131         & 0.194                  \\
Male                   & 0           & 0.143  & -     & 0.028   & 0.200      & 0.063     & 0.131         & 0.055                  \\
No Hats                & 0           & 1      & 1     & -       & 0.500      & 0.125     & 0.133         & 0.231                  \\
No Glasses             & 0           & 1      & 1     & 0.037   & -          & 0.100     & 0.148         & 0.397                  \\
No Beards              & 0           & 1      & 1     & 0.083   & 0.333      & -         & 0.098         & 0.373                  \\
\begin{tabular}[c]{@{}l@{}}No Beard\\ No Hats\end{tabular}          & 0           & 1      & 1     & 0.1     & 0.2        & 0.029     & -             & 0.371                  \\
\begin{tabular}[c]{@{}l@{}}No Glasses\\ No Smiles\\ No Ties\end{tabular} & 0           & 1      & 1     & 0.063   & 0.125      & 0.167     & 0.205         & -                      \\ \hline
\end{tabular}
\caption{The full results for the recovery scores ($\mathcal{R}_{\text{score}}$) of the CLIP ResNet.}
 \label{tbl:supp-unique_rnclip}
\end{table*}

\begin{table*}[] \centering
\begin{tabular}{lllllllll}
                    & Full CelebA & Female & Male  & No Hats & No Glasses & No Beards & \begin{tabular}[c]{@{}l@{}}No Beard\\ No Hats\end{tabular} & \begin{tabular}[c]{@{}l@{}}No Glasses\\ No Smiles\\ No Ties\end{tabular} \\ \hline
Full CelebA            & -           & 1.000  & 1.000 & 0.200   & 0.091      & 0.050     & 0.276         & 0.363                  \\
Female                 & 0           & -      & 0.333 & 0.040   & 0.045      & 0.056     & 0.054         & 0.140                  \\
Male                   & 0           & 1      & -     & 0.500   & 0.111      & 0.063     & 0.140         & 0.199                  \\
No Hats                & 0           & 0.5    & 1     & -       & 0.200      & 0.040     & 0.076         & 0.369                  \\
No Glasses             & 0           & 0.333  & 1     & 0.111   & -          & 0.063     & 0.042         & 0.388                  \\
No Beards              & 0           & 0.143  & 0.143 & 0.125   & 0.1        & -         & 0.086         & 0.209                  \\
\begin{tabular}[c]{@{}l@{}}No Beard\\ No Hats\end{tabular}          & 0           & 0.143  & 0.5   & 0.031   & 0.053      & 0.032     & -             & 0.192                  \\
\begin{tabular}[c]{@{}l@{}}No Glasses\\ No Smiles\\ No Ties\end{tabular} & 0           & 1      & 1     & 0.033   & 0.125      & 0.125     & 0.108         & -                      \\ \hline
\end{tabular}
\caption{The full results for the recovery scores ($\mathcal{R}_{\text{score}}$) of the ImageNet ViT.}
 \label{tbl:supp-unique_vit}
\end{table*}

\begin{table*}[] \centering
\begin{tabular}{lllllllll}
                       & Full CelebA & Female & Male  & No Hats & No Glasses & No Beards & \begin{tabular}[c]{@{}l@{}}No Beard\\ No Hats\end{tabular} & \begin{tabular}[c]{@{}l@{}}No Glasses\\ No Smiles\\ No Ties\end{tabular} \\ \hline
Full CelebA            & -           & 0.500  & 1.000 & 0.027   & 1.000      & 0.067     & 0.264         & 0.387                  \\
Female                 & 0           & -      & 0.200 & 0.034   & 0.032      & 0.036     & 0.048         & 0.188                  \\
Male                   & 0           & 0.333  & -     & 0.091   & 0.125      & 0.045     & 0.229         & 0.232                  \\
No Hats                & 0           & 1      & 1     & -       & 0.063      & 0.067     & 0.170         & 0.365                  \\
No Glasses             & 0           & 1      & 0.5   & 0.067   & -          & 0.053     & 0.066         & 0.189                  \\
No Beards              & 0           & 0.5    & 1     & 0.091   & 0.333      & -         & 0.073         & 0.127                  \\
\begin{tabular}[c]{@{}l@{}}No Beard\\ No Hats\end{tabular}          & 0           & 0.2    & 0.5   & 0.029   & 0.056      & 0.038     & -             & 0.369                  \\
\begin{tabular}[c]{@{}l@{}}No Glasses\\ No Smiles\\ No Ties\end{tabular} & 0           & 1      & 1     & 0.032   & 0.125      & 0.037     & 0.107         & -                      \\ \hline
\end{tabular}
\caption{The full results for the recovery scores ($\mathcal{R}_{\text{score}}$) of the CLIP ViT.}
 \label{tbl:supp-unique_clipvit}
\end{table*}

\begin{table*}[] \centering
\begin{tabular}{lllllllll}
                       & Full CelebA & Female & Male  & No Hats & No Glasses & No Beards & \begin{tabular}[c]{@{}l@{}}No Beard\\ No Hats\end{tabular} & \begin{tabular}[c]{@{}l@{}}No Glasses\\ No Smiles\\ No Ties\end{tabular} \\ \hline
Full CelebA            & -           & 0.200  & 0.250 & 0.100   & 0.077      & 0.077     & 0.156         & 0.068                  \\
Female                 & 0           & -      & 0.200 & 0.038   & 0.056      & 0.033     & 0.044         & 0.119                  \\
Male                   & 0           & 1      & -     & 0.053   & 0.143      & 0.042     & 0.104         & 0.511                  \\
No Hats                & 0           & 0.091  & 1     & -       & 0.063      & 0.028     & 0.046         & 0.194                  \\
No Glasses             & 0           & 0.25   & 1     & 0.067   & -          & 0.034     & 0.047         & 0.081                  \\
No Beards              & 0           & 0.5    & 1     & 0.045   & 0.083      & -         & 0.072         & 0.356                  \\
\begin{tabular}[c]{@{}l@{}}No Beard\\ No Hats\end{tabular}          & 0           & 0.25   & 0.333 & 0.067   & 0.059      & 0.034     & -             & 0.068                  \\
\begin{tabular}[c]{@{}l@{}}No Glasses\\ No Smiles\\ No Ties\end{tabular} & 0           & 1      & 0.5   & 0.027   & 0.091      & 0.043     & 0.163         & -                      \\ \hline
\end{tabular}
\caption{The full results for the recovery scores ($\mathcal{R}_{\text{score}}$) of the MAE ViT.}
 \label{tbl:supp-unique_mae}
\end{table*}


\begin{table*}[] \centering
\begin{tabular}{llllllll}
              & Female & Male  & No Hats & No Glasses & No Beards & \begin{tabular}[c]{@{}l@{}}No Beard\\ No Hats\end{tabular} & \begin{tabular}[c]{@{}l@{}}No Glasses\\ No Smiles\\ No Ties\end{tabular} \\ \hline
Full CelebA        & 0.413 & 0.458 & 0.3715 & 0.374 & 0.314 & 0.387     & 0.355            \\
Female       & -        & 0.329 & 0.3615 & 0.320 & 0.352 & 0.360     & 0.334            \\
Male         & -        & -        & 0.3732 & 0.426 & 0.364 & 0.352     & 0.381            \\
No Hats      & -        & -        & -        & 0.400 & 0.300 & 0.325     & 0.379            \\
No Glasses   & -        & -        & -        & -        & 0.336 & 0.305     & 0.343            \\
No Beards    & -        & -        & -        & -        & -        & 0.302     & 0.295            \\
\begin{tabular}[c]{@{}l@{}}No Beard\\ No Hats\end{tabular} & -        & -        & -        & -        & -        & -            & 0.341            
\end{tabular}
\caption{The full results for the alignment scores ($\mathcal{A}_{\text{score}}$) of the SeFa method.}
 \label{tbl:supp-cosine-sefa}
\end{table*}

\begin{table*}[]\centering
\begin{tabular}{llllllll}
              & Female & Male  & No Hats & No Glasses & No Beards & \begin{tabular}[c]{@{}l@{}}No Beard\\ No Hats\end{tabular} & \begin{tabular}[c]{@{}l@{}}No Glasses\\ No Smiles\\ No Ties\end{tabular} \\ \hline
Full CelebA   & 0.475  & 0.489 & 0.652   & 0.598      & 0.618     & 0.615                                                      & 0.525                                                                    \\
Female        & -      & 0.367 & 0.523   & 0.457      & 0.559     & 0.559                                                      & 0.345                                                                    \\
Male          & -      & -     & 0.466   & 0.401      & 0.372     & 0.431                                                      & 0.450                                                                    \\
No Hats       & -      & -     & -       & 0.512      & 0.620     & 0.505                                                      & 0.518                                                                    \\
No Glasses    & -      & -     & -       & -          & 0.556     & 0.502                                                      & 0.503                                                                    \\
No Beards     & -      & -     & -       & -          & -         & 0.563                                                      & 0.477                                                                    \\
\begin{tabular}[c]{@{}l@{}}No Beard\\ No Hats\end{tabular} & -      & -     & -       & -          & -         & -                                                          & 0.439                                                                    \\  \hline
\end{tabular}

\caption{The full results for the alignment scores ($\mathcal{A}_{\text{score}}$) of the Hessian loss.}
 \label{tbl:supp-cosine-hessian}
\end{table*}

\begin{table*}[]\centering
\begin{tabular}{llllllll}
              & Female & Male  & No Hats & No Glasses & No Beards & \begin{tabular}[c]{@{}l@{}}No Beard\\ No Hats\end{tabular} & \begin{tabular}[c]{@{}l@{}}No Glasses\\ No Smiles\\ No Ties\end{tabular} \\ \hline
Full CelebA   & 0.519  & 0.511 & 0.556   & 0.533      & 0.512     & 0.593                                                      & 0.579                                                                    \\
Female        & -      & 0.412 & 0.452   & 0.513      & 0.550     & 0.613                                                      & 0.388                                                                    \\
Male          & -      & -     & 0.373   & 0.474      & 0.425     & 0.448                                                      & 0.426                                                                    \\
No Hats       & -      & -     & -       & 0.460      & 0.485     & 0.576                                                      & 0.482                                                                    \\
No Glasses    & -      & -     & -       & -          & 0.484     & 0.630                                                      & 0.491                                                                    \\
No Beards     & -      & -     & -       & -          & -         & 0.550                                                      & 0.501                                                                    \\
\begin{tabular}[c]{@{}l@{}}No Beard\\ No Hats\end{tabular} & -      & -     & -       & -          & -         & -                                                          & 0.496                                                                    \\ \hline
\end{tabular}

\caption{The full results for the alignment scores ($\mathcal{A}_{\text{score}}$) of the LatentCLR loss.}
 \label{tbl:supp-cosine-latentclr}
\end{table*}

\begin{table*}[]\centering
\begin{tabular}{llllllll}
              & Female & Male  & No Hats & No Glasses & No Beards & \begin{tabular}[c]{@{}l@{}}No Beard\\ No Hats\end{tabular} & \begin{tabular}[c]{@{}l@{}}No Glasses\\ No Smiles\\ No Ties\end{tabular} \\ \hline
Full CelebA   & 0.566  & 0.477 & 0.567   & 0.555      & 0.570     & 0.562                                                      & 0.513                                                                    \\
Female        & -      & 0.430 & 0.606   & 0.609      & 0.577     & 0.592                                                      & 0.505                                                                    \\
Male          & -      & -     & 0.508   & 0.503      & 0.429     & 0.456                                                      & 0.453                                                                    \\
No Hats       & -      & -     &         & 0.596      & 0.553     & 0.583                                                      & 0.553                                                                    \\
No Glasses    & -      & -     & -       & -          & 0.572     & 0.559                                                      & 0.564                                                                    \\
No Beards     & -      & -     & -       & -          & -         & 0.616                                                      & 0.567                                                                    \\
\begin{tabular}[c]{@{}l@{}}No Beard\\ No Hats\end{tabular} & -      & -     & -       & -          & -         & -                                                          & 0.499                                                                    \\ \hline
\end{tabular}
 \caption{The full results for the alignment scores ($\mathcal{A}_{\text{score}}$) of the Voynov method.}
 \label{tbl:supp-cosine-voynov}
\end{table*}

\begin{table*}[]\centering
\begin{tabular}{llllllll}
              & Female & Male  & No Hats & No Glasses & No Beards & \begin{tabular}[c]{@{}l@{}}No Beard\\ No Hats\end{tabular} & \begin{tabular}[c]{@{}l@{}}No Glasses\\ No Smiles\\ No Ties\end{tabular} \\ \hline
Full CelebA   & 0.523  & 0.452 & 0.505   & 0.528      & 0.519     & 0.491                                                      & 0.495                                                                    \\
Female        & -      & 0.413 & 0.481   & 0.522      & 0.490     & 0.511                                                      & 0.449                                                                    \\
Male          & -      & -     & 0.488   & 0.451      & 0.443     & 0.409                                                      & 0.384                                                                    \\
No Hats       & -      & -     & -       & 0.498      & 0.515     & 0.523                                                      & 0.465                                                                    \\
No Glasses    & -      & -     & -       & -          & 0.497     & 0.517                                                      & 0.503                                                                    \\
No Beards     & -      & -     & -       & -          & -         & 0.544                                                      & 0.484                                                                    \\
\begin{tabular}[c]{@{}l@{}}No Beard\\ No Hats\end{tabular} & -      & -     & -       & -          & -         & -                                                          & 0.512                                                                    \\ \hline
\end{tabular}
 \caption{The full results for the alignment scores ($\mathcal{A}_{\text{score}}$) of the Jacobian loss.}
 \label{tbl:supp-cosine-jacobian}
\end{table*}

\begin{table*}[] \centering
\begin{tabular}{llllllll}
              & Female & Male  & No Hats & No Glasses & No Beards & \begin{tabular}[c]{@{}l@{}}No Beard\\ No Hats\end{tabular} & \begin{tabular}[c]{@{}l@{}}No Glasses\\ No Smiles\\ No Ties\end{tabular} \\ \hline
Full CelebA   & 0.457  & 0.403 & 0.740   & 0.792      & 0.461     & 0.643         & 0.489                  \\
Female        & -      & 0.409 & 0.651   & 0.720      & 0.599     & 0.483         & 0.444                  \\
Male          & -      & -     & 0.508   & 0.377      & 0.328     & 0.360         & 0.390                  \\
No Hats       & -      & -     & -       & 0.611      & 0.778     & 0.414         & 0.560                  \\
No Glasses    & -      & -     & -       & -          & 0.698     & 0.659         & 0.649                  \\
No Beards     & -      & -     & -       & -          & -         & 0.652         & 0.556                  \\
\begin{tabular}[c]{@{}l@{}}No Beard\\ No Hats\end{tabular} & -      & -     & -       & -          & -         & -             & 0.492                  \\ \hline
\end{tabular}
\caption{The full results for the alignment scores ($\mathcal{A}_{\text{score}}$) of the ImageNet Trained ResNet.}
 \label{tbl:supp-cosine-vanilla}
\end{table*}

\begin{table*}[] \centering
\begin{tabular}{llllllll}
              & Female & Male  & No Hats & No Glasses & No Beards & \begin{tabular}[c]{@{}l@{}}No Beard\\ No Hats\end{tabular} & \begin{tabular}[c]{@{}l@{}}No Glasses\\ No Smiles\\ No Ties\end{tabular} \\ \hline
Full CelebA   & 0.069  & 0.272 & 0.419   & 0.494      & 0.556     & 0.543         & 0.014                  \\
Female        & -      & 0.155 & 0.411   & 0.230      & 0.191     & 0.354         & 0.056                  \\
Male          & -      & -     & 0.248   & 0.392      & 0.086     & 0.338         & 0.231                  \\
No Hats       & -      & -     & -       & 0.370      & 0.279     & 0.346         & 0.569                  \\
No Glasses    & -      & -     & -       & -          & 0.494     & 0.405         & 0.447                  \\
No Beards     & -      & -     & -       & -          & -         & 0.331         & 0.510                  \\
\begin{tabular}[c]{@{}l@{}}No Beard\\ No Hats\end{tabular} & -      & -     & -       & -          & -         & -             & 0.403                  \\ \hline
\end{tabular}
\caption{The full results for the alignment scores ($\mathcal{A}_{\text{score}}$) of the Attribute Classifier ResNet.}
 \label{tbl:supp-cosine-att}
\end{table*}

\begin{table*}[] \centering
\begin{tabular}{llllllll}
              & Female & Male  & No Hats & No Glasses & No Beards & \begin{tabular}[c]{@{}l@{}}No Beard\\ No Hats\end{tabular} & \begin{tabular}[c]{@{}l@{}}No Glasses\\ No Smiles\\ No Ties\end{tabular} \\ \hline
Full CelebA   & 0.615  & 0.357 & 0.750   & 0.825      & 0.619     & 0.803         & 0.649                  \\
Female        & -      & 0.439 & 0.643   & 0.644      & 0.404     & 0.565         & 0.525                  \\
Male          & -      & -     & 0.384   & 0.444      & 0.417     & 0.175         & 0.289                  \\
No Hats       & -      & -     & -       & 0.508      & 0.310     & 0.744         & 0.641                  \\
No Glasses    & -      & -     & -       & -          & 0.717     & 0.682         & 0.557                  \\
No Beards     & -      & -     & -       & -          & -         & 0.568         & 0.495                  \\
\begin{tabular}[c]{@{}l@{}}No Beard\\ No Hats\end{tabular} & -      & -     & -       & -          & -         & -             & 0.474                  \\ \hline
\end{tabular}
\caption{The full results for the alignment scores ($\mathcal{A}_{\text{score}}$) of the Robust ResNet.}
 \label{tbl:supp-cosine-robust}
\end{table*}

\begin{table*}[] \centering
\begin{tabular}{llllllll}
              & Female & Male  & No Hats & No Glasses & No Beards & \begin{tabular}[c]{@{}l@{}}No Beard\\ No Hats\end{tabular} & \begin{tabular}[c]{@{}l@{}}No Glasses\\ No Smiles\\ No Ties\end{tabular} \\ \hline
Full CelebA   & 0.753  & 0.451 & 0.772   & 0.791      & 0.894     & 0.580         & 0.656                  \\
Female        & -      & 0.283 & 0.733   & 0.684      & 0.639     & 0.474         & 0.337                  \\
Male          & -      & -     & 0.371   & 0.435      & 0.396     & 0.337         & 0.314                  \\
No Hats       & -      & -     & -       & 0.815      & 0.679     & 0.715         & 0.505                  \\
No Glasses    & -      & -     & -       & -          & 0.748     & 0.608         & 0.623                  \\
No Beards     & -      & -     & -       & -          & -         & 0.706         & 0.507                  \\
\begin{tabular}[c]{@{}l@{}}No Beard\\ No Hats\end{tabular} & -      & -     & -       & -          & -         & -             & 0.723                  \\ \hline
\end{tabular}
\caption{The full results for the alignment scores ($\mathcal{A}_{\text{score}}$) of the CLIP ResNet.}
 \label{tbl:supp-cosine-clipRN}
\end{table*}

\begin{table*}[] \centering
\begin{tabular}{llllllll}
              & Female & Male  & No Hats & No Glasses & No Beards & \begin{tabular}[c]{@{}l@{}}No Beard\\ No Hats\end{tabular} & \begin{tabular}[c]{@{}l@{}}No Glasses\\ No Smiles\\ No Ties\end{tabular} \\ \hline
Full CelebA   & 0.308  & 0.417 & 0.433   & 0.486      & 0.455     & 0.409         & 0.416                  \\
Female        & -      & 0.251 & 0.348   & 0.476      & 0.468     & 0.495         & 0.365                  \\
Male          & -      & -     & 0.168   & 0.275      & 0.304     & 0.331         & 0.363                  \\
No Hats       & -      & -     & -       & 0.409      & 0.484     & 0.499         & 0.390                  \\
No Glasses    & -      & -     & -       & -          & 0.347     & 0.363         & 0.417                  \\
No Beards     & -      & -     & -       & -          & -         & 0.344         & 0.164                  \\
\begin{tabular}[c]{@{}l@{}}No Beard\\ No Hats\end{tabular} & -      & -     & -       & -          & -         & -             & 0.363                  \\ \hline
\end{tabular}
\caption{The full results for the alignment scores ($\mathcal{A}_{\text{score}}$) of the ImageNet ViT.}
 \label{tbl:supp-cosine-vit}
\end{table*}

\begin{table*}[] \centering
\begin{tabular}{llllllll}
              & Female & Male  & No Hats & No Glasses & No Beards & \begin{tabular}[c]{@{}l@{}}No Beard\\ No Hats\end{tabular} & \begin{tabular}[c]{@{}l@{}}No Glasses\\ No Smiles\\ No Ties\end{tabular} \\ \hline
Full CelebA   & 0.418  & 0.373 & 0.496   & 0.358      & 0.579     & 0.539         & 0.499                  \\
Female        & -      & 0.169 & 0.343   & 0.399      & 0.368     & 0.504         & 0.258                  \\
Male          & -      & -     & 0.183   & 0.381      & 0.213     & 0.187         & 0.225                  \\
No Hats       & -      & -     & -       & 0.448      & 0.577     & 0.422         & 0.513                  \\
No Glasses    & -      & -     & -       & -          & 0.529     & 0.385         & 0.440                  \\
No Beards     & -      & -     & -       & -          & -         & 0.509         & 0.440                  \\
\begin{tabular}[c]{@{}l@{}}No Beard\\ No Hats\end{tabular} & -      & -     & -       & -          & -         & -             & 0.368                  \\ \hline
\end{tabular}
\caption{The full results for the alignment scores ($\mathcal{A}_{\text{score}}$) of the CLIP ViT.}
 \label{tbl:supp-cosine-clipvit}
\end{table*}

\begin{table*}[] \centering
\begin{tabular}{llllllll}
              & Female & Male  & No Hats & No Glasses & No Beards & \begin{tabular}[c]{@{}l@{}}No Beard\\ No Hats\end{tabular} & \begin{tabular}[c]{@{}l@{}}No Glasses\\ No Smiles\\ No Ties\end{tabular} \\ \hline
Full CelebA   & 0.513  & 0.385 & 0.439   & 0.294      & 0.469     & 0.417         & 0.255                  \\
Female        & -      & 0.203 & 0.456   & 0.313      & 0.312     & 0.521         & 0.342                  \\
Male          & -      & -     & 0.371   & 0.286      & 0.238     & 0.201         & 0.306                  \\
No Hats       & -      & -     & -       & 0.322      & 0.411     & 0.378         & 0.302                  \\
No Glasses    & -      & -     & -       & -          & 0.459     & 0.427         & 0.286                  \\
No Beards     & -      & -     & -       & -          & -         & 0.301         & 0.324                  \\
\begin{tabular}[c]{@{}l@{}}No Beard\\ No Hats\end{tabular} & -      & -     & -       & -          & -         & -             & 0.236                  \\ \hline
\end{tabular}
\caption{The full results for the alignment scores ($\mathcal{A}_{\text{score}}$) of the MAE ViT.}
 \label{tbl:supp-cosine-mae}
\end{table*}

\end{document}